%% file: main.tex
\documentclass{article}

\PassOptionsToPackage{numbers, compress}{natbib}
\PassOptionsToPackage{hidelinks}{hyperref}  
\PassOptionsToPackage{hyphens}{url}         

\usepackage[final]{neurips_data_2024}

\input{packages}

\input{macros}
\input{math_commands}

\title{Probing the Critical Point (CritPt) of AI Reasoning: \\a Frontier Physics Research Benchmark}
\author{%
\textbf{Minhui Zhu$^{1*}$, Minyang Tian$^{1,2*}$, Xiaocheng Yang$^{2}$, Tianci Zhou$^{3}$, Lifan Yuan$^{2}$, } \\
\textbf{Penghao Zhu$^{4}$, Eli Chertkov$^{5}$, Shengyan Liu$^{2}$, Yufeng Du$^{2}$, Ziming Ji$^{6}$, Indranil Das$^{2}$, } \\
\textbf{Qingzhi Chen$^{2}$, Junyi Cao$^{2}$, Yufeng Du$^{7}$, Jiabin Yu$^{8}$, Peixue Wu$^{9}$, Jinchen He$^{1,10}$, Yifan Su$^{11}$, } \\
\textbf{Yikun Jiang$^{6}$, Yujie Zhang$^{12,9}$, Chang Liu$^{13}$, Ze-Min Huang$^{14}$, Weizhen Jia$^{15}$, Yunkai Wang$^{12,9}$, } \\
\textbf{Farshid Jafarpour$^{16}$, Yong Zhao$^{1}$, Xinan Chen$^{2}$, Jessie Shelton$^{2}$, Aaron W. Young$^{17}$, } \\
\textbf{John Bartolotta$^{5}$, Wenchao Xu$^{18,19}$, Yue Sun$^{20}$, Anjun Chu$^{21}$, Victor Colussi$^{5}$,} \\
\textbf{Chris Akers$^{22}$, Nathan Brooks$^{23}$, Wenbo Fu$^{2}$, Jinchao Zhao$^{24}$, Marvin Qi$^{21}$, Anqi Mu$^{11}$, } \\
\textbf{Yubo Yang$^{25}$, Allen Zang$^{21}$, Yang Lyu$^{26}$, Peizhi Mai$^{2}$, Christopher Wilson$^{22}$, Xuefei Guo$^{2}$, } \\
\textbf{Juntai Zhou$^{2}$, Daniel Inafuku$^{2}$, Chi Xue$^{5}$, Luyu Gao$^{27}$, Ze Yang$^{2}$, Ya\"ir Hein$^{16}$, } \\
\textbf{Yonatan Kahn$^{28,29}$, Kevin Zhou$^{26}$, Di Luo$^{30}$, John Drew Wilson$^{22}$, Jarrod T. Reilly$^{22}$, } \\
\textbf{Dmytro Bandak$^{5}$, Ofir Press$^{31}$, Liang Yang$^{32}$, Xueying Wang$^{33}$, Hao Tong$^{2}$, } \\
\textbf{Nicolas Chia$^{1}$, Eliu Huerta$^{1,2,21}$, Hao Peng$^{2}$}
\vspace{0.5em}\\
$^{1}$Argonne National Laboratory \hspace{0.2em} $^{2}$University of Illinois Urbana-Champaign \hspace{0.2em} $^{3}$Virginia Tech \hspace{0.2em} \\
$^{4}$Ohio State University \hspace{0.2em} $^{5}$Independent \hspace{0.2em} $^{6}$Northeastern University \hspace{0.2em} $^{7}$Caltech \hspace{0.2em} \\
$^{8}$University of Florida \hspace{0.2em} $^{9}$University of Waterloo \hspace{0.2em} $^{10}$University of Maryland, College Park \hspace{0.2em} \\
 $^{11}$Columbia University \hspace{0.2em} $^{12}$Perimeter Institute for Theoretical Physics \hspace{0.2em} \\
$^{13}$University of Connecticut \hspace{0.2em} $^{14}$University of Cologne \hspace{0.2em} $^{15}$The Chinese University of Hong Kong \hspace{0.2em} \\
$^{16}$Utrecht University \hspace{0.2em} $^{17}$Harvard University \hspace{0.2em} $^{18}$ETH Zürich \hspace{0.2em} $^{19}$Paul Scherrer Institute \hspace{0.2em} \\
 $^{20}$University of Washington Seattle \hspace{0.2em} $^{21}$University of Chicago \hspace{0.2em} $^{22}$University of Colorado Boulder \hspace{0.2em} \\
 $^{23}$Chi 3 Optics \hspace{0.2em} $^{24}$Hong Kong University of Science and Technology \hspace{0.2em} $^{25}$Hofstra University \\
 \hspace{0.2em} $^{26}$University of California, Berkeley \hspace{0.2em} $^{27}$Carnegie Mellon University \hspace{0.2em} \\
$^{28}$University of Toronto \hspace{0.2em} $^{29}$Vector Institute \hspace{0.2em} $^{30}$University of California, Los Angeles \hspace{0.2em} \\
$^{31}$Princeton University \hspace{0.2em} $^{32}$University of California San Diego \hspace{0.2em} \\
$^{33}$National Institute of Theory and Mathematics in Biology \hspace{0.2em}
\vspace{0.2em} \\
$^{*}$ Equal contribution lead authors. \\
Correspondence to
\href{mailto:minhui.zhu@anl.gov}{minhui.zhu@anl.gov},
\href{mailto:mtian8@illinois.edu}{mtian8@illinois.edu}.
\vspace{0.3em} \\
\textcolor{Classic_Blue}{\faGlobe} \href{https://critpt.com}{critpt.com} \hspace{0.5em}
\faGithub \;\href{https://github.com/CritPt-Benchmark/CritPt}{CritPt} \hspace{0.4em}
\includegraphics[height=1.0em]{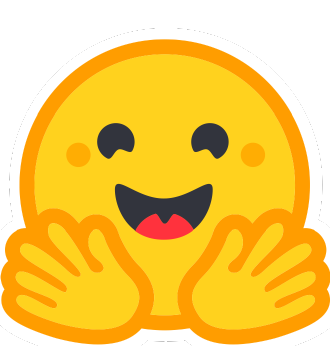} \href{https://huggingface.co/datasets/CritPt-Benchmark/CritPt}{CritPt}
}

\begin{document}
\maketitle
\input{text/00-abstract}
\input{text/01-intro}
\input{text/02-design}

\input{text/03-eval}

\input{text/04-result}
\input{text/05-conclusion}

\begin{ack}
We gratefully acknowledge insightful discussions with Nigel Goldenfeld and Carl M. Bender on physics reasoning and current AI limitations.
We also thank Zixiang Lu, Qiuling Fan, Semih Kacmaz, Ruixing Zhang, Xiaoning Wang, Sang Hyun Choi, Ethan Lake, Tong Wang, Carlo Siebenschuh, for their help on problem development.
We are appreciative of conversations with Xiao Chen, Andrew Lucas, Stephen Taylor, and Helvi Witek regarding problem formulation and student mentorship.
Finally, we thank the following physicists for sharing their experiences and perspectives on AI (in alphabetical order): Botao Du, Andrew Guo, Peter Johnsen, Zhiru Liu, Takumi Matsuzawa, Chenyu Tian, Zihan Wang, and Bo Zou.

We thank Artificial Analysis for hosting future evaluations on 70 \CritPt test challenges. The leaderboard will be continually updated at \url{https://artificialanalysis.ai/evaluations/critpt}.

This work was supported by Laboratory Directed Research and Development (LDRD) funding from Argonne National Laboratory, provided by the Director, Office of Science, of the U.S. Department of Energy (DOE) under Contract No. DE-AC02-06CH11357. 
We gratefully acknowledge computing resources provided by: the Argonne Leadership Computing Facility (ALCF), a DOE Office of Science User Facility under Contract DE-AC02-06CH11357;
the Delta advanced computing and data resources supported, by the National Science Foundation (NSF) (award OAC 2005572) and the State of Illinois. 
DeltaAI (award OAC 2320345) is a joint effort of the University of Illinois Urbana-Champaign and its National Center for Supercomputing Applications.
Eliu Huerta also acknowledges partial support from NSF grants OAC-2209892 and OAC-2514142. 
Yong Zhao's material is based upon work supported by the U.S. Department of Energy, Office of Science, Office of Nuclear Physics through Contract No.~DE-SC0012704, No.~DE-AC02-06CH11357, within the framework of Scientific Discovery through Advanced Computing (SciDAC) award Fundamental Nuclear Physics at the Exascale and Beyond, and under the umbrella of the Quark-Gluon Tomography (QGT) Topical Collaboration with Award DE-SC0023646.

 \end{ack}

\newpage
\appendix
\input{text/06-SI}

\newpage
\bibliographystyle{unsrtnat-abbr}
\bibliography{main}

\end{document}

%% file: packages.tex
\usepackage[utf8]{inputenc} 
\usepackage[T1]{fontenc}    
\usepackage[colorlinks=true,
            citecolor=black,      
            linkcolor=blue,     
            urlcolor=blue]{hyperref}
\usepackage{url}            
\usepackage{booktabs}       
\usepackage{amsfonts}       
\usepackage{nicefrac}       
\usepackage{microtype}      
\usepackage{xcolor}         
\usepackage{standalone}
\usepackage{latexsym}
\usepackage{amsmath}
\usepackage{amssymb}
\usepackage{amsthm}
\usepackage{csquotes}
\usepackage{graphicx}
\usepackage{subcaption}
\usepackage{array}
\usepackage{tabu}
\usepackage{makecell}
\usepackage{paralist}
\usepackage{cases}
\usepackage{diagbox}
\usepackage{enumitem}
\usepackage{soul}
\usepackage{multirow}
\usepackage{verbatim}
\usepackage{tabulary}
\usepackage{tabularx}
\usepackage[mathscr]{euscript}
\usepackage{mathtools}
\usepackage{algorithm}
\usepackage{algpseudocode}
\usepackage{stmaryrd}
\usepackage{tikz-dependency}
\usetikzlibrary{automata,decorations.markings,arrows,positioning,matrix,calc,patterns,angles,quotes,calc}
\usepackage{adjustbox}
\usepackage{xspace}
\usepackage{afterpage}
\usepackage{bm}
\usepackage{color}
\usepackage[toc,page]{appendix}
\usepackage{makecell}
\usepackage{boldline}
\usepackage[shortcuts]{extdash}  
\usepackage{comment}
\usepackage{blindtext}
\usepackage{capt-of}
\usepackage{varwidth}
\usepackage{pifont}
\usepackage{wrapfig}
\usepackage{pgfplots}
\usepackage{threeparttable}
\usepackage{caption}
\usepackage[table]{xcolor}
\usepackage{listings}
\usepackage{changepage}
\usepackage[most]{tcolorbox}
\usepackage{colortbl}        
\usepackage{pgfmath}         
\usepackage{xstring}
\usepackage{doi}
\usepackage{fontawesome5} 

\DeclareFixedFont{\ttb}{T1}{txtt}{bx}{n}{12} 
\DeclareFixedFont{\ttm}{T1}{txtt}{m}{n}{12}  

\definecolor{deepblue}{rgb}{0,0,0.5}
\definecolor{deepred}{rgb}{0.6,0,0}
\definecolor{deepgreen}{rgb}{0,0.5,0}
\definecolor{Classic_Blue}{RGB}{15, 76, 129}

\tcbuselibrary{minted,breakable,xparse,skins}

\definecolor{bg}{gray}{0.95}
\DeclareTCBListing{mintedbox}{O{}m!O{}}{%
  breakable=true,
  listing engine=minted,
  listing only,
  minted language=#2,
  minted style=default,
  minted options={%
    linenos,
    gobble=0,
    breaklines=true,
    breakafter=,,
    fontsize=\small,
    numbersep=8pt,
    #1},
  boxsep=0pt,
  left skip=0pt,
  right skip=0pt,
  left=25pt,
  right=0pt,
  top=3pt,
  bottom=3pt,
  arc=5pt,
  leftrule=0pt,
  rightrule=0pt,
  bottomrule=2pt,
  toprule=2pt,
  colback=bg,
  colframe=orange!70,
  enhanced,
  overlay={%
    \begin{tcbclipinterior}
    \fill[orange!20!white] (frame.south west) rectangle ([xshift=20pt]frame.north west);
    \end{tcbclipinterior}},
  #3}

\makeatletter
\newcommand*{\sectionbookmark}[1][]{%
  \bookmark[%
    level=section,%
    dest=\@currentHref,%
    #1%
  ]%
}
\makeatother

\usepackage{abstract}
\definecolor{myback}{RGB}{249,248,242}
\definecolor{myframe}{RGB}{58, 83, 135}

\newtcolorbox{exampleproblem}{
  enhanced,
  breakable,                 
  colback=myback,
  colframe=myframe,
  boxrule=1.0pt,
  arc=3mm,
  left=10pt,right=10pt,top=8pt,bottom=10pt,
  before skip=12pt,   
  after skip=12pt,    
  fontupper=\setlength{\parskip}{4pt}\setlength{\parindent}{0pt}
}

\newlist{hanglist}{itemize}{1}
\setlist[hanglist]{%
  label={},            
  labelsep=0pt,        
  leftmargin=2em,      
  itemindent=-2em,     
  listparindent=0pt,   
  parsep=1.5pt,          
  itemsep=0.5\baselineskip  
}

%% file: macros.tex
\DeclareSymbolFont{extraup}{U}{zavm}{m}{n}
\DeclareMathSymbol{\vardiamond}{\mathalpha}{extraup}{87}

\newcolumntype{L}[1]{>{\raggedright\let\newline\\\arraybackslash\hspace{0pt}}m{#1}}
\newcolumntype{C}[1]{>{\centering\let\newline\\\arraybackslash\hspace{0pt}}m{#1}}
\newcolumntype{R}[1]{>{\raggedleft\let\newline\\\arraybackslash\hspace{0pt}}m{#1}}

\algrenewcommand{\algorithmiccomment}[1]{\leavevmode$\triangleright$ #1}

\setul{1pt}{.4pt}

\DeclareFixedFont{\ttb}{T1}{txtt}{bx}{n}{12} 
\DeclareFixedFont{\ttm}{T1}{txtt}{m}{n}{12}  

\definecolor{lightgrey}{rgb}{0.9,0.9,0.9}
\definecolor{darkgrey}{rgb}{0.4,0.4,0.4}

\definecolor{myPurple}{RGB}{150,50,220} 

\newcommand{\CritPt}{\textit{CritPt}\xspace}

\definecolor{promptbg}{rgb}{0.96, 0.97, 0.98}   
\definecolor{promptframe}{rgb}{0.3, 0.45, 0.65} 

\definecolor{Red-Orange}{RGB}{226, 178, 168}
\definecolor{Yellow}{RGB}{240, 229, 179}
\definecolor{Teal}{RGB}{179, 217, 217}

\newcommand{\colorscore}[1]{%
  \begingroup
  \StrLeft{#1}{1}[\first]%
  \ifnum\first<2\relax
    \cellcolor{Red-Orange}{#1}%
  \else\ifnum\first<4\relax
    \cellcolor{Yellow}{#1}%
  \else
    \cellcolor{Teal}{#1}%
  \fi\fi
  \endgroup
}

%% file: math_commands.tex
\usepackage{amsmath,amsfonts,bm}

\def\eqref#1{equation~\ref{#1}}

\def\1{\bm{1}}

\DeclareMathAlphabet{\mathsfit}{\encodingdefault}{\sfdefault}{m}{sl}
\SetMathAlphabet{\mathsfit}{bold}{\encodingdefault}{\sfdefault}{bx}{n}



%% file: text/00-abstract.tex
\begin{abstract}

While large language models (LLMs) with reasoning capabilities are progressing rapidly on high-school math competitions and coding, can they reason effectively through complex, open-ended challenges found in frontier physics research?
And crucially, what kinds of reasoning tasks do physicists actually want LLMs to assist with?
To address these questions, we present the \textbf{\CritPt} (\textit{Complex Research using Integrated Thinking - Physics Test}, pronounced ``critical point''), the first benchmark designed to test LLMs on unpublished, research-level reasoning tasks that broadly covers modern physics research areas, including condensed matter, quantum physics, atomic, molecular \& optical physics, astrophysics, high energy physics, mathematical physics, statistical physics, nuclear physics, nonlinear dynamics, fluid dynamics and biophysics. 
\CritPt consists of 71 composite research challenges designed to simulate full-scale research projects at the entry level, which are also decomposed to 190 simpler checkpoint tasks for more fine-grained insights.
All problems are newly created by over 50 active physics researchers based on their own research.
Every problem is hand-curated to admit a guess-resistant and machine-verifiable answer and is evaluated by an automated grading pipeline heavily customized for advanced physics-specific output formats.
We find that while current state-of-the-art LLMs show early promise on isolated checkpoints, they remain far from being able to reliably solve full research-scale challenges: the best average accuracy among base models is only 5.7\% , achieved by GPT-5 (high), moderately rising to around 10\% when equipped with coding tools. 
We provide a secured public evaluation pipeline and online grading server, enabling independent validation and continual monitoring of frontier model progress.
Through the realistic yet standardized evaluation offered by \CritPt, we highlight a large disconnect between current model capabilities and realistic physics research demands, offering a foundation to guide the development of scientifically grounded AI tools.
\end{abstract}

%% file: text/01-intro.tex
\section{Introduction}

Modern physics research encounters increasingly complex systems, specialized tools, and interdisciplinary collaborations~\cite{anderson1972more, sinatra2015century}. 
Yet the core standards cannot be compromised: mathematical rigor, true creativity, precise execution and consistency between theory and experiment are all essential. 
The reality and demanding nature of the subject together set a high entry bar to become a physics researcher and make true breakthroughs more and more difficult to achieve.

Large language models (LLMs)~\cite{vaswani2017attention,devlin2019bert,raffel2020exploring} show promise in assisting research workflows, for example by identifying relevant literature~\cite{delgado2025transforming,scherbakov2025emergence,pramanick2024spiqa}, synthesizing scientific knowledge across domains~\cite{gao2023enabling,wang2024autosurvey,asai2024openscholar,skarlinski2024language,cuicurie}.
However, these applications largely involve \textit{recombining existing information}
and differ fundamentally from the kind of \textit{original reasoning} required to solve research problems in physics. 
And unlike in natural language tasks where redundancy may mask shallow errors, math and science problems can be unforgiving: a single flawed inference can invalidate the entire solution.

Recently, reasoning-oriented LLMs\footnote{In this paper, reasoning-oriented models including GPT-5 (high), o3, o4-mini, Gemini 2.5 Pro/Flash, DeepSeek R1 and Claude Opus 4 are evaluated~\cite{gpt5,o3ando4-mini,gemni-2-5,guo2025Deepseek,claude4}. General-purpose chat models, including GPT-5 (minimal), GPT-4o and Llama-4 Maverick~\cite{gpt5,gpt4o, llama4} are also included for comparison.} 
have made progress on structured multi-step problem solving~\cite{jaech2024openai, guo2025Deepseek}. 
These systems usually use encapsulated think tokens as an intermediate process before generating a final answer~\cite{guo2025Deepseek,comanici2025gemini,yang2025qwen3}.
They are typically fine-tuned on STEM-focused corpora and optimized for multi-step reasoning tasks, using techniques such as Chain-of-Thought prompting~\cite{wei2022chain}, reinforcement learning from verifiable rewards~\cite{guo2025Deepseek}, tool use \cite{schick2023toolformer} including code execution \cite{wang2024executable,yuan2024craft} and web search \cite{lewis2020rag}, and scaling inference-time computation \cite{wang2023selfconsistency,snell2025scaling}.
Empirically, these models exhibit behaviors that resemble human reasoning, such as decomposing long problems into coherent substeps, exploring alternatives through trial-and-error, and verifying intermediate results with internal heuristics or external checkers \cite{hazrahave, gandhi2025cognitive,guo2025Deepseek}.
As a result, these models see striking gains in relatively well-structured reasoning tasks, such as general coding~\cite{chen2021evaluating,austin2021program,jimenezswe}, high-school academic competitions~\cite{google-imo,el2025competitive,balunovic2025matharena,he2024olympiadbench,jainlivecodebench,qiu2025phybench}, as well as assignments from high school to graduate-level courses~\cite{hendrycks2measuring, hendrycks2020measuring,wang2024mmlu,wang2023scibench,xu2025ugphysics,rein2024gpqa}. 
Their performance remains limited but shows progress on expert-level mathematics and science benchmarks~\cite{tian2024scicode,glazer2024frontiermath,chung2025theoretical,phan2025humanity}.
This motivates a critical next step in AI evaluation:
assessing LLMs in authentic scientific research environments characterized by broad scopes, open-ended questions, deep decision trees and sparse solution spaces.
Realistically, can the latest LLMs meaningfully assist physicists with the reasoning tasks in frontier research? If so, to what extent and at what cost?

In this paper, we introduce \textbf{\CritPt} (\textit{Complex Research using Integrated Thinking - Physics Test}, pronounced ``critical point''), a benchmark designed to evaluate LLMs’ reasoning ability in realistic physics research workflows across diverse frontier topics. 
Our main evaluations are guided by the following lines of inquiry:

\begin{itemize}[itemsep=3pt,topsep=0pt,parsep=0pt,partopsep=0pt,label=$\circ$]

\item \textbf{Can LLMs solve unseen physics research problems beyond their training data?}

The goal of research is to make new discoveries, not to repeat known exercises. 
While grand open problems are out of reach and hard to verify, can LLMs solve unseen entry-level research problems, where the basic methods and concepts are established, but require nontrivial synthesis and original reasoning to reach a full solution?

\item \textbf{What reasoning tasks can LLMs help with in realistic physics research workflows today?}

A full-scale research project can often be decomposed into smaller steps or first addressed in a simplified form. 
In collaborative settings, these modular tasks can be distributed across team members.
What types of modular reasoning tasks may LLMs start to assist today? 

\item \textbf{Can we trust LLMs' reasoning traces and responses in physics research contexts?} 

Physics concepts and methods are deeply tied to context-dependent assumptions, where subtle errors in seemingly plausible answers can mislead, especially for those without expert judgment.
As a prerequisite check, how reliable are LLMs when tackling complex, unstructured problems in advanced physics, particularly those lying at the boundary of their current capabilities?
\end{itemize}

\CritPt provides a powerful framework for assessing 
the value of LLMs in realistic physics research workflows, an essential but underexplored component in defining AI's future role in scientific discovery~\cite{wang2023scientific}.
\CritPt contains 71 complex, composite \textit{challenges} to simulate full-scale research projects at the entry level, and 190 modular \textit{checkpoints} decomposed from the full challenges to offer more traceable and fine-grained insights on simpler subtasks.

However, designing a benchmark to meet these goals comes with significant practical and technical obstacles.
In \CritPt, we come up with the following design features to address these key obstacles: 

\begin{itemize}[itemsep=5pt,topsep=0pt,parsep=1pt,partopsep=0pt]
\item \textbf{Frontier research problems standardized by physics experts.}

Research-level problems are underrepresented in LLM benchmarks, as they demand significantly more domain expertise to adapt and validate than textbook-style problems. 
Consequently, our understanding of AI's ability in physics tends to center around well-structured problems or focus on a specific discipline, overlooking the complex and open-ended reasoning ability needed for real scientific discovery.

To better represent the depth and breadth of modern physics research, \CritPt is developed through a 9-month close collaboration between AI researchers and physics experts from nearly all major physics subfields (Sec.~\ref{sec:source}). 
Together, we iteratively co-design the content choice, dataset structure and evaluation infrastructure through multiple review stages (Sec.~\ref{sec:review}).
All the problems are based on experts' own research expertise to faithfully represent the \textbf{realistic} reasoning demands at the frontier of modern physics, while offering practical signals and accessible insights for the LLM developers.

\item \textbf{Leakage-resistant and reasoning-focused design.} 

Benchmarks sourced directly from public materials or generated from LLMs are susceptible to contamination since such materials are often included in model training data. 
This potentially leads to inflated performance via memorization or retrieval rather than \textit{genuine reasoning}~\cite{wu2024reasoning,balepur2024artifacts,deng2024investigating}, and often suffers from quick saturation and lack of utility value~\cite{ott2022mapping}.
Further, simple problem formats such as multiple-choice flattens problem complexity, allowing shortcut guessing and easy hacking~\cite{li2024open,balepur-etal-2025-best}. 
Fully open sourced benchmarks can be compromised over time, due to misuse or unintended contamination~\cite{dodge2021documenting,golchintime,roberts2023cutoff}.

We mitigate these risks through strict design criteria (Sec~\ref{sec:criteria}). 
All problems in the \CritPt are \textbf{unpublished}, hand-curated by physics experts to be well-defined and self-contained with \textbf{search-proof} answers.
In addition to an \textbf{open-ended} question format, problems are constructed to have \textbf{guess-resistant} final answers such as arrays of floating-point numbers and complicated symbolic expressions~\cite{glazer2024frontiermath}. 
We publish one example challenge with checkpoints, solutions, and error analysis (Sec.~\ref{sec:structure}).
The solutions to the other 70 challenges (test set) are kept private to avoid potential contamination.

\item \textbf{Physics-informed scalable auto-grading pipeline.} 

Grading physics problems is traditionally resource-intensive, requiring experts to verify all steps, recognize valid alternative paths, and detect subtle loopholes. 
Some use LLM judges, which can be unreliable due to sensitivity to superficial factors such as prompt wording or answer format, especially when evaluating content beyond the judge's own capacity~\cite{wang2024large,ye2024justice}.
An alternative is grading the final answer only.
However, for open-ended problems in advanced physics, even automating final-answer grading is complicated by technical challenges like parsing free-form LLM outputs and standardizing advanced physics notations~\cite{laskar-etal-2024-systematic}.

Our evaluation framework is canonical, scalable and physics-informed (Sec.~\ref{sec:eval}).
All final answers are \textbf{machine-verifiable} by careful design.
The LLM answers are first normalized into \textbf{structured} code blocks, then scored using \textbf{custom scripts} that supports numerical values, SymPy-compatible symbolic expressions~\cite{sympy}, and executable Python functions with test cases~\cite{tian2024scicode,chung2025theoretical}. 
Our autograder also accounts for physically meaningful \textbf{error tolerances} and equivalent forms specified by physics experts. 
We leverage this automation to host an online evaluation server, which securely facilitates independent, scalable validation by the research community alongside continual, timely monitoring by industry efforts.

\end{itemize}

Overall, our physics experts consider \CritPt challenges comparable in difficulty to the kind of warm-up research exercises that a hands-on principal investigator might assign to junior graduate students, which require solid physics training and some domain expertise, yet remain accessible through thoughtful exploration.
In this sense, this benchmark intends to probe the \textbf{critical point} of AI reasoning: the transition from producing plausible responses based on superficial pattern recognition, to genuinely reasoning through real-world problems in frontier physics research.

In Sec.~\ref{sec:res}, we show that current state-of-the-art LLMs are making early progress on isolated checkpoints, but remain far from being able to reliably solve full research-scale challenges.
Even the best-performing base model on \CritPt, GPT-5 (high), reaches only 5.7\% average accuracy on challenges, with most other models scoring near zero. 
When equipped with tools (code interpreter and web search), GPT-5 (high) improves modestly to 12.6\% accuracy.
More stringent evaluation metrics reveal that current LLMs still lack reliability when tackling research-level problems, underscoring the gap between today’s models and the demands of realistic physics research workflow.

%% file: text/02-design.tex
\section{Design choices of \CritPt}

We begin by describing the data sources and coverage of \CritPt in Sec.~\ref{sec:source}, followed by the technical problem criteria in Sec.~\ref{sec:criteria}. The data creation and review process are outlined in Sec.~\ref{sec:review}, and Sec.~\ref{sec:structure} presents the structure of a \CritPt challenge with an illustrative example.

\subsection{Source and coverage: hand-curated research challenges from the physics community} \label{sec:source}

We source our benchmark data from the problems and reasoning tasks that physicists encounter in real research practice.
Because modern physics is highly specialized, this is only possible through a large-scale collaboration with more than 50 physics researchers across 30 institutions worldwide, including senior Ph.D. students, postdocs, and professors.
Each contributor crafts problems based on their own research expertise, producing a dataset that has both depth of realistic research and diversity in \textit{disciplines}, \textit{topics} and \textit{flavors} in the modern physics landscape.

\captionsetup[table]{skip=6pt}
\begin{table}[h!]
\centering
\renewcommand{\arraystretch}{1.4}
\begin{threeparttable}
\small

\rowcolors{2}{gray!10}{white}

\begin{tabular}{
  >{\raggedright\arraybackslash}m{5.6cm}
  >{\centering\arraybackslash}m{1.4cm}
  >{\centering\arraybackslash}m{1.6cm}
  >{\centering\arraybackslash}m{1.4cm}
  >{\centering\arraybackslash}m{1.6cm}
}
\toprule
\textbf{Research Area} & \textbf{Challenges} & \textbf{\% of Total} & \textbf{Checkpoints} & \textbf{\% of Total} \\
\midrule
Condensed Matter Physics                                 & 25 & 35.2\% & 70 & 36.8\% \\
\mbox{Quantum Information, Science \& Technology}        & 17 & 23.9\% & 43 & 22.6\% \\
Atomic, Molecular \& Optical                             & 14 & 19.7\% & 42 & 22.1\% \\
High Energy Physics                                      & 10 & 14.1\% & 30 & 15.8\% \\
Mathematical Physics                                     & 10 & 14.1\% & 22 & 11.6\% \\
Gravitation, Cosmology \& Astrophysics                   &  9 & 12.7\% & 25 & 13.2\% \\
Statistical Physics \& Thermodynamics                    &  9 & 12.7\% & 24 & 12.6\% \\
Nuclear Physics                                          &  7 &  9.9\% & 19 & 10.0\% \\
Nonlinear Dynamics                                       &  4 &  5.6\% & 12 &  6.3\% \\
Fluid Dynamics                                           &  2 &  2.8\% &  6 &  3.2\% \\
Biophysics                                               &  2 &  2.8\% &  4 &  2.1\% \\
\midrule
\rowcolor{white}
\textbf{Total}                                           & 71 &        & 190 &        \\
\rowcolor{white}
\textbf{Covering Multiple Areas}                         & 33 & 46.5\% &  88 & 46.3\% \\
\bottomrule
\end{tabular}
\caption{The physics research \textit{areas} covered by \CritPt's challenges and checkpoints.}
\label{tab:discipline}
\end{threeparttable}
\end{table}

\begin{figure}[h!]
    \centering
    \includegraphics[width=0.7\linewidth]{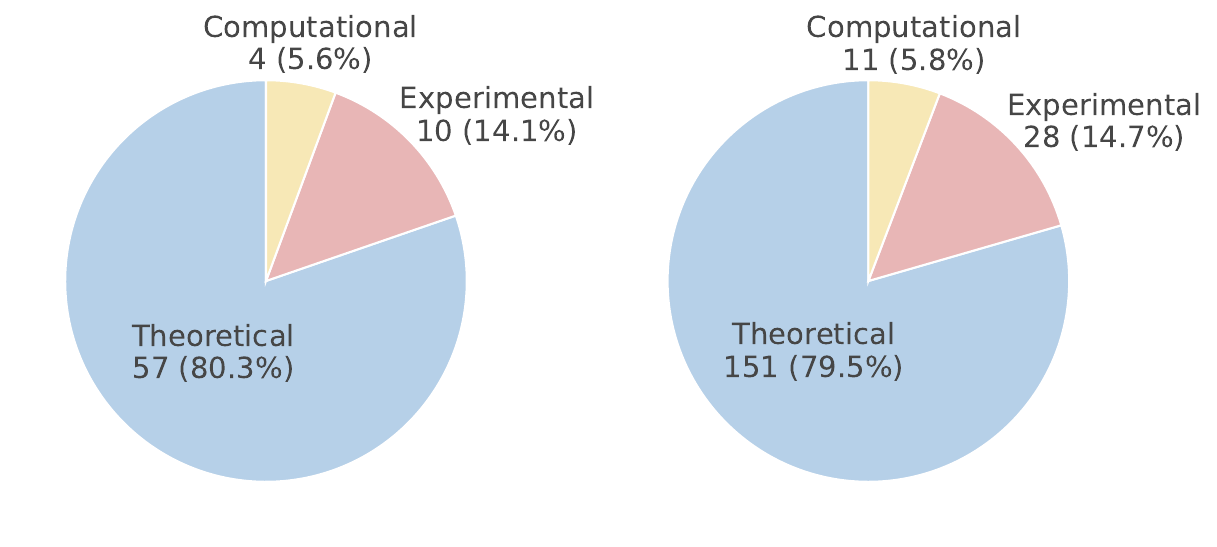}
    \caption{\CritPt's challenges (\textbf{left}) and checkpoints (\textbf{right}) cover three flavors of physics research -- theoretical, experimental, and computational -- encountered by physics researchers.}
    \label{fig:flavor}
\end{figure}

As shown in Table~\ref{tab:discipline}, 71 challenges and 190 checkpoints in \CritPt cover a broad range of modern physics research \textit{disciplines}.\footnote{Our categorization is based on a modified version of the Physics Subject Heading (PhySH) classification scheme created by the American Physical Society~\cite{physh}.}
Among them, 33 challenges and 88 checkpoints cover two or more disciplines, reflecting the growing interdisciplinarity of today's physics research~\cite{sinatra2015century}.
Within these disciplines, we cover \textit{topics} ranging from quantum error correction (relevant to industry) to  
string theory (quest for fundamental particles), 
from cell dynamics (small scales) to black holes (large scales), from nonlinear optics (experimental techniques) to delicate asymptotics of special functions (math tricks). 
Most problems are related to experts' own publications on high-profile physics journals, such as Nature, Science, Physical Review series.
For detailed coverage, see the list of challenges in~\ref{SI:problem_list}. 

\CritPt also covers three major \textit{flavors} of physics research: theoretical, experimental, and computational, as shown in Fig.~\ref{fig:flavor}. 
This three-way grouping reflects how physicists commonly describe themselves.\footnote{This also aligns with the \textit{technique} facet of the PhySH classification~\cite{physh}.} 
Here, we cannot exhaustively cover all types of tasks in each category and instead sample a representative cross-section of reasoning tasks.
For example, an experimental challenge cannot ask an LLM to run equipment directly but can focus on designing or interpreting an experiment under realistic constraints.

Notably, not all problems in \CritPt are about polished research questions that one finds in publications. 
Some problems are inspired by less celebrated but essential aspects of real research, such as failed trials, tedious intermediate calculations, or subtle insights that are rarely documented in papers.
These ``insider'' elements can only be provided by domain experts, and help further differentiate between pattern matching from genuine reasoning.
An LLM capable of reliably solving these challenges would mark a major breakthrough in AI for science. 

\subsection{Benchmark criteria: leakage-resistant and reasoning-focused design}\label{sec:criteria}

With the source and domain coverage of our data established, the next obstacle is to standardize inherently unstructured research-level problems into a benchmark format that provides accurate signals of genuine reasoning, while allowing scalable evaluation.
To guide this process, we define the following technical criteria for constructing \CritPt problems:

\begin{itemize}
\item\textbf{Search-proof but solvable.} 
All problems in \CritPt are newly created and carefully constructed in a way such that their final answers cannot be retrieved through web search.
Meanwhile, they are possible to solve with the publicly known knowledge (i.e., no confidential or private information is needed).
All questions are crafted to be well-posed with unambiguous constraints and verifiable final answers. 
Solving them should demonstrate a deep understanding of the physical scenario, correct application of methods under coherent assumptions, and precise multi-step reasoning and execution.
\CritPt's problems mainly fall into three categories or their combinations: 
(1) modified versions of published results to test out-of-distribution generalization, which simulates the realistic research scenario of a follow-up project based on existing results; 
(2) a non-trivial application of a method to a specific system, which tests understanding and utilizing a method under different physical constraints; 
(3) non-trivial intermediate steps of a calculation not explicitly shown in a paper, which tests the ability to reproduce published results and understand enough to fill in gaps in the context. 
We note that these niche contents are also unlikely to appear in future publications, but mirrors frequently occurring tasks in daily research activity.

\item \textbf{Open-ended Q\&A format with verifiable answers.}
We adopt open-ended question formats with various answer formats, mostly numbers or symbolic expressions.
All the symbols, conventions and physical units are explicitly given in the problems to prepare for canonical grading later.
If an answer expression is too complicated for reliable symbolic manipulation in SymPy or admits too many equivalent forms, we ask models to return a Python function as the answer and evaluate it with test cases.
In rare cases that asking a question with a binary or categorical answer (e.g., ``Yes/No") is essential, we ask a set of related questions and consider the model's solution correct only when all are answered correctly, mirroring how real scientific understanding often requires consistency across multiple angles.

\item\textbf{Guess-resistant construction tailored for physics contents.}
Though search-proof by construction,  physics results, particularly the elegant and memorable ones, often take on some commonly occurring values, such as 0, 1/2 or $\pi$, regardless of the system variation or the derivation path. 
For example, in condensed matter physics, many systems are extremely complex, but universal quantities such as topological numbers are often shared by systems with different microscopic details.
To mitigate the risk of guessing, we carefully choose the physical systems and the quantities to ask that distinguish between correct and incorrect reasoning paths, to ensure that models must follow the intended sequence of physical reasoning to arrive at the correct conclusion.
Each final answer usually contains at least one non-universal quantity in a complicated format, such as floating-point numbers with several-decimal precision, large integers or dimension-dependent symbolic expressions.
\end{itemize}

We note that our design criteria on answer formats are partially inspired by SciCode~\cite{tian2024scicode}, FrontierMath~\cite{glazer2024frontiermath} and TPBench~\cite{chung2025theoretical}.

\subsection{Quality control: iterative development and multi-level expert review}\label{sec:review}

Guided by the benchmark criteria above, each challenge in \CritPt goes through an extensive iterative creation and a multi-stage review process.
Every data contributor, benchmark coordinator, problem reviewer and scientific writer, holds a Ph.D. in physics or is an active physics Ph.D. student engaged in frontier physics research. 

The data collection follows the workflow below:
\begin{enumerate}[itemsep=3pt,topsep=0pt,parsep=0pt,partopsep=0pt]
    \item \textbf{Initial creation:} 
    The coordinators first provide each physics expert annotator with an at least hour-long introduction to LLMs and the benchmark design criteria.
    Experts then create problems based on their research expertise. 
    Each submission includes a solution often more detailed than a typical journal paper, containing step-by-step explanations, algebraic derivations, numerical codes, supporting data, references, and occasionally alternative solutions.
    
    \item \textbf{Iterative revision:}
    The initial draft of each problem undergoes an iterative reviewing process between the expert and coordinators, typically with three or more rounds and up to ten for extremely complex cases.
    AI researchers and physics experts also jointly analyze LLM responses to make sure authentic, domain-relevant reasoning is being tested rather than spurious artifacts, such as formatting issues, ambiguous prompting, or subtle loopholes.
    We avoid cherry-picking based on specific behavior of a particular model to ensure fair comparisons and long-term utility.
    \item \textbf{Expert review:}
    After iterative reviewing, each problem undergoes high-level peer review by researchers in closely related areas, while
    technical derivations and algebraic steps are validated by additional physics experts. 
    Final write-ups are edited by a science writing specialist for clarity and accessibility.
\end{enumerate}
On average, it takes more than 40 hours of expert effort to create one full challenge in \CritPt. 
All \CritPt's contributors and consultants (see Acknowledgment) are given access to leading LLMs and encouraged to experiment with them. 
Experts' first-hand observations of model performance, limitations, and behaviors have deeply transformed our benchmark design throughout a 7-month collaboration.
As a result, \CritPt not only reflects realistic reasoning demands that physicists themselves care about, but is also an effort to provide direct and actionable feedback for AI developers, including those without an advanced physics background.  

\subsection{Structure of a challenge: an example}\label{sec:structure}
We illustrate the structure and design of a full \CritPt challenge with an example, ``Quantum Error Detection'', in this section.

Each \textbf{challenge} is designed as a self-contained, research-style problem at the level of a junior researcher. 
The \textit{setup} section provides all necessary background and context, mimicking how a mentor would define the scope and clarify assumptions when on-boarding the researcher. 
The \textit{challenge question} then poses the central research inquiry within this setup, requiring complex multi-step original reasoning to solve. 

Although most contributing experts agree that current LLMs lag far behind human-level research reasoning in their fields, they also recognize the potential for models to assist with more focused or granular tasks. 
To capture this, each \CritPt challenge is decomposed into 2–4 \textbf{checkpoint} questions to isolate specific reasoning steps. 
While this decomposition naturally reduces complexity, these checkpoints still preserve the depth and reasoning demands of realistic research workflow, going well beyond mechanical steps like formula substitutions. 
They include but are not limited to: filling in intermediate steps in a long derivation, solving simplified versions of the full task (e.g., 1D case before higher dimensions), or analyzing relevant special cases (e.g., behaviour in the high-temperature limit).
If LLMs can reliably handle such tasks, they can save physicists significant effort and speed up scientific discovery. 
We note that the \textit{setup} section is also provided to the models when evaluating checkpoints.

\begin{exampleproblem}
{\large\bfseries Challenge: Quantum Error Detection} \par\vspace{5pt}

\textit{Setup:} 

In quantum error correction, you encode quantum states into logical states made of many qubits in order to improve their resilience to errors. In quantum error detection, you do the same but can only detect the presence of errors and not correct them. In this problem, we will consider a single [[4,2,2]] quantum error detection code, which encodes two logical qubits into four physical qubits, and investigate how robust logical quantum operations in this code are to quantum errors.

Our convention is that the four physical qubits in the [[4,2,2]] code are labelled 0,1,2,3. The two logical qubits are labelled A and B. The stabilizers are $XXXX$ and $ZZZZ$, where $X$ and $Z$ are Pauli matrices. The logical $X$ and $Z$ operators on the two qubits are $X_A = XIXI$, $X_B=XXII$, $Z_A = ZZII$, $Z_B = ZIZI$, up to multiplication by stabilizers.

We will consider different state preparation circuits consisting of controlled not $CNOT_{ij}$ gates, where $CNOT_{ij}$ has control qubit $i$ and target qubit $j$. As a simple model of quantum errors in hardware, we will suppose that each $CNOT_{ij}$ gate in the circuit has a two qubit depolarizing error channel following it that produces one of the 15 non-identity two-qubit Paulis with equal probability $p/15$. The probability $p$ indicates the probability of an error in a single two-qubit gate. We will assess the logical infidelity of certain state preparation protocols as a function of the physical infidelity $p$.

\par\vspace{8pt}
\textit{Challenge question:} 

Suppose that we prepare a logical two-qubit $|00\rangle_{AB}$ state in the [[4,2,2]] code. To do so, we introduce an ancilla qubit, qubit 4, and use the following state preparation circuit:
\begin{equation*}
    M_4 (CNOT_{04}) (CNOT_{34}) (CNOT_{23}) (CNOT_{10}) (CNOT_{12}) (H_1) 
\end{equation*}

Note that this equation is written in matrix multiplication order, while the quantum operations in the circuit occur in the reverse order (from right-to-left in the above equation).  $H$ is a single-qubit Hadamard gate and $M$ is a single-qubit measurement. The ancilla is used to detect errors in the state preparation circuit and makes the circuit fault-tolerant. If the ancilla measurement is $|0\rangle$ ($|1\rangle$), the state preparation succeeds (fails).

What is the logical state fidelity of the final 2-qubit logical state at the end of the circuit as a function of two-qubit gate error rate $p$, assuming the state is post-selected on all detectable errors in the code and on the ancilla qubit measuring $|0\rangle$?

\par\vspace{3pt}
\textit{Answer:} 
\begin{equation*}
    F_{\rm{logical}}=1-\frac{\frac{16}{25} p^2 - \frac{128}{125} p^3 + \frac{2048}{3375} p^4 - \frac{32768}{253125} p^5}{1 - \frac{68}{15} p + \frac{704}{75} p^2 - \frac{32768}{3375} p^3 + \frac{
 253952}{50625} p^4 - \frac{262144}{253125} p^5}
\end{equation*}

\par\vspace{8pt}
\textbf{\large Checkpoints}
\begin{hanglist}

\item \textit{Checkpoint 1:} 

Suppose that we wish to prepare a logical two-qubit GHZ state  $(|00\rangle_{AB}+|11\rangle_{AB})/\sqrt{2}$ in the [[4,2,2]] code. To do so, we use the following state preparation circuit:
\begin{equation*}
 (CNOT_{03}) (H_0) (CNOT_{21}) (H_2). 
\end{equation*}

Note that this equation is written in matrix multiplication order, while the quantum operations in the circuit occur in the reverse order (from right-to-left in the above equation). $H$ is a single-qubit Hadamard gate.

What is the physical state fidelity of the final physical 4-qubit state at the end of the circuit as a function of the two-qubit gate error rate $p$?

\par\vspace{3pt}
\textit{Answer:} 
\begin{equation*}
    F_{\rm{physical}}=\left(1-\frac{12}{15}p\right)^2
\end{equation*}

\item\textit{Checkpoint 2:}

Suppose that we wish to prepare a logical two-qubit GHZ state  $(|00\rangle_{AB}+|11\rangle_{AB})/\sqrt{2}$ in the [[4,2,2]] code. To do so, we use the following state preparation circuit:
\begin{equation*}
 (CNOT_{03}) (H_0) (CNOT_{21}) (H_2). 
\end{equation*}

Note that this equation is written in matrix multiplication order, while the quantum operations in the circuit occur in the reverse order (from right-to-left in the above equation). $H$ is a single-qubit Hadamard gate.

What is the logical state fidelity of the final 2-qubit logical state at the end of the circuit as a function of the two-qubit gate error rate $p$, assuming the state is post-selected on all detectable errors in the code?

\par\vspace{3pt}
\textit{Answer:} 
\begin{equation*}
    F_{\rm{logical}}=1 - \frac{\frac{16}{75}p^2}{1-\frac{8}{5}p + \frac{64}{75}p^2}
\end{equation*}

\item\textit{Checkpoint 3:}

Suppose that we prepare a logical two-qubit $|00\rangle_{AB}$ state in the [[4,2,2]] code. To do so, we introduce an ancilla qubit, qubit 4, and use the following state preparation circuit:
\begin{equation*}
    M_4 (CNOT_{04}) (CNOT_{34}) (CNOT_{23}) (CNOT_{10}) (CNOT_{12}) (H_1)
\end{equation*}
Note that this equation is written in matrix multiplication order, while the quantum operations in the circuit occur in the reverse order (from right-to-left in the above equation). $H$ is a single-qubit Hadamard gate and $M$ is a single-qubit measurement. The ancilla is used to detect errors in the state preparation circuit and makes the circuit fault-tolerant. If the ancilla measurement is $|0\rangle$ ($|1\rangle$), the state preparation succeeds (fails).

What is the logical state fidelity of the final 2-qubit logical state at the end of the circuit as a function of two-qubit gate error rate $p$, assuming the state is post-selected on all detectable errors in the code and on the ancilla qubit measuring $|0\rangle$?

\par\vspace{3pt}
\textit{Answer:} 
\begin{equation*}
    F_{\rm{logical}}=1-\frac{\frac{16}{25} p^2 - \frac{128}{125} p^3 + \frac{2048}{3375} p^4 - \frac{32768}{253125} p^5}{1 - \frac{68}{15} p + \frac{704}{75} p^2 - \frac{32768}{3375} p^3 + \frac{
 253952}{50625} p^4 - \frac{262144}{253125} p^5}
\end{equation*}

\end{hanglist}

\end{exampleproblem}

See design ideas behind the example challenge in \ref{SI:example_idea} and detailed solutions on this \href{https://critpt.com/example}{web page}.

%% file: text/03-eval.tex
\section{Evaluation pipeline}\label{sec:eval}

We implement an automated evaluation framework combining a structured two-step generation protocol (Sec.~\ref{sec:eval_generation}) with a robust, canonical grading system (Sec.~\ref{sec:eval_grading}). This setup ensures both faithful assessment of reasoning quality and rigorous, scalable verification across diverse output formats.
Our evaluation pipeline and grading server for \CritPt challenges are hosted online for future testing and public access (Sec.~\ref{sec:eval_server}).

\subsection{Two-step answer generation from models}
\label{sec:eval_generation}
\begin{figure}[h!]
    \centering
    \includegraphics[width=1.02\linewidth]{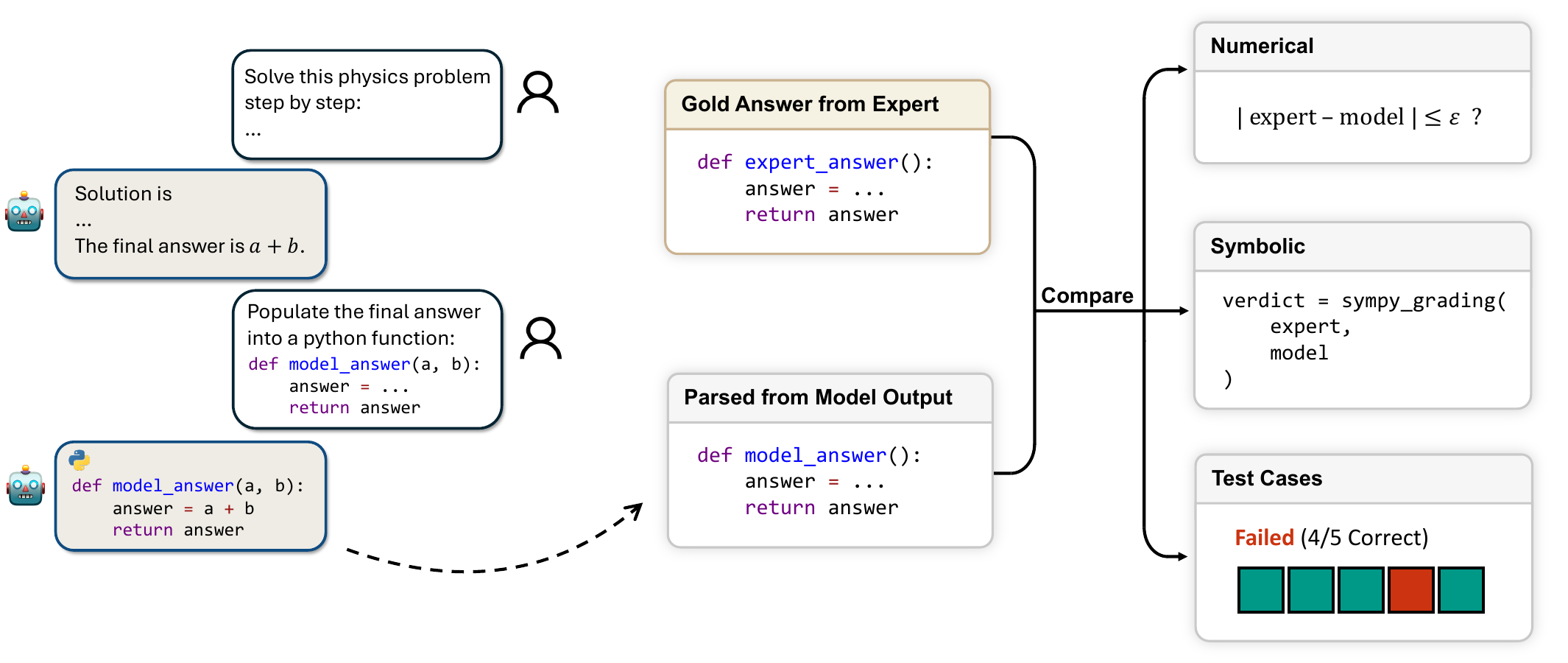}
    \caption{A schematic overview of the two-step generation process and the grading system. 
    \textbf{Left}: The two-step generation protocol separates problem-solving (first round) from answer formatting (second round). 
    \textbf{Right}: The automated grading system compares the model output against the gold answer from experts using scripts customized according to the expected answer format.}
    \label{fig:parse_grade}
\end{figure}

To disentangle the reasoning process from answer formatting, we adopt a two-step generation strategy, sketched in Fig.~\ref{fig:parse_grade} (Left):
\begin{itemize}[itemsep=2pt,topsep=0pt,parsep=1pt,partopsep=0pt]
\item In the first step, the model is prompted to generate a complete solution using free-form natural language and mathematical derivations (see~\ref{SI:prompt} for the system prompt). 
This allows the model to reason without constraints imposed by output formatting templates.
\item In the second step, we guide the model to extract and standardize its final answer into a designated code block template we provide (see~\ref{SI:prompt} for the parsing prompt and the example template). 
This template enforces a canonical structure suitable for parsing and grading. 
\end{itemize}
By separating solution reasoning from formatting, we avoid premature conversions (e.g., via \texttt{SymPy}) that may distort intermediate steps, and reduce parsing errors from inconsistent model output styles.

\subsection{Auto-grading system}
\label{sec:eval_grading}
We extract the final answer by parsing the Python code block generated in the second step, which contains an answer function. 
Parsing is designed to tolerate surrounding text while enforcing strict formatting within the code block. To ensure safety and compatibility, only a limited set of libraries (e.g., \texttt{math}, \texttt{numpy}, \texttt{sympy}, \texttt{scipy}) is allowed, and potentially unsafe operations are filtered.

Next, we compare the extracted model's answer function with the gold answer provided by physics expert through an automated grading system. 
Our automated grading system supports three primary answer types: numerical values, symbolic expressions, and Python functions, each with their own evaluation logic, sketched in Fig.~\ref{fig:parse_grade} (Right):

\begin{itemize}[itemsep=2pt,topsep=0pt,parsep=1pt,partopsep=0pt]

\item \textbf{Numerical values}: 
We assess correctness against gold answers using expert-provided tolerance ranges that generously account for the physical or numerical sensitivity of each problem. 
For exact results, a default precision of 12 significant digits is enforced to accommodate potential floating-point errors introduced by conversion to Python code.

\item \textbf{Symbolic expressions}:
For problems involving symbolic algebra, we use a hierarchical grading script based on \texttt{SymPy}, implemented underneath the \texttt{sympy\_grading()} function in Fig.~\ref{fig:parse_grade}. 
It starts with built-in equivalence checking and algebraic simplification, extended with custom routines tailored to the structure of each expression when standard simplification proves insufficient. 
We also accommodate issues that can be easily overlooked such as math object type conflicts in \texttt{SymPy}.

\item  \textbf{Python codes}:
When models return executable functions (e.g., for parametric solutions), we grade the model answer using curated test cases selected by physics experts~\cite{tian2024scicode}. These cases are chosen to probe physically meaningful parameter ranges and edge conditions.
\end{itemize}

For composite answers (e.g., tuples or dictionaries of results), we apply element-wise grading, and an answer is considered correct only if all components match the expert answer. 

To ensure runtime safety and isolation, each answer is executed in a sand-boxed environment with enforced resource limits: 30-second wall-clock timeout and bounded memory usage (or the resource limits specified by the expert for computationally heavy tasks; whichever is higher between the default and expert's specification).
This prevents pathological behaviors such as infinite loops or excessive allocation created by models, while still supporting computation-intensive problems.

Though labor-intensive to build upfront, this pipeline enables scalable, high-fidelity evaluation for diverse, complex output formats in advanced physics.

\subsection{Public evaluation pipeline for \CritPt challenges}
\label{sec:eval_server}
While the answers to 70 \CritPt test challenges are kept private to mitigate data leakage, we provide a public evaluation pipeline with an online grading server.\footnote{Detailed documentation on the public evaluation pipeline: \href{https://github.com/CritPt-Benchmark/CritPt}{https://github.com/CritPt-Benchmark/CritPt}}
Model developers can test their own models using the same streamlined generation and grading scripts used for our reported results in the next section, to obtain directly comparable results.
The grading server returns an accuracy as the fraction of correctly solved challenges over the full test set.
To discourage selective querying and benchmark hacking, the server only accepts complete batches of answers for all 70 challenges and is rate-limited to 10 submissions per account per 24-hour period. 
These design choices are intended to strike a balance between public accessibility and resistance to data leakage.

This secure public pipeline also allows collaboration with external partners without compromising private data. Notably, Artificial Analysis, an AI evaluation company, has independently validated our results on the \CritPt challenges using this pipeline~\cite{aa_critpt}. 
They are continually evaluating frontier models at a scale and frequency typically beyond the resources of an academic laboratory, providing the physics community with timely insights into the capabilities of the latest models.\footnote{We also thank Artificial Analysis for supporting the grading server, which enables large-scale public access.}



%% file: text/04-result.tex
\section{Results}\label{sec:res}

In this section, we evaluate 10 state-of-art models (configuration in Table~\ref{tab:model-list}) to directly answer the three lines of inquiry in the introduction. 
Evaluation is performed \textit{independently} on two levels: the full \textbf{challenges} (Sec.~\ref{sec:res_challenge}) and finer-grained \textbf{checkpoints} (Sec.~\ref{sec:res_checkpoint}). 
In these two sections, we report \textbf{average accuracy} over five runs as our primary metric to reduce high stochasticity in model behavior.
Next in Sec.~\ref{sec:res_conf}, we adopt a stricter metric, \textbf{consistently solved rate}, to probe reliability of model performance.
Table~\ref{tab:summary_metrics} is a brief summary of these aggregated accuracy statistics.
Detailed analysis of model responses beyond aggregated statistics is discussed in Sec.~\ref{sec:res_analysis}.

\captionsetup[table]{skip=6pt}
\begin{table}[h!]
\centering
\renewcommand{\arraystretch}{1.5}
\begin{threeparttable}
\small
\begin{tabular}{>{\raggedright}p{3.3 cm} p{3.5cm} p{3.5 cm} p{1.3 cm}}
\toprule
\textbf{Model Name} & \textbf{Reasoning Effort} & \textbf{Tool Use} & \textbf{Company} \\
\midrule
GPT-5 (high) & \texttt{high} & / & OpenAI \\
GPT-5 (high, code) & \texttt{high}  & code interpreter & OpenAI \\
GPT-5 (high, code \& web) & \texttt{high}  & code interpreter, web search & OpenAI \\
o3 (high) & \texttt{high} & / & OpenAI\\
o4-mini (high) & \texttt{high} & / &  OpenAI \\
Gemini 2.5 Pro & \texttt{reasoning\_tokens=27000} & / &  Google \\
Gemini 2.5 Flash & \texttt{Default} & / & Google \\
DeepSeek R1 & \texttt{Default} & / &  DeepSeek \\
Claude Opus 4 & \texttt{reasoning\_tokens=27000} & / & Anthropic  \\
\rowcolor{gray!10}
GPT-5 (minimal) & \texttt{minimal}\tnote{*} & / & OpenAI  \\
\rowcolor{gray!10}
Llama-4 Maverick & \texttt{/} & / & Meta \\
\rowcolor{gray!10}
GPT-4o & \texttt{/} & / & OpenAI  \\
\bottomrule
\end{tabular}
\begin{tablenotes}
\scriptsize
\item[*] GPT-5 (minimal) sets reasoning\_effort = minimal, which means zero reasoning tokens used.
\end{tablenotes}
\end{threeparttable}
\caption{Models and their API configurations used in our evaluation. Both reasoning-oriented models (white background) and  general-purpose chat models (grey background) are included. More details in~\ref{SI:eval_method}.}
\label{tab:model-list}
\end{table}

\definecolor{LightBlue}{RGB}{199,216,235}

\newcommand{\Vmax}{25.6}

\newcommand{\gradcell}[1]{%
  \pgfmathparse{min(100, round(100*(#1)/\Vmax))}%
  \edef\temp{\noexpand\cellcolor{LightBlue!\pgfmathresult}}%
  \temp#1%
}

\begin{table}[h!]
\centering
\renewcommand{\arraystretch}{1.5}
\begin{threeparttable}
\small
\begin{tabular}{
  >{\raggedright\arraybackslash}p{3.3cm}
  >{\centering\arraybackslash}p{1.4cm}
  >{\centering\arraybackslash}p{1.4cm}
  >{\centering\arraybackslash}p{1.4cm}
  >{\centering\arraybackslash}p{1.4cm}
  >{\centering\arraybackslash}p{1.4cm}
  >{\centering\arraybackslash}p{1.4cm}
  }
\toprule
\multirow{3}{*}[-1.5ex]{\textbf{Model Name}}
  & \multicolumn{3}{c}{\textbf{Average Accuracy (\%)}}
  & \multicolumn{3}{c}{\textbf{Consistently Solved Rate (\%)}} \\
\cmidrule(lr){2-4}\cmidrule(lr){5-7}
& \multirow{2}{*}{\textbf{Challenge} }
  & \multicolumn{2}{c}{\textbf{Checkpoint}}
  & \multirow{2}{*}{\textbf{Challenge}}
  & \multicolumn{2}{c}{\textbf{Checkpoint}} \\
\cmidrule(lr){3-4}\cmidrule(lr){6-7}
& & \textbf{w/o expert answer} & \textbf{w/ expert answer}
  & & \textbf{w/o expert answer} & \textbf{w/ expert answer} \\
\midrule
GPT-5 (high, code \& web) & \gradcell{12.6}              & \gradcell{21.4}              & \gradcell{24.5}              & \gradcell{10.0}             & \gradcell{14.4}             & \gradcell{17.6}             \\
GPT-5 (high, code)        & \gradcell{10.6}              & \gradcell{20.0}              & \gradcell{23.9}              & \phantom{0}\gradcell{8.6}   & \gradcell{15.5}             & \gradcell{17.6}             \\
GPT-5 (high)              & \phantom{0}\gradcell{5.7}    & \gradcell{15.3}              & \gradcell{20.0}              & \phantom{0}\gradcell{4.3}   & \gradcell{10.7}             & \gradcell{15.5}             \\
Gemini-2.5 Pro            & \phantom{0}\gradcell{2.0}    & \phantom{0}\gradcell{8.1}    & \phantom{0}\gradcell{9.1}    & \phantom{0}\gradcell{0.0}   & \phantom{0}\gradcell{4.3}   & \phantom{0}\gradcell{3.7}   \\
o3 (high)                 & \phantom{0}\gradcell{1.4}    & \phantom{0}\gradcell{7.4}    & \gradcell{10.6}              & \phantom{0}\gradcell{0.0}   & \phantom{0}\gradcell{3.7}   & \phantom{0}\gradcell{5.9}   \\
DeepSeek R1               & \phantom{0}\gradcell{1.1}    & \phantom{0}\gradcell{5.1}    & \phantom{0}\gradcell{6.6}    & \phantom{0}\gradcell{0.0}   & \phantom{0}\gradcell{2.1}   & \phantom{0}\gradcell{3.2}   \\
Gemini-2.5 Flash          & \phantom{0}\gradcell{1.1}    & \phantom{0}\gradcell{4.1}    & \phantom{0}\gradcell{4.6}    & \phantom{0}\gradcell{0.0}   & \phantom{0}\gradcell{2.1}   & \phantom{0}\gradcell{2.1}   \\
o4-mini (high)            & \phantom{0}\gradcell{0.6}    & \phantom{0}\gradcell{5.6}    & \phantom{0}\gradcell{7.1}    & \phantom{0}\gradcell{0.0}   & \phantom{0}\gradcell{2.7}   & \phantom{0}\gradcell{2.1}   \\
Claude Opus 4             & \phantom{0}\gradcell{0.3}    & \phantom{0}\gradcell{2.6}    & \phantom{0}\gradcell{3.0}    & \phantom{0}\gradcell{0.0}   & \phantom{0}\gradcell{0.5}   & \phantom{0}\gradcell{1.6}   \\
GPT-5 (minimal)           & \phantom{0}\gradcell{0.0}    & \phantom{0}\gradcell{2.6}    & \phantom{0}\gradcell{3.7}    & \phantom{0}\gradcell{0.0}   & \phantom{0}\gradcell{1.6}   & \phantom{0}\gradcell{2.1}   \\
Llama-4 Maverick          & \phantom{0}\gradcell{0.0}    & \phantom{0}\gradcell{2.2}    & \phantom{0}\gradcell{2.1}    & \phantom{0}\gradcell{0.0}   & \phantom{0}\gradcell{1.1}   & \phantom{0}\gradcell{0.5}   \\
GPT-4o                    & \phantom{0}\gradcell{0.0}    & \phantom{0}\gradcell{1.7}    & \phantom{0}\gradcell{1.2}    & \phantom{0}\gradcell{0.0}   & \phantom{0}\gradcell{0.0}   & \phantom{0}\gradcell{0.0}   \\
\bottomrule
\end{tabular}
\end{threeparttable}
\caption{A high-level summary of \CritPt evaluation results. }
\label{tab:summary_metrics}
\end{table}

\subsection{Challenge-level evaluation: can LLMs solve unseen research problems?}
\label{sec:res_challenge}

\begin{figure}[h!]
    \centering
    \includegraphics[width=1.0\linewidth]{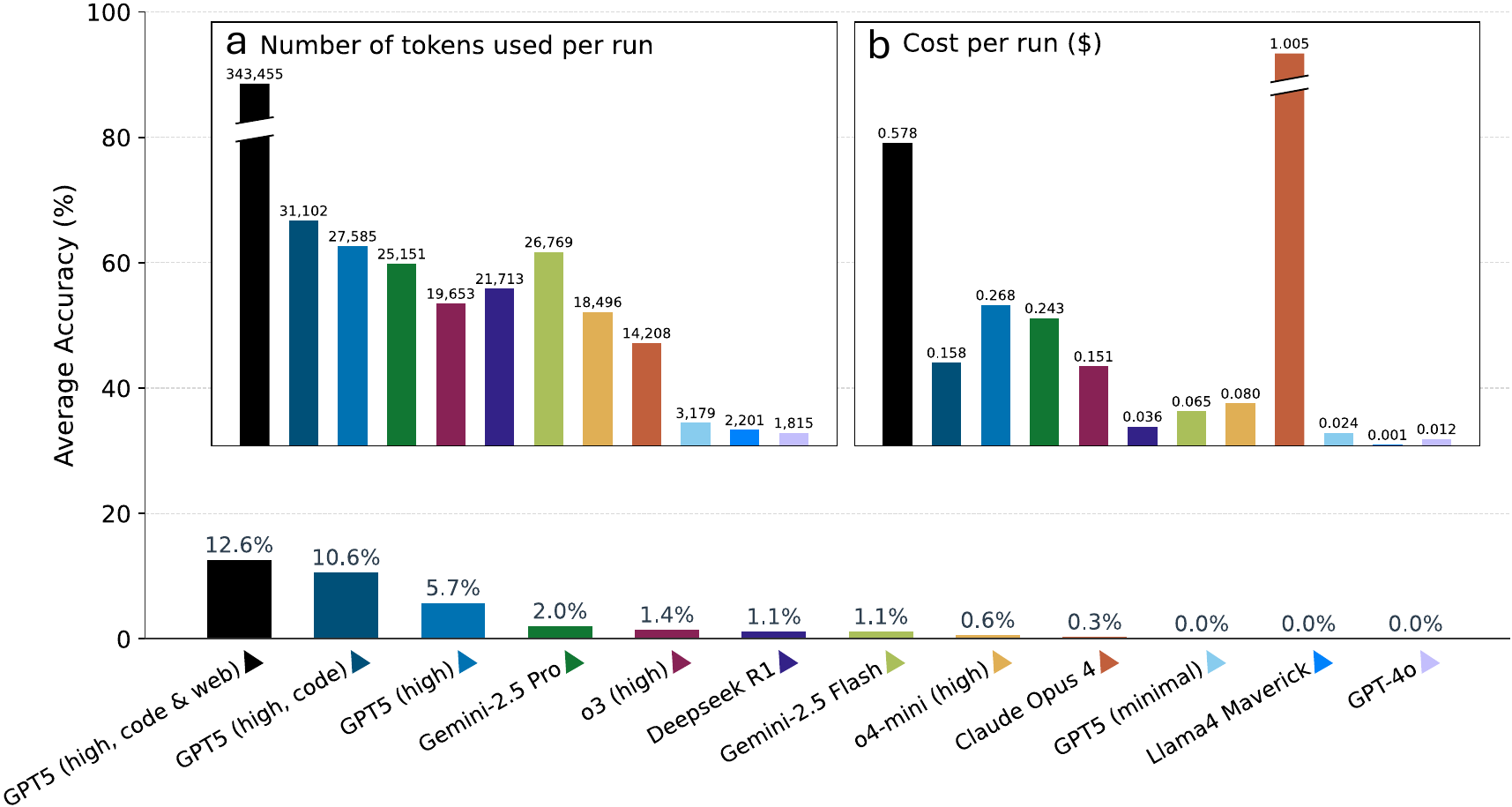}
    \caption{A comparison of 10 models' performance on 70 test \CritPt challenges. Each model is tested on every challenge in five independent runs.
    \textbf{Main plot}: the average accuracy over all runs and all challenges for each model. 
    \textbf{Inset a}: the average number of reasoning tokens used per run for each model.
    \textbf{Inset b}: the average cost (USD) per run, calculated from  token usage and API pricing for each model (\ref{SI:API_stat}).
    }
    \label{fig:perf_challenge}
\end{figure}

We first assess model performance on \CritPt challenges without intermediate supervision, testing their end-to-end complex reasoning in unseen full-scale research problems.  
As shown in Fig.~\ref{fig:perf_challenge}, all models score low in terms of average accuracy of five independent runs across 70 challenges in \CritPt test set. 
Among base models (no external tools), GPT-5 (high), achieves only 5.7\% as the best result, with all other models below or equal to 2\%.

Tool use can offer modest but meaningful gains. 
With access to the code interpreter, GPT-5 (high, code) improves to 10.6\%, a notable relative improvement from the base model, consistent with the importance of numerical tools in modern physics research.
Adding web search brings only a marginal increase to 12.6\%, confirming that \textit{CritPt}’s search-proof design effectively resists short-cutting via retrieval and emphasizes genuine reasoning.

As expected, all the general-purpose chat models score zero, while reasoning-oriented models start making progress, however small, suggesting some emerging capability from explicitly structured reasoning processes.
Meanwhile, with long reasoning chains and more verbose outputs, these reasoning-oriented models consume significantly more resources. 
As shown in \textbf{Inset a} of Fig.~\ref{fig:perf_challenge}, 
these models consume much more tokens per run than general-purpose models. 
Note that GPT-5 (high, code \& web) uses an order of magnitude more tokens than other reasoning models, primarily caused by large amounts of web-retrieved contents counted as extra input tokens, not necessarily indicating deeper thinking.
For a more detailed token consumption breakdown by type and number, please see~\ref{SI:API_stat}.
Overall, this shows that all reasoning-oriented models engage deeply with the these challenges, which are indeed difficult even for LLMs with long context windows and extensive computational resources.

Inference cost can be another critical constraint for scaling LLM usage in research. 
\textbf{Inset b} of Fig.~\ref{fig:perf_challenge} reports the average cost per run on \CritPt challenges, determined by total token usage and each vendor’s standard API pricing (see ~\ref{SI:API_stat} for details).
The large token consumption by advanced models naturally drives up cost, but the gain in performance is often disproportionate.
This highlights the need for \textit{efficient} reasoning: performance gains should not rely solely on extended context length, especially in scientific domains where the solution space is sparse and brute-force exploration is ineffective. 
Pricing policies further impact cost, as high-performing models are typically commercial.
For example, DeepSeek R1 is relatively affordable thanks to its low per-token rates, whereas Claude Opus 4 is the most expensive due to its premium pricing.

In summary, current LLMs are far away from being able to solve unseen physics research problem at a junior-researcher level.
While coding tools help make a small initial step, they are not sufficient to overcome fundamental reasoning bottlenecks.
Notably, GPT-5 was released near the very end of our data collection cycle, but only shows limited progress over its predecessor, o3 (high), on \CritPt challenges. 
This suggests that a realistic benchmark like \CritPt has the potential to resist new generations of models for a substantial amount of time.

\subsection{Checkpoint-level evaluation: smaller tasks that LLMs can assist today?}
\label{sec:res_checkpoint}

To better differentiate current model capabilities and isolate failure modes, each \CritPt challenge is decomposed into 2–4 checkpoint questions. 
These checkpoints are evaluated sequentially in a multi-turn conversation format, simulating how researchers might naturally interact with an assistant system during problem solving.

\begin{figure}[h!]
    \centering
    \includegraphics[width=0.9\linewidth]{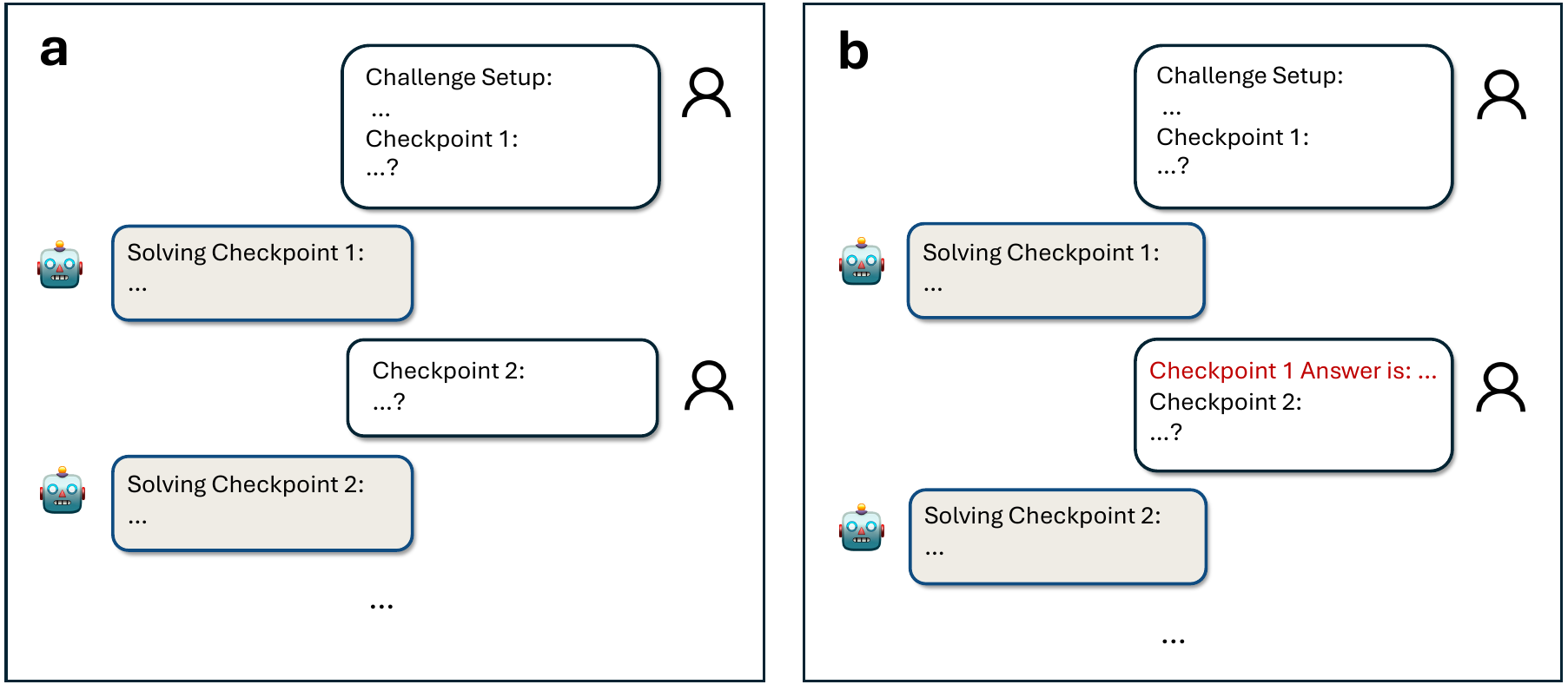}
    \caption{Schematic of the two experimental setups for evaluating sequential checkpoints within a multi-turn conversation. 
    (\textbf{a}) Self-carryover without expert answer: The model's own output from the previous checkpoint is used as context for the next one.
    (\textbf{b}) Oracle carryover with expert answers: The correct answer (shown in red) to the previous checkpoint is provided before the model attempts the next checkpoint.}
    \label{fig:checkpoint_setups}
\end{figure}

We use two evaluation setups for checkpoints, illustrated in Fig.~\ref{fig:checkpoint_setups}, to test reasoning effectiveness in different research scenarios:

\begin{itemize}
\item \textbf{Self-carryover (without expert answer):} 
The model proceeds sequentially, using only its own previous outputs (Fig.~\ref{fig:checkpoint_setups}a). 
This setting captures a realistic scenario where the overall problem is decomposable, but intermediate results remain uncertain and errors can propagate and compound.

\item \textbf{Oracle carryover (with expert answers):} 
The model also proceeds sequentially, but is given ground-truth answer to the prior checkpoint before attempting the next (Fig.~\ref{fig:checkpoint_setups}b). 
This setup intends to test isolated local tasks by removing upstream error effects, or assess whether a model can effectively use answers to relevant tasks as hints.
\end{itemize}

\begin{figure}[h!]
    \centering
    \includegraphics[width=1.0\linewidth]{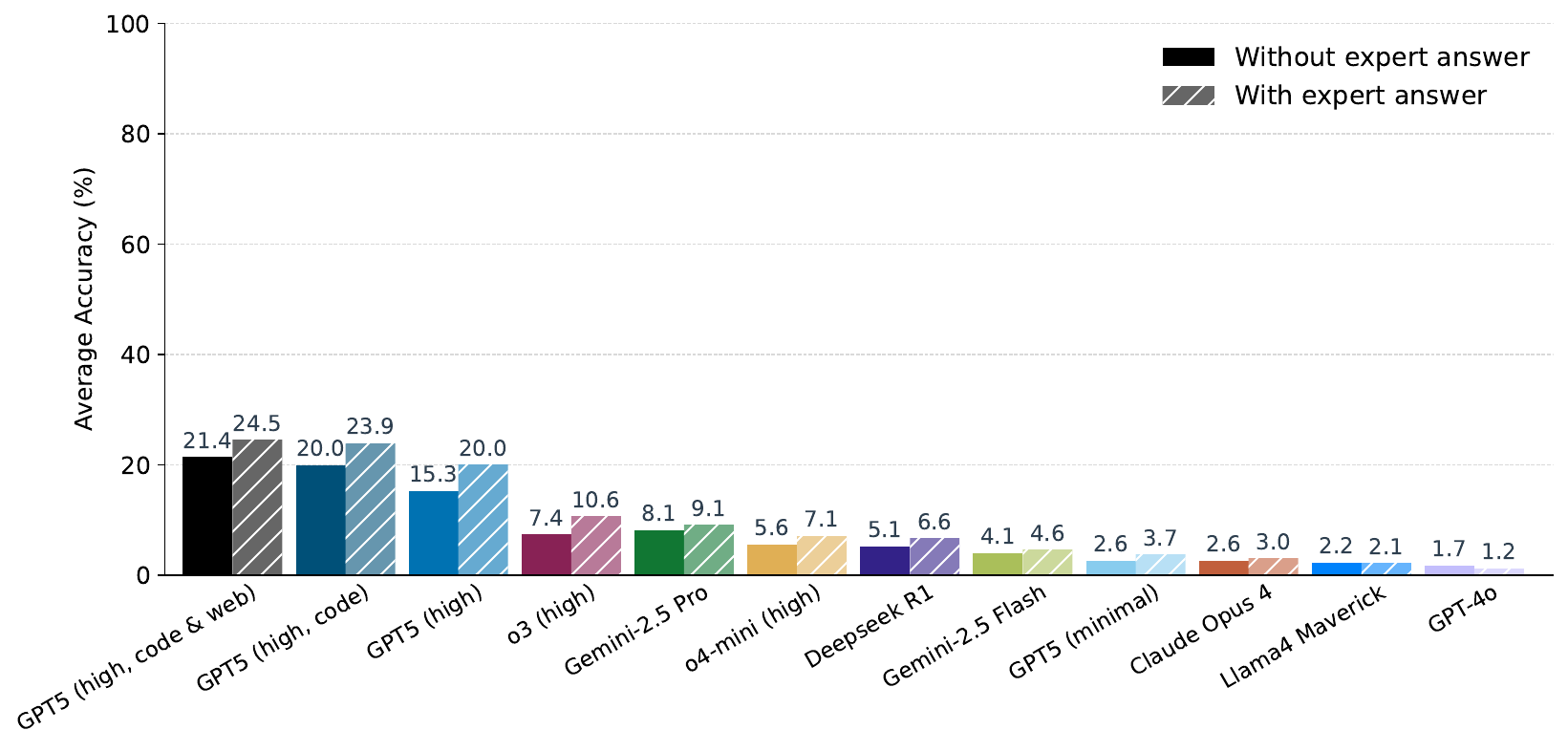}
    \caption{A comparison of 10 models' performance on the 187 test \CritPt checkpoints.
    The average accuracy is aggregated over all runs and all checkpoints, in two setups respectively.
    Solid bar reports the self-carryover without expert answer, while the hatched pattern reports oracle carryover with expert answers.}
    \label{fig:perf_checkpoint}
\end{figure}

As shown in Fig.~\ref{fig:perf_checkpoint}, current LLMs show early promise in assisting with checkpoints, more localized or well-scoped tasks compared to challenges.
GPT-5 (high) again outperforms all other models. 
In the self-carryover without expert answer (solid bar),  GPT-5 (high) reaches 15.3\%.
This improves to 20.0\% with code interpreter and reaches 21.4\% when also equipped with web search.
The next best performers are Gemini 2.5 Pro (8.1\%) and o3 (high) (7.4\%), followed by o4-mini (high) (5.6\%), DeepSeek R1 (5.1\%) and Gemini 2.5 Flash (4.1\%).
The remaining models fall below 3\%, where small absolute differences may reflect statistical noise rather than meaningful performance gaps, so the ordering should be interpreted with caution.

Almost all models benefit from expert answer injection to earlier checkpoints. 
The largest gains are observed for GPT-5 (high) family and o3 (high), which improve by more than 3\% in the oracle-carryover setting (hatched bar), indicating the ability to leverage correct intermediate results to improve downstream reasoning.

Though still difficult, \CritPt checkpoints seems to fall within the improving front for next generation leading models.
This aligns with the feedback from our physics experts, particularly theorists, who have begun cautiously experimenting with LLMs in their daily research, and occasionally find them correct for small and well-defined reasoning tasks.
However, LLMs are not fully correct most of the time, so experts must dedicate considerable time to verifying the convoluted yet plausible outputs. 
This process can sometimes exceed the time required to solve the problem independently.
These observations motivate the next section, an attempt to assess reliability of LLM performance with a stricter evaluation metric.

\subsection{Reliability metric: can we trust LLM outputs?}
\label{sec:res_conf}
In the regime of highly complex and open-ended problems, a robust and trustworthy reasoning process is particularly important. 
Subtle mistakes from hallucination can be hard to identify and can mislead users who are learning new things and lack expert-level judgment. 
As a prerequisite check of the models' reliability, we introduce a stricter performance metric: a problem is deemed as \textbf{consistently solved} only if at least four out of five runs give correct final answers.\footnote{Caveats: this is a heuristic reliability check based on the number of runs we have given our resource constraints. It can also be viewed as a 4/5 super majority vote (or pass@4/5). We are \textit{by no means} claiming statistical sufficiency from five samples nor that 80\% accuracy implies high reliability. We welcome sponsorship or third-party host to run more tests.}

\begin{figure}[h!]
    \centering
    \includegraphics[width=1.0\linewidth]{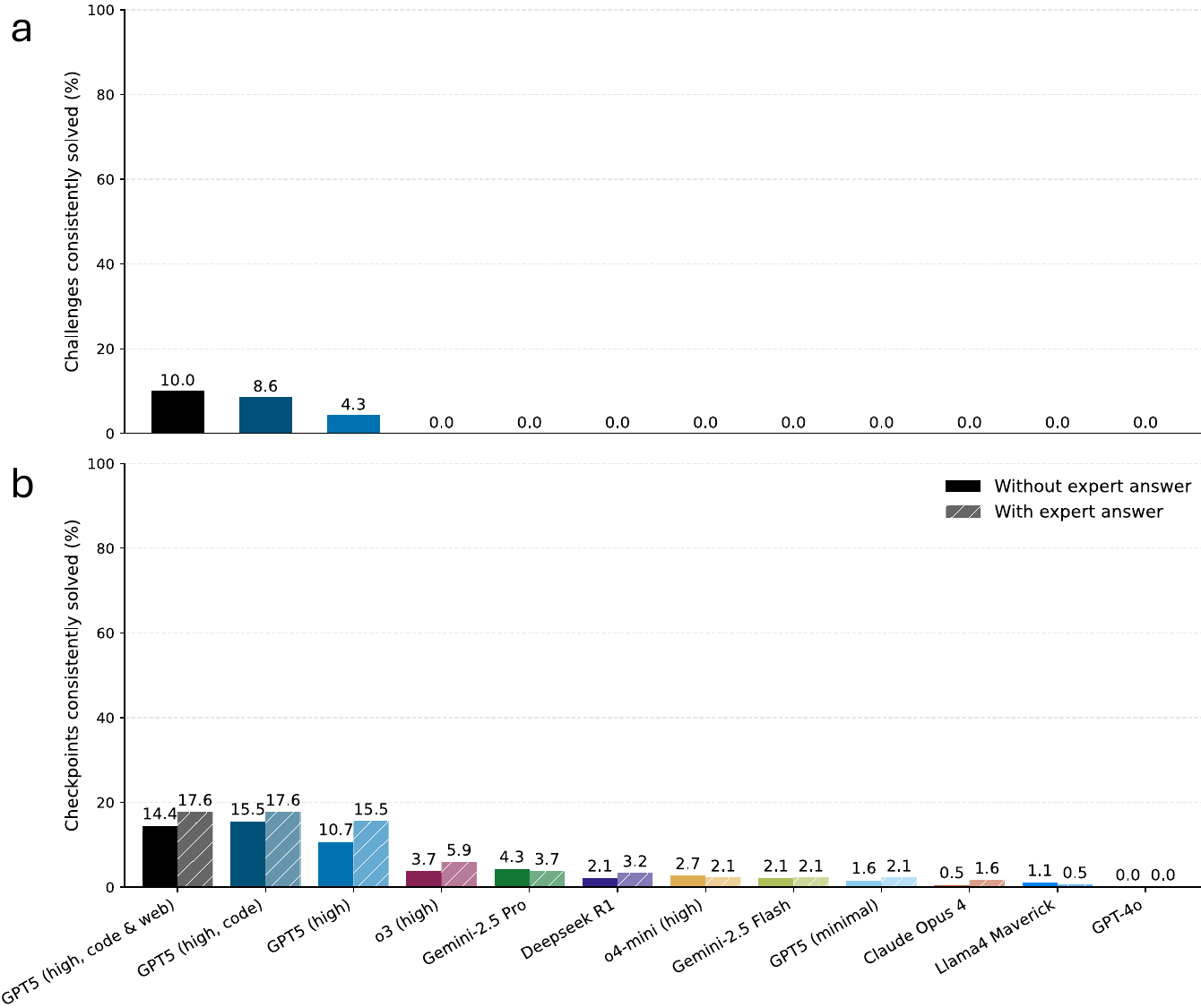}
    \caption{Percentage of \CritPt problems consistently solved by models.
    A problem is considered consistently solved if at least four out of five independent runs yield the correct final answer.
    (\textbf{a}) Percentage of challenges consistently solved. 
    (\textbf{b}) Percentage of checkpoints consistently solved.
    }
    \label{fig:perf_consistency}
\end{figure}

Applying this criterion leads to a sharp drop in performance across all models, suggesting high stochasticity in model behavior in complex physics research context.

At the full challenge level (Fig.~\ref{fig:perf_consistency}a), only GPT-5 (high) is able to solve any problems consistently~\cite{kalai2025language}. 
The base model scores only 4.3\% (3 out of 70 challenges).
With tool use, this improves to 8.6\% (code) and 10.0\% (code \& web). 
All other reasoning-oriented models drop to zero under this stricter measure.

At the checkpoint level (Fig.~\ref{fig:perf_consistency}b), leading models are able to consistently solve a very limited number. 
Interestingly, GPT-5 (high, code) outperforms GPT-5 (high, code \& web) with a small margin, further supporting the search-proof design of \CritPt even at the checkpoint level. 

These results imply that current LLMs cannot yet be trusted to reason consistently in high-stakes research contexts. 
Improving reasoning reliability through better uncertainty calibration, more powerful external validations or other advances remains an open challenge.
Meanwhile, the message to the physics community is clear: while LLMs may be useful for exploring or prototyping small subtasks, caution and expert oversight may be necessary in advanced research contexts.

\subsection{Detailed analysis of full model responses}
\label{sec:res_analysis}

Beyond aggregated accuracy metrics, we further analyze model behavior at the level of individual challenges and walk through the full model responses (including reasoning traces when available) with physics experts, which helps surface more qualitative insights not captured by final answer correctness alone. 

To support an efficient review process with experts and overcome the limitations of traditional LLM infrastructure, we develop an interactive visualization platform.
This tool allows experts to quickly browse model responses, systematically compare performance across tasks and model families, and identify error patterns or interesting behaviors.
Table~\ref{tab:qec_complete} illustrates the visualization of the example challenge, where we can immediately observe an unexpected behavior: adding web search tool on top of coding for GPT-5 (high) actually degrades the model performance for this particular challenge. 
The full interactive demo including all model responses is available at \href{https://critpt.com/example}{critpt.com/example}.

\begin{table}[h!]
\centering
\renewcommand{\arraystretch}{1.6}
\begin{threeparttable}
\small
\begin{tabular}{
  >{\raggedright\arraybackslash}m{3.3cm}
  >{\centering\arraybackslash}m{1.2cm}
  >{\centering\arraybackslash}m{1.5cm}
  >{\centering\arraybackslash}m{1.4cm}
  >{\centering\arraybackslash}m{1.4cm}
  >{\centering\arraybackslash}m{1.4cm}
  >{\centering\arraybackslash}m{1.4cm}
}
\toprule
\multirow{2}{*}{\textbf{Model Name}}
  & \multirow{2}{*}{\textbf{Challenge}}
  & \multirow{2}{*}{\textbf{Checkpoint 1}}
  & \multicolumn{2}{c}{\textbf{Checkpoint 2}}
  & \multicolumn{2}{c}{\textbf{Checkpoint 3}} \\
\cmidrule(lr){4-5}\cmidrule(lr){6-7}
& & & \textbf{w/o expert answer} & \textbf{w/ expert answer}
  & \textbf{w/o expert answer}   & \textbf{w/ expert answer} \\
\midrule
GPT-5 (high, code \& web) & \colorscore{2/5} & \colorscore{5/5} & \colorscore{5/5} & \colorscore{5/5} & \colorscore{3/5} & \colorscore{3/5} \\
GPT-5 (high, code)        & \colorscore{4/5} & \colorscore{5/5} & \colorscore{5/5} & \colorscore{5/5} & \colorscore{5/5} & \colorscore{4/5} \\
GPT-5 (high)              & \colorscore{0/5} & \colorscore{5/5} & \colorscore{5/5} & \colorscore{5/5} & \colorscore{0/5} & \colorscore{0/5} \\
Gemini-2.5 Pro            & \colorscore{0/5} & \colorscore{3/5} & \colorscore{3/5} & \colorscore{1/5} & \colorscore{0/5} & \colorscore{0/5} \\
o3 (high)                 & \colorscore{0/5} & \colorscore{5/5} & \colorscore{2/5} & \colorscore{2/5} & \colorscore{0/5} & \colorscore{0/5} \\
DeepSeek R1               & \colorscore{0/5} & \colorscore{3/5} & \colorscore{3/5} & \colorscore{2/5} & \colorscore{0/5} & \colorscore{0/5} \\
o4-mini (high)            & \colorscore{0/5} & \colorscore{5/5} & \colorscore{1/5} & \colorscore{0/5} & \colorscore{0/5} & \colorscore{0/5} \\
Gemini-2.5 Flash          & \colorscore{0/5} & \colorscore{2/5} & \colorscore{0/5} & \colorscore{0/5} & \colorscore{0/5} & \colorscore{0/5} \\
Claude Opus 4             & \colorscore{0/5} & \colorscore{0/5} & \colorscore{0/5} & \colorscore{0/5} & \colorscore{0/5} & \colorscore{0/5} \\
GPT-5 (minimal)           & \colorscore{0/5} & \colorscore{0/5} & \colorscore{0/5} & \colorscore{0/5} & \colorscore{0/5} & \colorscore{0/5} \\
Llama-4 Maverick          & \colorscore{0/5} & \colorscore{0/5} & \colorscore{0/5} & \colorscore{0/5} & \colorscore{0/5} & \colorscore{0/5} \\
GPT-4o                    & \colorscore{0/5} & \colorscore{0/5} & \colorscore{0/5} & \colorscore{0/5} & \colorscore{0/5} & \colorscore{0/5} \\
\bottomrule
\end{tabular}
\caption{Detailed breakdown of evaluation results for the example challenge, ``quantum error detection.''}
\label{tab:qec_complete}
\end{threeparttable}
\end{table}

This interface streamlines non-AI researchers' engagement with LLM outputs at scale, by abstracting away the technical complexity of API access and evaluation infrastructures, thereby making it easier to provide structured feedback. 
For the example challenge, the expert's detailed feedback is provided in~\ref{SI:example_analysis}. 
The expert notes that although GPT-5 (high) achieves the highest final-answer accuracy, its output formatting is cluttered and difficult to follow, potentially limiting its usefulness in realistic workflows, and this may be a solvable problem in natural language processing.
Another interesting observation is that GPT-5 (high) often calls tools in ways that diverge from expert expectations: using code execution even when an analytical solution would be simpler, or performing excessive web retrieval before assessing relevance. 
These behaviors reveal a gap between current LLM decision heuristics and human-expert research intuition.


%% file: text/05-conclusion.tex
\section{Conclusion}
In this work, we introduce \CritPt, a physics reasoning benchmark designed by experts to probe the capacity of LLMs to meet the authentic reasoning demands of frontier research. 
Moving beyond traditional evaluation formats, \CritPt's self-contained challenges mimic how a mentor frames a problem for a junior researcher, while its modular checkpoints allow for a fine-grained analysis of reasoning capabilities without sacrificing scientific depth. 
By assembling a diverse set of unpublished, guess-resistant problems developed by 50+ active physicists, we provide the first systematic testbed grounded in the realistic workflows, complexity, and failure sensitivity of modern physics.

Our evaluation results demonstrate a striking gap between current model performance and the depth, rigor, creativity and precision required for physics research. 
While advanced reasoning models such as GPT-5 (high) show early promise on narrowly scoped tasks, particularly when augmented with coding tools, their ability to maintain coherence and correctness through full-scale challenges remains minimal. 
Furthermore, our reliability analysis shows that even correct answers are often not consistently reproducible, highlighting a severe deficit in the robustness required for high-stakes scientific applications.

Beyond a benchmark, \CritPt represents a collaborative bridge between the physics and AI communities.
For AI developers, it translates the abstract reasoning demands of physics research into a standardized dataset and provides concrete and accessible signals via a physics-informed automatic evaluation pipeline, providing immediate high-quality feedback for model development. 
For physicists, it offers a grounded introduction to the current capabilities and limitations of AI assistants, supported by an interactive visualization tool that enables domain experts to efficiently inspect, analyze, and critique model outputs at scale. 
This process creates a shared vocabulary, clarifying the requirements for future systems that aim not just to answer textbook questions, but to engage with the complex and iterative workflows of scientific discovery. 

Looking forward, \CritPt is a powerful and scientifically grounded framework to measure progress. 
It offers a clear metric of how far AI has come and, more importantly, how far it has yet to go to augment genuine discovery in physics. 
We hope this benchmark will not only guide the development of more capable and reliable reasoning models but also catalyze deeper conversations between the physics and AI communities on what it truly means for a machine to reason in a scientific context.

%% file: text/06-SI.tex
\section{Appendix}
\subsection{List of \CritPt challenges}\label{SI:problem_list}

\href{https://critpt.com/example}{Example challenge}~\cite{Gottesman2016,Vuillot2017,Linke2017,Harper2019}:
\begin{itemize}
    \item Quantum error detection 
\end{itemize}

\href{https://github.com/CritPt-Benchmark/CritPt}{Test set (70 challenges)} 
\cite{komar2014quantum,zhang2021distributed,zang2024quantum,zang2025enhancing,PhysRevD.23.347,LINDE1982389,PhysRevLett.48.1220,LINDE1983177,PhysRevLett.65.3233,PhysRevD.52.5529,Nima_Arkani-Hamed_2003,PhysRevLett.108.261302,PhysRevD.86.043530,PhysRevD.90.023501,Maleknejad_2016,Fujita_2022,PhysRevD.81.123529,PhysRevD.110.043521,Shamit_Kachru_2003,witten1978some,goldschmidt1986kosterlitz,moore1991nonabelions,milovanovic1996edge,schiller2023superconductivity,cao2024signatures,may2024theory,kitaev2006anyons,nayak2008non,hastings2013metaplectic,clarke2013exotic,lindner2012fractionalizing,cheng2012superconducting,barkeshli2013theory,fendley2012parafermionic,verlinde1988fusion,frohlich2004kramers,frohlich2007duality,francesco2012conformal,chang2019topological,huang2024arxiv,fan2024diagnostics,dennis2002topological,nishimori2001statistical,zaletel2014arxiv,cirac2021rmp,huang2025topological,sun2011nearly,neupert2011fractional,mai20231,mai2023topological,mai2024topological,PhysRevB.94.035135,PhysRevD.94.106002,Izubuchi:2018srq,Moch:2004pa,Su:2022fiu,ji2021large,ji2013parton,Ji:2014gla,horodecki2009general,smith2025additivity,lesniewski1999monotone,hiai2016contraction,hoang2016torsional,zhang2025scalable,rieser2022tunable,manceau2017improving,nehra2022few,Murase:2018iyl,Padovani:2019xcv,boyer1968quantum,brown1986geonium,brown1986cyclotron,barton1988quantum,fan2023measurement,Kitagawa1993squeezed,wineland1994squeezed,ma2011quantum,chu2021quantum,barberena2024trade,goluskin2016internally,liu2024fixed,chandrasekhar2013hydrodynamic,busse1967stability,Manneville2006,liu2022single,busse1978non,nield2006convection,nield2019brief,trevisan1987mass,hewitt2020vigorous,Beacom:2006tt,Alenezi:2025kwl,bartolotta2022entropy,bellman_reciprocity_1961,jeffrey_table_2007,turner_quantum_2018,zhou_hydrodynamic_2023,hallatschek_acceleration_2014,zhou_emergent_2019,chen2007interactions,tsang2016quantum,meyer2015connectivity,collins2006integration,reilly2024speeding,wilson2024entangled,jager2022lindblad,zhang2022building,horvat2021quantum,PhysRevB.109.115154,PhysRevB.106.085142,PhysRevB.72.104510,greiner2002quantum,bloch2012quantum,gross2017quantum,gross2021quantum,young2022tweezer,kim2017,fossfeig2020,Barratt2021,Chertkov2022,Niu2021,Zhang2022,DeCross2022,PhysRevX.8.031029,PhysRevResearch.2.023348,Qi2019determininglocal,Powell1956,Jafarpour2018,Barber2021,hein2024asymptotic,levien2025size,hinshelwood1952136,Gray2006CirculantReview,hein2024competition,Dauxois2009AutocatRing,TogashiKaneko2001,zhao2022thermodynamics,zhao2023failure,zhao2023proof,mai2024new,ma2025charge,la2025luttinger,jain2023dipole,jain2024dipole,stahl2023fracton,glorioso2023goldstone,yang2020electronic,yang2020quantum,hiraoka2020direct,holzmann2016theory,drummond2008finite,chiesa2006finite,gerard2018lax,lenzmann2020derivation,zhou2015solitons,lenzmann2018short,stanley1986enumerative,simion2000noncrossing,zhou2019emergent,hsu2013dynamical,fradkin2009scaling,parker2017entanglement,brunet2016some,chatterjee2016multiple,turner2018weak,nahum2017quantum,zhou2017operator,noyes1970strong,PhysRevC.47.1876,PhysRevA.83.063614,PhysRevLett.120.100401,PhysRevA.99.043604,fefferman2012ambient,graham2009extended,jia2021obstruction,jia2023weyl,henningson1998holographic,ciambelli2020weyl,safronova2012ytterbium,mitroy2010theory,le2013dynamical,Tang_2018,fujita2011aspects,grinberg2021proper,geng2025microscopic,chua2024hartle,horvat2021interference,chen2024information,maisriml2025acquisition,kochen2011problem,spekkens2005contextuality,klyachko2008simple,kunjwal2015kochen,zhang2025quantifiers,zhang2503reassessing,dotsenko1984conformal,dijkgraaf1988c,aharony2004hagedorn,kinney2007index,yu2025universal,yu2025wilson,yu2024quantum,roy2014band,yang2025engineering,hernandez2013attosecond,dorney2019controlling,rego2019generation,brooks2024circularly,PhysRevLett.121.241102,PhysRevD.101.123003,PhysRevD.104.103030,PhysRevResearch.1.033187,PhysRevLett.128.121101,Vermeulen:2021uf,hall2022advancedligolisacosmic,PhysRevD.103.L051702,PhysRevLett.121.061102,PhysRevD.105.063030,PhysRevD.103.103002,sunko2023spin,donoway2024multimodal,sie2017large,sie2015Valley,overhauser1971observability,Pospelov:2007mp,PhysRevB.76.233411,adam2007self,PhysRevB.101.115115,PhysRevB.101.115140,PhysRevB.92.014403,zhu2025emergent,PhysRevB.111.L100402,feng2025transient,su2017spatially,abbamonte2025collective,mitrano2024exploring,zhou2025distinguishability,hirche2022contraction,PhysRevLett.122.086402,PhysRevB.108.085117,PhysRevLett.132.036501,PhysRevB.109.045147,PhysRevB.109.205121}:

\begin{enumerate}
    \item Holographic Weyl anomaly
    \item Population growth rate from stochastic model of growth and cell-size regulation
    \item Geodesic in AdS/BCFT of a black hole
    \item Orbital angular momentum conservation in high harmonic generation
    \item Marchenko-Pastur entropy
    \item Twisted Bilayer MoTe$_2$
    \item Noisy quantum sensing
    \item Scalar spectrum in Nieh–Yan gravity
    \item Nieh-Yan modified gravity with Torsion in FRW space-time
    \item Axion inflation with Chern-Simons
    \item Gapped edge of the Moore-Read state
    \item Parafermion zero modes tunneling
    \item Verlinde lines in the Moore-Read CFT
    \item High/low-temperature duality in Ising Torus
    \item Decohered Affleck-Kennedy-Lieb-Tasaki (AKLT) model
    \item Interacting Chern insulator
    \item Zero temperature entropy in Sachdev-Ye-Kitaev models
    \item Optical binding force
    \item Cascade optical parametric amplifier
    \item Torsional levitated optomechanics
    \item Numerical LaMET matching
    \item Single particle Holevo Information
    \item One-loop correction of quasi-PDF 
    \item LaMET matching in Coulomb gauge
    \item Minimum Doppler factor of a relativistic jet
    \item Spherical cavity shifts
    \item One-axis twisting model with dissipation
    \item Optical conductivity of the Hubbard model
    \item Hubbard model in an optical lattice
    \item Orthogonal non-isometric maps
    \item Linear stability analysis of Rayleigh-B\'enard convection
    \item Rayleigh-Darcy convection with mixed boundary conditions
    \item Condensation of three types of particles
    \item Quantum tensor networks
    \item Quantum inverse problem
    \item Oscillation amplitude in transient dynamics of autocatalytic cycles
    \item Quantum geometry
    \item Scattering rate of the Hatsugai-Kohmoto model
    \item Alteration of cavity field coherences due to atom-cavity interaction
    \item Hydrodynamic modes in Schwinger-Keldysh
    \item Energy in many body quantum systems
    \item Graphene minimal conductivity
    \item Disclination charge
    \item Ground state in Kitaev honeycomb model
    \item Goniopolarity in semiconductor
    \item PXP scar
    \item Integrals of motion
    \item Lattice Gaussian sum
    \item Long-range light cone
    \item Many-body NC Partitions
    \item Random walk $a_3(t)$
    \item Efimov effect in three body problem
    \item Extended obstruction tensors
    \item Magic wavelengths for Yb isotopes
    \item INS cross-section for scattering from an oscillator
    \item Dark matter detection with Cosmic Explorer
    \item LIGO with modified mirrors for ultralight vector dark matter
    \item Magnetic space group identification
    \item Charge density wave diffraction
    \item Superresolution
    \item Quantum search time
    \item Quantum games in multi-slit interference
    \item Noise robustness in Kochen–Specker and Spekkens contextuality
    \item Conformal correlators
    \item Constructing fermionic matrix operators
    \item Generating function of index
    \item Quantum capacity for quantum channels
    \item Quantum $f$-divergence
    \item Contraction coefficients for quantum relative entropy
    \item Row-Twirling channel on qubit lattice

\end{enumerate}

\newpage
\subsection{Prompts for two-step answer generation from model}\label{SI:prompt}
\begin{tcolorbox}[
  enhanced,
  title=Step 1: system prompt for full problem solving,
  colback=myback,
  colframe=myframe,
  fonttitle=\bfseries,
  coltitle=white,
  colbacktitle=promptframe,
  left=4mm, right=4mm, top=4mm, bottom=3mm,
  attach boxed title to top left={yshift=-2mm, xshift=5mm},
  boxed title style={arc=2mm, drop shadow},
  arc=3mm,
  boxrule=1pt,
  drop shadow,
]
You are a physics research assistant specializing in solving complex research-level problems using precise, step-by-step reasoning.

\vspace{0.5em}
\textbf{Input}

Problems will be provided in Markdown format.

\vspace{0.5em}

\textbf{Output (Markdown format)}

\begin{enumerate}
    \item \textbf{Step-by-Step Derivation} -- Show every non-trivial step in the solution. Justify steps using relevant physical laws, theorems, mathematical identities (or numerical codes)~\footnote{ Content in the parenthesis is only given when code interpreter is enabled.}.
    
    \item \textbf{Mathematical Typesetting} -- Use LaTeX for all mathematics: \texttt{\$...\$} for inline expressions, \texttt{\$\$...\$\$} for display equations.
    
    \item \textbf{Conventions and Units} -- Follow the unit system and conventions specified in the problem.
    
    \item \textbf{Final Answer} -- At the end of the solution, start a new line with \textbf{``Final Answer:''}, and present the final result.
    
    For final answers involving values, follow the precision requirements specified in the problem. If no precision is specified:
    \begin{itemize}
        \item If an exact value is possible, provide it (e.g., $\sqrt{2}$, $\pi/4$).
        \item If exact form is not feasible, retain at least 12 significant digits in the result.
    \end{itemize}
    
    \item \textbf{Formatting Compliance} -- If the user requests a specific output format (e.g., code, table), provide the final answer accordingly.
\end{enumerate}

\end{tcolorbox}

\definecolor{codegray}{rgb}{0.6, 0.6, 0.6}
\definecolor{pythonkeyword}{rgb}{0.58, 0.13, 0.55}   
\definecolor{pythonstring}{rgb}{0.0, 0.5, 0.0}     
\definecolor{pythoncomment}{rgb}{0.5, 0.5, 0.5}    

\begin{tcolorbox}[
  enhanced,
  title=Step 2: parsing prompt for answer formatting,
  colback=myback,
  colframe=myframe,
  fonttitle=\bfseries,
  coltitle=white,
  colbacktitle=promptframe,
  left=4mm, right=4mm, top=4mm, bottom=3mm,
  attach boxed title to top left={yshift=-2mm, xshift=5mm},
  boxed title style={arc=2mm, drop shadow},
  arc=3mm,
  boxrule=1pt,
  drop shadow,
]
Populate your final answer into the code template provided below. This step is purely for formatting/display purposes. No additional reasoning or derivation should be performed. Do not import any modules or packages beyond what is provided in the template.

\begin{tcblisting}{
  listing engine=listings,
  listing only,
  arc=0mm,
  boxrule=0pt,
  colback=white,
  left=1mm, right=0mm, top=0mm, bottom=0mm, % Tight fit
  listing options={
    language=Python,
    basicstyle=\ttfamily\fontsize{8pt}{9.5pt}\selectfont\color{black},
    commentstyle=\color{pythoncomment},
    keywordstyle=\color{pythonkeyword}\bfseries,
    stringstyle=\color{pythonstring},
    numbers=left,
    numberstyle=\tiny\color{codegray},
    showstringspaces=false,
    upquote=true,
  }
}
import sympy as sp
p = sp.symbols('p')

def answer(p):
    r"""
    Return the expression of the logical state fidelity 
    in SymPy format.

    Inputs
    -------
    p: sympy.Symbol, two-qubit gate error rate, $p$

    Outputs
    -------
    F_logical: sympy.Expr, logical state fidelity
    """

    ----------- FILL IN YOUR RESULTS BELOW ------------
    F_logical = ...  # a SymPy expression of inputs
    ---------------------------------------------------

    return F_logical
\end{tcblisting}
\end{tcolorbox}

\newpage
\subsection{Detailed API setup used in evaluation}\label{SI:eval_method}

We evaluate eight proprietary models across four major providers (OpenAI, Google, DeepSeek, Anthropic) using Inspect AI~\cite{UK_AI_Security_Institute_Inspect_AI_Framework_2024}, and one open-source model (Meta Llama-4 Maverick) via Together AI~\cite{togetherai}. 

From OpenAI, we evaluate four models. These include their flagship reasoning model, GPT-5 \cite{gpt5}, as well as two earlier-generation models from the o-series: the large o3, designed for extended multi-step reasoning, and its lighter variant, o4-mini, optimized for latency and cost \cite{o3ando4-mini}. We also assess GPT-4o, a general-purpose multimodal model~\cite{gpt4o}.
From Google, we evaluate Gemini 2.5 Pro, their most advanced “thinking model” to date, which excels at complex reasoning, coding, and scientific tasks~\cite{gemni-2-5}. We also include Gemini 2.5 Flash, a lightweight, high-throughput variant engineered for speed and efficiency, while retaining core reasoning capabilities~\cite{gemni-2-5}.
From DeepSeek, we evaluate DeepSeek R1, an open-weight model trained with multi-stage reinforcement learning to enable multi-step reasoning and self-reflection~\cite{guo2025Deepseek}. It achieves performance comparable to OpenAI-o1-1217~\cite{jaech2024openai}.
From Anthropic, we include Claude Opus 4, their flagship model for complex coding and reasoning, with improved precision in instruction-following~\cite{claude4}.
Finally, from Meta, we evaluate Llama-4 Maverick, an open‑weight, instruction‑tuned multimodal model with strong performance on a broad range of widely reported benchmarks~\cite{llama4}.

For reproducibility, model configurations including API names and configurations are summarized in Table~\ref{tab:model-config}. 

\captionsetup[table]{skip=6pt}

\begin{table}[h!]
\centering
\renewcommand{\arraystretch}{1.7}
\begin{threeparttable}
\small
\begin{tabular}{>{\raggedright}p{3.3cm} p{3.9cm} p{3.7cm} p{2.0cm}}
\toprule
\textbf{Model Name} & \textbf{API Name} & \textbf{Reasoning Effort} & \textbf{Tool Use} \\
\midrule
GPT-5 (high, code \& web) & \texttt{gpt5-2025-08-07} & \texttt{high} & code interpreter, web search \\
GPT-5 (high, code) & \texttt{gpt5-2025-08-07} & \texttt{high} & code interpreter \\
GPT-5 (high) & \texttt{gpt5-2025-08-07} & \texttt{high} & / \\
o3 (high) & \texttt{o3-2025-04-16} & \texttt{high} & / \\
o4-mini (high) & \texttt{o4-mini-2025-04-16} & \texttt{high} & / \\
Gemini 2.5 Pro & \texttt{gemini-2.5-pro} & \texttt{reasoning\_tokens=27000} & / \\
Gemini 2.5 Flash & \texttt{gemini-2.5-flash} & \texttt{Default} & / \\
DeepSeek R1 & \texttt{deepseek-reasoner} & \texttt{Default} & / \\
Claude Opus 4 & \texttt{claude-opus-4-20250514} & \texttt{reasoning\_tokens=27000} & / \\
\rowcolor{gray!10}
GPT-5 (minimal) & \texttt{gpt5-2025-08-07} & \texttt{minimal}\tnote{*} & / \\
\rowcolor{gray!10}
Llama-4 Maverick & \parbox[t]{4.0cm}{\texttt{Llama-4-Maverick-17B-}\\\texttt{128E-Instruct-FP8}} & \texttt{/} & / \\
\rowcolor{gray!10}
GPT-4o & \texttt{chatgpt-4o-latest} & \texttt{/} & / \\
\bottomrule
\end{tabular}
\begin{tablenotes}
\scriptsize
\item[*] GPT-5 (minimal) sets \texttt{reasoning\_effort=minimal}, which means zero reasoning tokens used.
\end{tablenotes}
\end{threeparttable}
\caption{API configurations used in our evaluation.}
\label{tab:model-config}
\end{table}

\newpage
\subsection{API usage statistics}\label{SI:API_stat}
All evaluations are conducted using the standard API access provided by each company, ensuring scalability and consistency across models.
As shown in Table~\ref{tab:challenge-token-price}, for each challenge, we perform five independent runs, and then calculate the average statistics per run, including number of input tokens, cached input tokens,  reasoning tokens, response tokens, which adds up to total tokens. 

The average cost per run (in USD) is determined by recorded tokens in each category and their respective pricing (in the unit of USD per million tokens).
The input cost is computed by charging cached input tokens at the cached input rate and the remaining input tokens at the standard input rate. 
The output cost is computed from response tokens and reasoning tokens at the output rate. 
These two components then sum up to the total cost.

When models are augmented with tools, both input tokens and cached input tokens increase significantly. 
This is due to tool use inducing a multi-turn interaction: each round produces output that is fed back as input to the next round. Repeated context like prior turns is served from cache, inflating the cached input token counts. 
By contrast, reasoning tokens and output tokens in the table reflect only the final round, so intermediate generation across earlier turns appears on the input side rather than as additional output.

\newcolumntype{L}[1]{>{\raggedright\arraybackslash}p{#1}}
\newcolumntype{R}[1]{>{\raggedleft\arraybackslash}p{#1}}

\captionsetup[table]{skip=6pt}
\begin{table}[h!]
\centering
\renewcommand{\arraystretch}{1.6}
\begin{threeparttable}
\small
\begin{tabular}{
  L{2.3cm} | 
  R{0.8cm}R{0.7cm} | 
  R{0.8cm}R{0.7cm} | 
  R{0.8cm}R{0.7cm} | 
  R{0.8cm}R{0.7cm} | 
  R{0.9cm}R{0.9cm}   
}
\toprule
\multirow{2}{*}{\textbf{Model Name}} 
& \multicolumn{2}{c}{\textbf{Input}} 
& \multicolumn{2}{c}{\textbf{Cached Input}} 
& \multicolumn{2}{c}{\textbf{Reasoning}} 
& \multicolumn{2}{c}{\textbf{Response}} 
& \multicolumn{2}{c}{\textbf{Total}} \\
\cmidrule(lr){2-3}\cmidrule(lr){4-5}\cmidrule(lr){6-7}\cmidrule(lr){8-9}\cmidrule(lr){10-11}
& \textbf{Tokens} & \textbf{Price (\$/1M)}
& \textbf{Tokens} & \textbf{Price (\$/1M)}
& \textbf{Tokens} & \textbf{Price (\$/1M)}
& \textbf{Tokens} & \textbf{Price (\$/1M)}
& \textbf{Tokens} & \textbf{Cost/run (\$)}\\
\midrule
GPT-5 (high, code \& web) & 330569 & 1.25  & 291050 & 0.125 & 10600 & 10.0 & 2286 & 10.0 & 343455 & 0.578 \\
GPT-5 (high, code)        & 17728  & 1.25  & 16539  & 0.125 & 11379 & 10.0 & 1996 & 10.0 & 31102  & 0.158 \\
GPT-5 (high)              & 864    & 1.25  & 55     & 0.125 & 24686 & 10.0 & 2035 & 10.0 & 27585  & 0.268 \\
o3 (high)                 & 864    & 2.0   & 200    & 0.5   & 17356 & 8.0  & 1322 & 8.0  & 19653  & 0.151 \\
o4-mini (high)            & 864    & 1.1   & 140    & 0.275 & 16096 & 4.4  & 1536 & 4.4  & 18496  & 0.079 \\
Gemini 2.5 Pro            & 1012   & 1.25  & 0      & 0.31  & 20943 & 10.0 & 3196  & 10.0 & 25151  & 0.243 \\
Gemini 2.5 Flash          & 1012   & 0.3   & 0      & 0.075 & 22031 & 2.5  & 3726  & 2.5  & 26769  & 0.065 \\
DeepSeek R1               & 821    & 0.56  & 320    & 0.07  & 19656 & 1.68 & 1236 & 1.68 & 21713  & 0.036 \\
Claude Opus 4             & 1005   & 15.0  & 0      & 1.5   & 4566  & 75.0 & 8637 & 75.0 & 14208  & 1.005 \\
GPT-5 (minimal)           & 864    & 1.25  & 24     & 0.125 & 0     & 10.0 & 2315  & 10.0 & 3179   & 0.024 \\
GPT-4o                    & 865    & 2.5   & 216    & 1.25  & 0     & 10.0 & 950   & 10.0 & 1815   & 0.012 \\
Llama-4 Maverick          & 818    & 0.27  & 0      & 0.0   & 0     & 0.85 & 1383  & 0.85 & 2201   & 0.001 \\
\bottomrule
\end{tabular}
\caption{Average token usage, pricing and cost across models in the evaluation of \CritPt challenges.}
\label{tab:challenge-token-price}
\end{threeparttable}
\end{table}

\newpage
\subsection{Detailed analysis on example challenge: Quantum error detection}\label{SI:example}

\subsubsection{Design idea from expert}\label{SI:example_idea}
Quantum error correction (QEC) — the science of how to scalably suppress quantum noise in digital quantum computers — is a rapidly developing field that is becoming increasingly important to the development of quantum computing hardware. A key goal of QEC is to find QEC codes, protocols for encoding many physical qubits into logical qubits, with desirable properties, such as low time and space overhead and large error suppression. In this problem, we analyze some properties of the simplest possible QEC code, a small error detection code that can only detect errors and not correct them. Given this code’s small size, its properties can be worked out analytically, making it a useful test bed for understanding conceptually properties of more complicated large-scale QEC codes that can only be studied with numerical simulations, such as Monte Carlo sampling. This problem illustrates a set of questions a QEC researcher interested in this QEC code might ask to better understand its practical performance for a specific task, preparing a logical quantum state encoded in the code.

In subproblem 1, we describe a logical state preparation protocol for a specific quantum state in the QEC code and ask what the \textit{physical} fidelity for that protocol is in terms of the error rate p in an idealized quantum computer with noisy gates. This quantity measures how much the quantum state is affected by the noise on the quantum computer without the help of error detection. It is a simple calculation that can be done by hand that requires knowing a little bit about the properties of stabilizers and Pauli matrices, common and well-known topics in quantum information and QEC.

In subproblem 2, we perform a similar calculation but now instead compute the \textit{logical} fidelity of the logically encoded quantum state. This quantity measures how much the logical quantum state is affected by noise, and should be lower than the physical fidelity if the QEC code is working well. Comparing the logical and physical fidelities of operations in a QEC code is key to understanding the performance of the QEC code, so is a calculation of interest to researchers in the field. This calculation is a bit more involved than the previous subproblem, but still only relies on understanding the mathematics of stabilizers and Pauli matrices though in a more complicated setting. With patience this problem can be worked out by hand, but it is easier to solve it by writing some simple code that performs some combinatorics. The combinatorial calculation is necessary to understand how all of the different possible quantum gate errors lead to logical errors in the QEC code.

In subproblem 3, we now consider a different logical state preparation protocol for the same QEC code and ask for the logical state fidelity for that protocol. This state preparation protocol is significantly more complicated than the previous one. To solve this problem, one needs to write code to perform the combinatorics. Moreover, this code needs to be much more complicated than that required for subproblem 1. In addition to performing counting, it also needs to simulate how gate errors propagate through a quantum circuit, which involves running a non-trivial algorithm. Being able to answer problems such as this quickly, which can be stated quite simply but whose solution involves complex multi-step reasoning and code development, would be quite valuable to QEC researchers looking to quickly test many QEC codes.

\subsubsection{Expert feedback on model responses}\label{SI:example_analysis}
\begin{itemize}
\item \textbf{Subproblem 1:}
Many models are able to solve this subproblem. Essentially all models understand the problem setup and the logic of how to obtain a solution.There is a sharp divide between models that are able to solve the problem correctly (e.g., o3 (high), o4-mini (high), GPT-5 (high), DeepSeek R1, Gemini 2.5 Pro) and those that are not. The incorrect models tend to immediately fail in the first steps of the calculation or make an illogical claim without support that leads to the wrong answer. Sometimes, the output of the model is long, convoluted, and horribly formatted, making it nearly impossible to tell where a logical mistake could have been made. Even among the correct models, there is quite a bit of separation in the quality of answers. o3 (high) produces some of the clearest, simple, and easy-to-understand derivations of the final solution. Other models, such as GPT-5 (high), o4-mini (high), and Gemini 2.5 Pro, produce answers with poor formatting (such as excessive use of bullet points) that are unnecessarily difficult to understand. These answers would be less useful for researchers, since it would make it difficult for them to verify the correctness of the result. DeepSeek R1 and Gemini 2.5 Pro appear to be at the edge of being able to solve the problem. They often answer correctly, but occasionally make an algebra mistake during the calculation. Other models, such as Gemini 2.5 Flash, GPT-4o, Llama-4 Maverick , would get lost in the derivation and often resort to guessing answers. The GPT-5 (high, code) and GPT-5 (high, code \& search) write code to solve the problem, which works but is completely unnecessary as this problem can be solved with simple counting that can be done by hand.

\item \textbf{Subproblem 2:}
In this subproblem, almost all of the models understand the high-level idea, which involves counting up different possible errors, assigning them probabilities, and combining the probabilities together in the correct way. However, most models are unable to execute the details of the calculation correctly. For the models able to solve the problem, when they give the correct answer, their explanations are generally clear, concise, and readable, making them useful for a researcher to understand the problem. When these models give incorrect answers, the derivations tend to become more vague, unclear, and poorly formatted. For the best performing models, when they give an incorrect answer the output tends to be well formatted and plausible, with more subtle errors hidden in the algebra or assumptions. This makes these models a bit dangerous to use, since their output can appear reasonable on their face but actually be completely wrong. The models that always give incorrect answers tend to have either minimal or confusing logic and produce wildly different solutions in each attempt, suggesting that they are guessing. At least in these models, it is quite apparent that the solutions are unreliable, which makes them less dangerous than the sometimes subtly-incorrect models. The GPT-5 (high, code) can reliably solve this problem. They write and execute code, which is not strictly necessary to solve the problem, but is a reasonable approach that works. However, despite giving consistently correct answers, all GPT-5's solutions are formatted quite poorly, often as a list of bullet points with incomplete sentences or vague comments. In principle, a researcher could decipher the material, but it certainly would be more useful if it was presented in a cleaner format, such as one might find in course lecture notes or a textbook.

\item \textbf{Subproblem 3:}
All models except for GPT-5 (high, code) and GPT-5 (high, code \& search) completely fail on this subproblem. This subproblem requires using the same general reasoning as subproblem 2, but now in a more complicated setting that involves writing and executing code to perform the counting of quantum errors. All other models (including GPT-5 (high)) perform rampant guessing, inventing the numbers needed to get a final solution. The best incorrect models tend to produce answers that are some combination of vague, confusing, poorly formatted, or excessively short. The worst incorrect models simply repeat the information presented in the prompt and guess an answer (such as 1). Most incorrect models produce different random answers in each trial. In all cases, it would be easy for a researcher to see that these models are unable to give a correct solution. However, in no case did the incorrect models indicate that they did not know how to solve the problem, instead giving long and complicated justifications for clearly incorrect solutions. Even though GPT-5 (high, code) could fairly consistently produce the correct answer, its solutions were again poorly formatted (e.g., everything was a list of vague bullet points) making them much less useful than they could be. The poor formatting tends to make me trust the model less, since I have seen poor formatting coincide with incorrect logic in other models. Therefore, even though GPT-5 can often get the right answer here, its presentation to the user can still be significantly improved to make it more trustworthy for research use cases.

\newpage

\end{itemize}

%% file: main.bbl
\begin{thebibliography}{318}
\providecommand{\natexlab}[1]{#1}
\providecommand{\url}[1]{\texttt{#1}}
\expandafter\ifx\csname urlstyle\endcsname\relax
  \providecommand{\doi}[1]{doi: #1}\else
  \providecommand{\doi}{doi: \begingroup \urlstyle{rm}\Url}\fi

\bibitem[()Anderson]{anderson1972more}
P.~W. Anderson.
\newblock More is different: broken symmetry and the nature of the hierarchical structure of science.
\newblock \emph{Science}, 177\penalty0 (4047):\penalty0 393--396, 1972.

\bibitem[Sinatra et~al.(2015)Sinatra, Deville, Szell, Wang, and Barab{\'a}si]{sinatra2015century}
R.~Sinatra, P.~Deville, M.~Szell, D.~Wang, and A.-L. Barab{\'a}si.
\newblock {A century of physics}.
\newblock \emph{Nature Physics}, 11\penalty0 (10):\penalty0 791--796, 2015.

\bibitem[Vaswani et~al.(2017)Vaswani, Shazeer, Parmar, Uszkoreit, Jones, Gomez, Kaiser, and Polosukhin]{vaswani2017attention}
A.~Vaswani, N.~Shazeer, N.~Parmar, J.~Uszkoreit, L.~Jones, A.~N. Gomez, {\L}.~Kaiser, and I.~Polosukhin.
\newblock Attention is all you need.
\newblock \emph{Advances in neural information processing systems}, 30, 2017.

\bibitem[Devlin et~al.(2019)Devlin, Chang, Lee, and Toutanova]{devlin2019bert}
J.~Devlin, M.-W. Chang, K.~Lee, and K.~Toutanova.
\newblock {BERT}: Pre-training of deep bidirectional transformers for language understanding.
\newblock pages, 4171--4186, 2019.

\bibitem[Raffel et~al.(2020)Raffel, Shazeer, Roberts, Lee, Narang, Matena, Zhou, Li, and Liu]{raffel2020exploring}
C.~Raffel, N.~Shazeer, A.~Roberts, K.~Lee, S.~Narang, M.~Matena, Y.~Zhou, W.~Li, and P.~J. Liu.
\newblock Exploring the limits of transfer learning with a unified text-to-text transformer.
\newblock \emph{Journal of machine learning research}, 21\penalty0 (140):\penalty0 1--67, 2020.

\bibitem[Delgado-Chaves et~al.(2025)Delgado-Chaves, Jennings, Atalaia, Wolff, Horvath, Mamdouh, Baumbach, and Baumbach]{delgado2025transforming}
F.~M. Delgado-Chaves, M.~J. Jennings, A.~Atalaia, J.~Wolff, R.~Horvath, Z.~M. Mamdouh, J.~Baumbach, and L.~Baumbach.
\newblock Transforming literature screening: The emerging role of large language models in systematic reviews.
\newblock \emph{Proceedings of the National Academy of Sciences}, 122\penalty0 (2):\penalty0 e2411962122, 2025.

\bibitem[Scherbakov et~al.(2025)Scherbakov, Hubig, Jansari, Bakumenko, and Lenert]{scherbakov2025emergence}
D.~Scherbakov, N.~Hubig, V.~Jansari, A.~Bakumenko, and L.~A. Lenert.
\newblock The emergence of large language models as tools in literature reviews: a large language model-assisted systematic review.
\newblock \emph{Journal of the American Medical Informatics Association}, 32\penalty0 (6):\penalty0 1071--1086, 2025.

\bibitem[Pramanick et~al.(2024)Pramanick, Chellappa, and Venugopalan]{pramanick2024spiqa}
S.~Pramanick, R.~Chellappa, and S.~Venugopalan.
\newblock {SPIQA}: A dataset for multimodal question answering on scientific papers.
\newblock \emph{Advances in Neural Information Processing Systems}, 37:\penalty0 118807--118833, 2024.

\bibitem[Gao et~al.(2023)Gao, Yen, Yu, and Chen]{gao2023enabling}
T.~Gao, H.~Yen, J.~Yu, and D.~Chen.
\newblock Enabling large language models to generate text with citations.
\newblock In \emph{{The 2023 Conference on Empirical Methods in Natural Language Processing}}, 2023.

\bibitem[Wang et~al.(2024)Wang, Guo, Yao, Zhang, Zhang, Wu, Zhang, Dai, Wen, Ye, et~al.]{wang2024autosurvey}
Y.~Wang, Q.~Guo, W.~Yao, H.~Zhang, X.~Zhang, Z.~Wu, M.~Zhang, X.~Dai, Q.~Wen, W.~Ye, et~al.
\newblock Autosurvey: Large {L}anguage {M}odels can automatically write surveys.
\newblock \emph{Advances in neural information processing systems}, 37:\penalty0 115119--115145, 2024.

\bibitem[Asai et~al.(2024)Asai, He, Shao, Shi, Singh, Chang, Lo, Soldaini, Feldman, D'arcy, et~al.]{asai2024openscholar}
A.~Asai, J.~He, R.~Shao, W.~Shi, A.~Singh, J.~C. Chang, K.~Lo, L.~Soldaini, S.~Feldman, M.~D'arcy, et~al.
\newblock Openscholar: Synthesizing scientific literature with retrieval-augmented {LM}s.
\newblock \emph{arXiv preprint arXiv:2411.14199}, 2024.

\bibitem[Skarlinski et~al.(2024)Skarlinski, Cox, Laurent, Braza, Hinks, Hammerling, Ponnapati, Rodriques, and White]{skarlinski2024language}
M.~D. Skarlinski, S.~Cox, J.~M. Laurent, J.~D. Braza, M.~Hinks, M.~J. Hammerling, M.~Ponnapati, S.~G. Rodriques, and A.~D. White.
\newblock Language agents achieve superhuman synthesis of scientific knowledge.
\newblock \emph{arXiv preprint arXiv:2409.13740}, 2024.

\bibitem[Cui et~al.(2025)Cui, Shamsi, Cheon, Ma, Li, Tikhanovskaya, Norgaard, Mudur, Plomecka, Raccuglia, et~al.]{cuicurie}
H.~Cui, Z.~Shamsi, G.~Cheon, X.~Ma, S.~Li, M.~Tikhanovskaya, P.~C. Norgaard, N.~Mudur, M.~B. Plomecka, P.~Raccuglia, et~al.
\newblock {CURIE:} evaluating {LLMs} on multitask scientific long-context understanding and reasoning.
\newblock In \emph{{The Thirteenth International Conference on Learning Representations}}, 2025.

\bibitem[({\natexlab{a}})OpenAI]{gpt5}
OpenAI.
\newblock Introducing {GPT-5}, 2025{\natexlab{a}}.
\newblock https://openai.com/index/introducing-gpt-5/.

\bibitem[({\natexlab{b}})OpenAI]{o3ando4-mini}
OpenAI.
\newblock {Introducing OpenAI o3 and o4-mini}, 2025{\natexlab{b}}.
\newblock https://openai.com/index/introducing-o3-and-o4-mini/.

\bibitem[({\natexlab{c}})Google]{gemni-2-5}
Google.
\newblock Gemini 2.5: Our most intelligent {AI} model, 2025{\natexlab{c}}.
\newblock https://blog.google/technology/google-deepmind/gemini-model-thinking-updates-march-2025/\#gemini-2-5-thinking.

\bibitem[Guo et~al.(2025)Guo, Yang, Zhang, Song, Wang, Zhu, Xu, Zhang, Ma, Bi, et~al.]{guo2025Deepseek}
D.~Guo, D.~Yang, H.~Zhang, J.~Song, P.~Wang, Q.~Zhu, R.~Xu, R.~Zhang, S.~Ma, X.~Bi, et~al.
\newblock {DeepSeek-R1} incentivizes reasoning in {LLMs} through reinforcement learning.
\newblock \emph{Nature}, 645\penalty0 (8081):\penalty0 633--638, 2025.

\bibitem[()Anthropic]{claude4}
Anthropic.
\newblock Introducing {Claude 4}, 2025.
\newblock https://www.anthropic.com/news/claude-4.

\bibitem[Hurst et~al.(2024)Hurst, Lerer, Goucher, Perelman, Ramesh, Clark, Ostrow, Welihinda, Hayes, Radford, et~al.]{gpt4o}
A.~Hurst, A.~Lerer, A.~P. Goucher, A.~Perelman, A.~Ramesh, A.~Clark, A.~Ostrow, A.~Welihinda, A.~Hayes, A.~Radford, et~al.
\newblock {GPT-4o system card}.
\newblock \emph{arXiv preprint arXiv:2410.21276}, 2024.

\bibitem[()Meta]{llama4}
Meta.
\newblock The {L}lama 4 herd: The beginning of a new era of natively multimodal {AI} innovation, 2025.
\newblock https://ai.meta.com/blog/llama-4-multimodal-intelligence/.

\bibitem[Jaech et~al.(2024)Jaech, Kalai, Lerer, Richardson, El-Kishky, Low, Helyar, Madry, Beutel, Carney, et~al.]{jaech2024openai}
A.~Jaech, A.~Kalai, A.~Lerer, A.~Richardson, A.~El-Kishky, A.~Low, A.~Helyar, A.~Madry, A.~Beutel, A.~Carney, et~al.
\newblock {OpenAI} o1 system card.
\newblock \emph{arXiv preprint arXiv:2412.16720}, 2024.

\bibitem[Comanici et~al.(2025)Comanici, Bieber, Schaekermann, Pasupat, Sachdeva, Dhillon, Blistein, Ram, Zhang, Rosen, et~al.]{comanici2025gemini}
G.~Comanici, E.~Bieber, M.~Schaekermann, I.~Pasupat, N.~Sachdeva, I.~Dhillon, M.~Blistein, O.~Ram, D.~Zhang, E.~Rosen, et~al.
\newblock Gemini 2.5: Pushing the frontier with advanced reasoning, multimodality, long context, and next generation agentic capabilities.
\newblock \emph{arXiv preprint arXiv:2507.06261}, 2025.

\bibitem[Yang et~al.(2025)Yang, Li, Yang, Zhang, Hui, Zheng, Yu, Gao, Huang, Lv, et~al.]{yang2025qwen3}
A.~Yang, A.~Li, B.~Yang, B.~Zhang, B.~Hui, B.~Zheng, B.~Yu, C.~Gao, C.~Huang, C.~Lv, et~al.
\newblock {Qwen3 technical report}.
\newblock \emph{arXiv preprint arXiv:2505.09388}, 2025.

\bibitem[Wei et~al.(2022)Wei, Wang, Schuurmans, Bosma, Xia, Chi, Le, Zhou, et~al.]{wei2022chain}
J.~Wei, X.~Wang, D.~Schuurmans, M.~Bosma, F.~Xia, E.~Chi, Q.~V. Le, D.~Zhou, et~al.
\newblock Chain-of-thought prompting elicits reasoning in large language models.
\newblock \emph{Advances in neural information processing systems}, 35:\penalty0 24824--24837, 2022.

\bibitem[Schick et~al.(2023)Schick, Dwivedi-Yu, Dess\'{\i}, Raileanu, Lomeli, Hambro, Zettlemoyer, Cancedda, and Scialom]{schick2023toolformer}
T.~Schick, J.~Dwivedi-Yu, R.~Dess\'{\i}, R.~Raileanu, M.~Lomeli, E.~Hambro, L.~Zettlemoyer, N.~Cancedda, and T.~Scialom.
\newblock Toolformer: language models can teach themselves to use tools.
\newblock In \emph{{Proceedings of the 37th International Conference on Neural Information Processing Systems}}, 2023.

\bibitem[Wang et~al.(2024)Wang, Chen, Yuan, Zhang, Li, Peng, and Ji]{wang2024executable}
X.~Wang, Y.~Chen, L.~Yuan, Y.~Zhang, Y.~Li, H.~Peng, and H.~Ji.
\newblock Executable code actions elicit better {LLM} agents.
\newblock In \emph{{Proceedings of the International Conference on Machine Learning}}, 2024.

\bibitem[Yuan et~al.(2024)Yuan, Chen, Wang, Fung, Peng, and Ji]{yuan2024craft}
L.~Yuan, Y.~Chen, X.~Wang, Y.~Fung, H.~Peng, and H.~Ji.
\newblock {CRAFT}: Customizing {LLM}s by creating and retrieving from specialized toolsets.
\newblock In \emph{{The Twelfth International Conference on Learning Representations}}, 2024.

\bibitem[Lewis et~al.(2020)Lewis, Perez, Piktus, Petroni, Karpukhin, Goyal, K\"{u}ttler, Lewis, Yih, Rockt\"{a}schel, Riedel, and Kiela]{lewis2020rag}
P.~Lewis, E.~Perez, A.~Piktus, F.~Petroni, V.~Karpukhin, N.~Goyal, H.~K\"{u}ttler, M.~Lewis, W.-t. Yih, T.~Rockt\"{a}schel, S.~Riedel, and D.~Kiela.
\newblock Retrieval-augmented generation for knowledge-intensive {NLP} tasks.
\newblock In \emph{{Proceedings of the 34th International Conference on Neural Information Processing Systems}}, 2020.

\bibitem[Wang et~al.(2023)Wang, Wei, Schuurmans, Le, Chi, Narang, Chowdhery, and Zhou]{wang2023selfconsistency}
X.~Wang, J.~Wei, D.~Schuurmans, Q.~V. Le, E.~H. Chi, S.~Narang, A.~Chowdhery, and D.~Zhou.
\newblock Self-consistency improves chain of thought reasoning in language models.
\newblock In \emph{{The Eleventh International Conference on Learning Representations }}, 2023.

\bibitem[Snell et~al.(2025)Snell, Lee, Xu, and Kumar]{snell2025scaling}
C.~V. Snell, J.~Lee, K.~Xu, and A.~Kumar.
\newblock Scaling {LLM} test-time compute optimally can be more effective than scaling parameters for reasoning.
\newblock In \emph{{The Thirteenth International Conference on Learning Representations}}, 2025.

\bibitem[Hazra et~al.(2025)Hazra, Venturato, Dos~Martires, and De~Raedt]{hazrahave}
R.~Hazra, G.~Venturato, P.~Z. Dos~Martires, and L.~De~Raedt.
\newblock Have large language models learned to reason? a characterization via 3-sat.
\newblock In \emph{{Second Conference on Language Modeling}}, 2025.

\bibitem[Gandhi et~al.(2025)Gandhi, Chakravarthy, Singh, Lile, and Goodman]{gandhi2025cognitive}
K.~Gandhi, A.~K. Chakravarthy, A.~Singh, N.~Lile, and N.~Goodman.
\newblock Cognitive behaviors that enable self-improving reasoners, or, four habits of highly effective {ST}a{R}s.
\newblock In \emph{{Second Conference on Language Modeling}}, 2025.

\bibitem[Chen et~al.(2021)Chen, Tworek, Jun, Yuan, Pinto, Kaplan, Edwards, Burda, Joseph, Brockman, et~al.]{chen2021evaluating}
M.~Chen, J.~Tworek, H.~Jun, Q.~Yuan, H.~P. D.~O. Pinto, J.~Kaplan, H.~Edwards, Y.~Burda, N.~Joseph, G.~Brockman, et~al.
\newblock Evaluating large language models trained on code.
\newblock \emph{arXiv preprint arXiv:2107.03374}, 2021.

\bibitem[Austin et~al.(2021)Austin, Odena, Nye, Bosma, Michalewski, Dohan, Jiang, Cai, Terry, Le, et~al.]{austin2021program}
J.~Austin, A.~Odena, M.~Nye, M.~Bosma, H.~Michalewski, D.~Dohan, E.~Jiang, C.~Cai, M.~Terry, Q.~Le, et~al.
\newblock Program synthesis with large language models.
\newblock \emph{arXiv preprint arXiv:2108.07732}, 2021.

\bibitem[Jimenez et~al.(2024)Jimenez, Yang, Wettig, Yao, Pei, Press, and Narasimhan]{jimenezswe}
C.~E. Jimenez, J.~Yang, A.~Wettig, S.~Yao, K.~Pei, O.~Press, and K.~R. Narasimhan.
\newblock {SWE}-bench: Can language models resolve real-world {G}it{H}ub issues?
\newblock In \emph{{The Twelfth International Conference on Learning Representations}}, 2024.

\bibitem[(){Google DeepMind}]{google-imo}
{Google DeepMind}.
\newblock Advanced version of {G}emini with deep think officially achieves gold-medal standard at the {I}nternational {M}athematical {O}lympiad, 2025.

\bibitem[El-Kishky et~al.(2025)El-Kishky, Wei, Saraiva, Minaiev, Selsam, Dohan, Song, Lightman, Clavera, Pachocki, et~al.]{el2025competitive}
A.~El-Kishky, A.~Wei, A.~Saraiva, B.~Minaiev, D.~Selsam, D.~Dohan, F.~Song, H.~Lightman, I.~Clavera, J.~Pachocki, et~al.
\newblock Competitive programming with large reasoning models.
\newblock \emph{arXiv preprint arXiv:2502.06807}, 2025.

\bibitem[Balunovi{\'c} et~al.(2025)Balunovi{\'c}, Dekoninck, Petrov, Jovanovi{\'c}, and Vechev]{balunovic2025matharena}
M.~Balunovi{\'c}, J.~Dekoninck, I.~Petrov, N.~Jovanovi{\'c}, and M.~Vechev.
\newblock {M}ath{A}rena: Evaluating llms on uncontaminated math competitions.
\newblock \emph{arXiv preprint arXiv:2505.23281}, 2025.

\bibitem[He et~al.(2024)He, Luo, Bai, Hu, Thai, Shen, Hu, Han, Huang, Zhang, et~al.]{he2024olympiadbench}
C.~He, R.~Luo, Y.~Bai, S.~Hu, Z.~Thai, J.~Shen, J.~Hu, X.~Han, Y.~Huang, Y.~Zhang, et~al.
\newblock {O}lympiad{B}ench: A challenging benchmark for promoting {AGI} with {O}lympiad-level bilingual multimodal scientific problems.
\newblock pages, 3828--3850, 2024.

\bibitem[Jain et~al.(2025)Jain, Han, Gu, Li, Yan, Zhang, Wang, Solar-Lezama, Sen, and Stoica]{jainlivecodebench}
N.~Jain, K.~Han, A.~Gu, W.-D. Li, F.~Yan, T.~Zhang, S.~Wang, A.~Solar-Lezama, K.~Sen, and I.~Stoica.
\newblock {L}ive{C}ode{B}ench: Holistic and contamination free evaluation of large language models for code.
\newblock In \emph{{The Thirteenth International Conference on Learning Representations}}, 2025.

\bibitem[Qiu et~al.(2025)Qiu, Guo, Song, Sun, Cai, Wei, Luo, Yin, Zhang, Hu, et~al.]{qiu2025phybench}
S.~Qiu, S.~Guo, Z.-Y. Song, Y.~Sun, Z.~Cai, J.~Wei, T.~Luo, Y.~Yin, H.~Zhang, Y.~Hu, et~al.
\newblock {PHYB}ench: Holistic evaluation of physical perception and reasoning in large language models.
\newblock \emph{arXiv preprint arXiv:2504.16074}, 2025.

\bibitem[Hendrycks et~al.(2021{\natexlab{a}})Hendrycks, Burns, Kadavath, Arora, Basart, Tang, Song, and Steinhardt]{hendrycks2measuring}
D.~Hendrycks, C.~Burns, S.~Kadavath, A.~Arora, S.~Basart, E.~Tang, D.~Song, and J.~Steinhardt.
\newblock Measuring mathematical problem solving with the {MATH} dataset.
\newblock In \emph{{Thirty-fifth Conference on Neural Information Processing Systems Datasets and Benchmarks Track (Round 2)}}, 2021{\natexlab{a}}.

\bibitem[Hendrycks et~al.(2021{\natexlab{b}})Hendrycks, Burns, Basart, Zou, Mazeika, Song, and Steinhardt]{hendrycks2020measuring}
D.~Hendrycks, C.~Burns, S.~Basart, A.~Zou, M.~Mazeika, D.~Song, and J.~Steinhardt.
\newblock Measuring massive multitask language understanding.
\newblock In \emph{{International Conference on Learning Representations}}, 2021{\natexlab{b}}.

\bibitem[Wang et~al.(2024{\natexlab{a}})Wang, Ma, Zhang, Ni, Chandra, Guo, Ren, Arulraj, He, Jiang, et~al.]{wang2024mmlu}
Y.~Wang, X.~Ma, G.~Zhang, Y.~Ni, A.~Chandra, S.~Guo, W.~Ren, A.~Arulraj, X.~He, Z.~Jiang, et~al.
\newblock {MMLU}-{P}ro: A more robust and challenging multi-task language understanding benchmark.
\newblock \emph{Advances in Neural Information Processing Systems}, 37:\penalty0 95266--95290, 2024{\natexlab{a}}.

\bibitem[Wang et~al.(2024{\natexlab{b}})Wang, Hu, Lu, Zhu, Zhang, Subramaniam, Loomba, Zhang, Sun, and Wang]{wang2023scibench}
X.~Wang, Z.~Hu, P.~Lu, Y.~Zhu, J.~Zhang, S.~Subramaniam, A.~R. Loomba, S.~Zhang, Y.~Sun, and W.~Wang.
\newblock {S}ci{B}ench: Evaluating college-level scientific problem-solving abilities of large language models.
\newblock pages, 50622--50649. PMLR, 2024{\natexlab{b}}.

\bibitem[Xu et~al.(2025)Xu, Xu, Xiao, Chen, Yan, ZHANG, Diao, Yang, and Wang]{xu2025ugphysics}
X.~Xu, Q.~Xu, T.~Xiao, T.~Chen, Y.~Yan, J.~ZHANG, S.~Diao, C.~Yang, and Y.~Wang.
\newblock {UGP}hysics: A comprehensive benchmark for undergraduate physics reasoning with large language models.
\newblock In \emph{{Forty-second International Conference on Machine Learning}}, 2025.

\bibitem[Rein et~al.(2024)Rein, Hou, Stickland, Petty, Pang, Dirani, Michael, and Bowman]{rein2024gpqa}
D.~Rein, B.~L. Hou, A.~C. Stickland, J.~Petty, R.~Y. Pang, J.~Dirani, J.~Michael, and S.~R. Bowman.
\newblock {GPQA}: A graduate-level google-proof q\&a benchmark.
\newblock In \emph{{First Conference on Language Modeling}}, 2024.

\bibitem[Tian et~al.(2024)Tian, Gao, Zhang, Chen, Fan, Guo, Haas, Ji, Krongchon, Li, et~al.]{tian2024scicode}
M.~Tian, L.~Gao, S.~Zhang, X.~Chen, C.~Fan, X.~Guo, R.~Haas, P.~Ji, K.~Krongchon, Y.~Li, et~al.
\newblock Scicode: A research coding benchmark curated by scientists.
\newblock \emph{Advances in Neural Information Processing Systems}, 37:\penalty0 30624--30650, 2024.

\bibitem[Glazer et~al.(2024)Glazer, Erdil, Besiroglu, Chicharro, Chen, Gunning, Olsson, Denain, Ho, Santos, et~al.]{glazer2024frontiermath}
E.~Glazer, E.~Erdil, T.~Besiroglu, D.~Chicharro, E.~Chen, A.~Gunning, C.~F. Olsson, J.-S. Denain, A.~Ho, E.~d.~O. Santos, et~al.
\newblock {FrontierMath:} a benchmark for evaluating advanced mathematical reasoning in {AI}.
\newblock \emph{arXiv preprint arXiv:2411.04872}, 2024.

\bibitem[Chung et~al.(2025)Chung, Gao, Kvasiuk, Li, M{\"u}nchmeyer, Rudolph, Sala, and Tadepalli]{chung2025theoretical}
D.~J. Chung, Z.~Gao, Y.~Kvasiuk, T.~Li, M.~M{\"u}nchmeyer, M.~Rudolph, F.~Sala, and S.~C. Tadepalli.
\newblock Theoretical physics benchmark ({TPB}ench)--a dataset and study of {AI} reasoning capabilities in theoretical physics.
\newblock \emph{arXiv preprint arXiv:2502.15815}, 2025.

\bibitem[Phan et~al.(2025)Phan, Gatti, Han, Li, Hu, Zhang, Zhang, Shaaban, Ling, Shi, et~al.]{phan2025humanity}
L.~Phan, A.~Gatti, Z.~Han, N.~Li, J.~Hu, H.~Zhang, C.~B.~C. Zhang, M.~Shaaban, J.~Ling, S.~Shi, et~al.
\newblock Humanity's last exam.
\newblock \emph{arXiv preprint arXiv:2501.14249}, 2025.

\bibitem[Wang et~al.(2023)Wang, Fu, Du, Gao, Huang, Liu, Chandak, Liu, Van~Katwyk, Deac, et~al.]{wang2023scientific}
H.~Wang, T.~Fu, Y.~Du, W.~Gao, K.~Huang, Z.~Liu, P.~Chandak, S.~Liu, P.~Van~Katwyk, A.~Deac, et~al.
\newblock Scientific discovery in the age of artificial intelligence.
\newblock \emph{Nature}, 620\penalty0 (7972):\penalty0 47--60, 2023.

\bibitem[Wu et~al.(2024)Wu, Qiu, Ross, Aky{\"u}rek, Chen, Wang, Kim, Andreas, and Kim]{wu2024reasoning}
Z.~Wu, L.~Qiu, A.~Ross, E.~Aky{\"u}rek, B.~Chen, B.~Wang, N.~Kim, J.~Andreas, and Y.~Kim.
\newblock Reasoning or reciting? exploring the capabilities and limitations of language models through counterfactual tasks.
\newblock pages, 1819--1862, 2024.

\bibitem[Balepur et~al.(2024)Balepur, Ravichander, and Rudinger]{balepur2024artifacts}
N.~Balepur, A.~Ravichander, and R.~Rudinger.
\newblock Artifacts or abduction: How do {LLMs} answer multiple-choice questions without the question?
\newblock pages, 10308--10330, 2024.

\bibitem[Deng et~al.(2024)Deng, Zhao, Tang, Gerstein, and Cohan]{deng2024investigating}
C.~Deng, Y.~Zhao, X.~Tang, M.~Gerstein, and A.~Cohan.
\newblock Investigating data contamination in modern benchmarks for large language models.
\newblock pages, 8698--8711, 2024.

\bibitem[Ott et~al.(2022)Ott, Barbosa-Silva, Blagec, Brauner, and Samwald]{ott2022mapping}
S.~Ott, A.~Barbosa-Silva, K.~Blagec, J.~Brauner, and M.~Samwald.
\newblock Mapping global dynamics of benchmark creation and saturation in artificial intelligence.
\newblock \emph{Nature Communications}, 13\penalty0 (1):\penalty0 6793, 2022.

\bibitem[Li et~al.(2024)Li, Guo, Guerin, and Lin]{li2024open}
Y.~Li, Y.~Guo, F.~Guerin, and C.~Lin.
\newblock An open-source data contamination report for large language models.
\newblock pages, 528--541, 2024.

\bibitem[Balepur et~al.(2025)Balepur, Rudinger, and Boyd-Graber]{balepur-etal-2025-best}
N.~Balepur, R.~Rudinger, and J.~L. Boyd-Graber.
\newblock Which of these best describes multiple choice evaluation with {LLM}s? {A}) forced {B}) flawed {C}) fixable {D}) all of the above.
\newblock pages, 3394--3418. Association for Computational Linguistics, 2025.

\bibitem[Dodge et~al.(2021)Dodge, Sap, Marasovi{\'c}, Agnew, Ilharco, Groeneveld, Mitchell, and Gardner]{dodge2021documenting}
J.~Dodge, M.~Sap, A.~Marasovi{\'c}, W.~Agnew, G.~Ilharco, D.~Groeneveld, M.~Mitchell, and M.~Gardner.
\newblock Documenting large webtext corpora: A case study on the colossal clean crawled corpus.
\newblock pages, 1286--1305, 2021.

\bibitem[()Golchin and Surdeanu]{golchintime}
S.~Golchin and M.~Surdeanu.
\newblock Time travel in {LLM}s: Tracing data contamination in large language models.
\newblock In \emph{{The Twelfth International Conference on Learning Representations}}, 2024.

\bibitem[Roberts et~al.(2023)Roberts, Thakur, Herlihy, White, and Dooley]{roberts2023cutoff}
M.~Roberts, H.~Thakur, C.~Herlihy, C.~White, and S.~Dooley.
\newblock To the cutoff... and beyond? a longitudinal perspective on llm data contamination.
\newblock In \emph{{The Twelfth International Conference on Learning Representations}}, 2023.

\bibitem[Wang et~al.(2024)Wang, Li, Chen, Cai, Zhu, Lin, Cao, Kong, Liu, Liu, et~al.]{wang2024large}
P.~Wang, L.~Li, L.~Chen, Z.~Cai, D.~Zhu, B.~Lin, Y.~Cao, L.~Kong, Q.~Liu, T.~Liu, et~al.
\newblock Large language models are not fair evaluators.
\newblock pages, 9440--9450, 2024.

\bibitem[Ye et~al.(2025)Ye, Wang, Huang, Chen, Zhang, Moniz, Gao, Geyer, Huang, Chen, et~al.]{ye2024justice}
J.~Ye, Y.~Wang, Y.~Huang, D.~Chen, Q.~Zhang, N.~Moniz, T.~Gao, W.~Geyer, C.~Huang, P.-Y. Chen, et~al.
\newblock Justice or prejudice? quantifying biases in {LLM}-as-a-judge.
\newblock In \emph{{International Conference on Learning Representations}}, 2025.

\bibitem[Laskar et~al.(2024)Laskar, Alqahtani, Bari, Rahman, Khan, Khan, Jahan, Bhuiyan, Tan, Parvez, Hoque, Joty, and Huang]{laskar-etal-2024-systematic}
M.~T.~R. Laskar, S.~Alqahtani, M.~S. Bari, M.~Rahman, M.~A.~M. Khan, H.~Khan, I.~Jahan, A.~Bhuiyan, C.~W. Tan, M.~R. Parvez, E.~Hoque, S.~Joty, and J.~Huang.
\newblock A systematic survey and critical review on evaluating large language models: Challenges, limitations, and recommendations.
\newblock pages, 13785--13816. Association for Computational Linguistics, 2024.

\bibitem[Meurer et~al.(2017)Meurer, Smith, Paprocki, \v{C}ert\'{i}k, Kirpichev, Rocklin, Kumar, et~al.]{sympy}
A.~Meurer, C.~P. Smith, M.~Paprocki, O.~\v{C}ert\'{i}k, S.~B. Kirpichev, M.~Rocklin, Kumar, et~al.
\newblock {S}ym{P}y: symbolic computing in {P}ython.
\newblock \emph{PeerJ Computer Science}, 3:\penalty0 e103, 2017.

\bibitem[phy()]{physh}
{PhySH -- Physics Subject Headings}.
\newblock \url{https://physh.org/about}.
\newblock Accessed: August 18, 2025.

\bibitem[(){Artificial Analysis}]{aa_critpt}
{Artificial Analysis}.
\newblock {CritPt Benchmark Leaderboard}.
\newblock \url{https://artificialanalysis.ai/evaluations/critpt}, 2026.
\newblock Accessed: 2026-05-07.

\bibitem[Kalai et~al.(2025)Kalai, Nachum, Vempala, and Zhang]{kalai2025language}
A.~T. Kalai, O.~Nachum, S.~S. Vempala, and E.~Zhang.
\newblock {Why language models hallucinate}.
\newblock \emph{arXiv preprint arXiv:2509.04664}, 2025.

\bibitem[({\natexlab{a}})Gottesman]{Gottesman2016}
D.~Gottesman.
\newblock Quantum fault tolerance in small experiments.
\newblock \emph{arXiv preprint arXiv:1610.03507}, 2016{\natexlab{a}}.

\bibitem[({\natexlab{b}})Vuillot]{Vuillot2017}
C.~Vuillot.
\newblock {Is error detection helpful on IBM 5Q chips?}
\newblock \emph{Quantum Inf. Comput.}, 18\penalty0 (11):\penalty0 0949, 2017{\natexlab{b}}.

\bibitem[Linke et~al.(2017)Linke, Gutierrez, Landsman, Figgatt, Debnath, Brown, and Monroe]{Linke2017}
N.~M. Linke, M.~Gutierrez, K.~A. Landsman, C.~Figgatt, S.~Debnath, K.~R. Brown, and C.~Monroe.
\newblock Fault-tolerant quantum error detection.
\newblock \emph{Sci. Adv.}, 3\penalty0 (10):\penalty0 e1701074, 2017.

\bibitem[()Harper and Flammia]{Harper2019}
R.~Harper and S.~T. Flammia.
\newblock Fault-tolerant logical gates in the {IBM} quantum experience.
\newblock \emph{Phys. Rev. Lett.}, 122:\penalty0 080504, 2019.

\bibitem[Komar et~al.(2014)Komar, Kessler, Bishof, Jiang, S{\o}rensen, Ye, and Lukin]{komar2014quantum}
P.~Komar, E.~M. Kessler, M.~Bishof, L.~Jiang, A.~S. S{\o}rensen, J.~Ye, and M.~D. Lukin.
\newblock A quantum network of clocks.
\newblock \emph{Nature Physics}, 10\penalty0 (8):\penalty0 582--587, 2014.

\bibitem[()Zhang and Zhuang]{zhang2021distributed}
Z.~Zhang and Q.~Zhuang.
\newblock Distributed quantum sensing.
\newblock \emph{Quantum Science and Technology}, 6\penalty0 (4):\penalty0 043001, 2021.

\bibitem[Zang et~al.(2024)Zang, Kolar, Gonzales, Chung, Gray, Kettimuthu, Zhong, and Saleem]{zang2024quantum}
A.~Zang, A.~Kolar, A.~Gonzales, J.~Chung, S.~K. Gray, R.~Kettimuthu, T.~Zhong, and Z.~H. Saleem.
\newblock Quantum advantage in distributed sensing with noisy quantum networks.
\newblock \emph{arXiv preprint arXiv:2409.17089}, 2024.

\bibitem[Zang et~al.(2025)Zang, Zheng, Maurer, Chong, Suchara, and Zhong]{zang2025enhancing}
A.~Zang, T.-X. Zheng, P.~C. Maurer, F.~T. Chong, M.~Suchara, and T.~Zhong.
\newblock Enhancing noisy quantum sensing by {GHZ} state partitioning.
\newblock \emph{arXiv preprint arXiv:2507.02829}, 2025.

\bibitem[({\natexlab{a}})Guth]{PhysRevD.23.347}
A.~H. Guth.
\newblock Inflationary universe: A possible solution to the horizon and flatness problems.
\newblock \emph{Phys. Rev. D}, 23:\penalty0 347--356, 1981{\natexlab{a}}.

\bibitem[({\natexlab{b}})Linde]{LINDE1982389}
A.~Linde.
\newblock A new inflationary universe scenario: A possible solution of the horizon, flatness, homogeneity, isotropy and primordial monopole problems.
\newblock \emph{Physics Letters B}, 108\penalty0 (6):\penalty0 389--393, 1982{\natexlab{b}}.

\bibitem[({\natexlab{c}})Albrecht and Steinhardt]{PhysRevLett.48.1220}
A.~Albrecht and P.~J. Steinhardt.
\newblock Cosmology for grand unified theories with radiatively induced symmetry breaking.
\newblock \emph{Phys. Rev. Lett.}, 48:\penalty0 1220--1223, 1982{\natexlab{c}}.

\bibitem[({\natexlab{d}})Linde]{LINDE1983177}
A.~Linde.
\newblock Chaotic inflation.
\newblock \emph{Physics Letters B}, 129\penalty0 (3):\penalty0 177--181, 1983{\natexlab{d}}.

\bibitem[Freese et~al.(1990)Freese, Frieman, and Olinto]{PhysRevLett.65.3233}
K.~Freese, J.~A. Frieman, and A.~V. Olinto.
\newblock Natural inflation with pseudo {Nambu-Goldstone} bosons.
\newblock \emph{Phys. Rev. Lett.}, 65:\penalty0 3233--3236, 1990.

\bibitem[()Kinney and Mahanthappa]{PhysRevD.52.5529}
W.~H. Kinney and K.~T. Mahanthappa.
\newblock Natural inflation from fermion loops.
\newblock \emph{Phys. Rev. D}, 52:\penalty0 5529--5537, 1995.

\bibitem[Arkani-Hamed et~al.(2003)Arkani-Hamed, Cheng, Creminelli, and Randall]{Nima_Arkani-Hamed_2003}
N.~Arkani-Hamed, H.-C. Cheng, P.~Creminelli, and L.~Randall.
\newblock Pseudonatural inflation.
\newblock \emph{Journal of Cosmology and Astroparticle Physics}, 2003\penalty0 (07):\penalty0 003, 2003.

\bibitem[({\natexlab{a}})Adshead and Wyman]{PhysRevLett.108.261302}
P.~Adshead and M.~Wyman.
\newblock Natural inflation on a steep potential with classical non-abelian gauge fields.
\newblock \emph{Phys. Rev. Lett.}, 108:\penalty0 261302, 2012{\natexlab{a}}.

\bibitem[({\natexlab{b}})Adshead and Wyman]{PhysRevD.86.043530}
P.~Adshead and M.~Wyman.
\newblock Gauge-flation trajectories in chromo-natural inflation.
\newblock \emph{Phys. Rev. D}, 86:\penalty0 043530, 2012{\natexlab{b}}.

\bibitem[Long et~al.(2014)Long, McAllister, and McGuirk]{PhysRevD.90.023501}
C.~Long, L.~McAllister, and P.~McGuirk.
\newblock Aligned natural inflation in string theory.
\newblock \emph{Phys. Rev. D}, 90:\penalty0 023501, 2014.

\bibitem[()Maleknejad]{Maleknejad_2016}
A.~Maleknejad.
\newblock Gravitational leptogenesis in axion inflation with {SU(2)} gauge field.
\newblock \emph{Journal of Cosmology and Astroparticle Physics}, 2016\penalty0 (12):\penalty0 027, 2016.

\bibitem[Fujita et~al.(2022)Fujita, Murai, Obata, and Shiraishi]{Fujita_2022}
T.~Fujita, K.~Murai, I.~Obata, and M.~Shiraishi.
\newblock Gravitational wave trispectrum in the axion-{SU(2)} model.
\newblock \emph{Journal of Cosmology and Astroparticle Physics}, 2022\penalty0 (01):\penalty0 007, 2022.

\bibitem[({\natexlab{a}})Gluscevic and Kamionkowski]{PhysRevD.81.123529}
V.~Gluscevic and M.~Kamionkowski.
\newblock Testing parity-violating mechanisms with cosmic microwave background experiments.
\newblock \emph{Phys. Rev. D}, 81:\penalty0 123529, 2010{\natexlab{a}}.

\bibitem[({\natexlab{b}})Wolf]{PhysRevD.110.043521}
W.~J. Wolf.
\newblock Minimizing the tensor-to-scalar ratio in single-field inflation models.
\newblock \emph{Phys. Rev. D}, 110:\penalty0 043521, 2024{\natexlab{b}}.

\bibitem[Kachru et~al.(2003)Kachru, Kallosh, Linde, Maldacena, McAllister, and Trivedi]{Shamit_Kachru_2003}
S.~Kachru, R.~Kallosh, A.~Linde, J.~Maldacena, L.~McAllister, and S.~P. Trivedi.
\newblock Towards inflation in string theory.
\newblock \emph{Journal of Cosmology and Astroparticle Physics}, 2003\penalty0 (10):\penalty0 013, 2003.

\bibitem[({\natexlab{a}})Witten]{witten1978some}
E.~Witten.
\newblock Some properties of the ($\psi$$\psi$) 2 model in two dimensions.
\newblock \emph{Nuclear Physics B}, 142\penalty0 (3):\penalty0 285--300, 1978{\natexlab{a}}.

\bibitem[({\natexlab{b}})Goldschmidt]{goldschmidt1986kosterlitz}
Y.~Y. Goldschmidt.
\newblock A {Kosterlitz-Thouless} phase transition associated with the supersymmetric sine-gordon theory.
\newblock \emph{Nuclear Physics B}, 270:\penalty0 29--38, 1986{\natexlab{b}}.

\bibitem[({\natexlab{c}})Moore and Read]{moore1991nonabelions}
G.~Moore and N.~Read.
\newblock {Nonabelions in the fractional quantum Hall effect}.
\newblock \emph{Nuclear Physics B}, 360\penalty0 (2-3):\penalty0 362--396, 1991{\natexlab{c}}.

\bibitem[({\natexlab{d}})Milovanovi{\'c} and Read]{milovanovic1996edge}
M.~Milovanovi{\'c} and N.~Read.
\newblock Edge excitations of paired fractional quantum {Hall} states.
\newblock \emph{Physical Review B}, 53\penalty0 (20):\penalty0 13559, 1996{\natexlab{d}}.

\bibitem[Schiller et~al.(2023)Schiller, Katzir, Stern, Berg, Lindner, and Oreg]{schiller2023superconductivity}
N.~Schiller, B.~A. Katzir, A.~Stern, E.~Berg, N.~H. Lindner, and Y.~Oreg.
\newblock {Superconductivity and fermionic dissipation in quantum Hall edges}.
\newblock \emph{Physical Review B}, 107\penalty0 (16):\penalty0 L161105, 2023.

\bibitem[Cao et~al.(2024)Cao, Kou, and Fradkin]{cao2024signatures}
J.~Cao, A.~Kou, and E.~Fradkin.
\newblock Signatures of parafermion zero modes in fractional quantum {Hall}--superconductor heterostructures.
\newblock \emph{Physical Review B}, 109\penalty0 (16):\penalty0 L161106, 2024.

\bibitem[May-Mann et~al.(2024)May-Mann, Stern, and Devakul]{may2024theory}
J.~May-Mann, A.~Stern, and T.~Devakul.
\newblock Theory of half-integer fractional quantum spin hall insulator edges.
\newblock \emph{arXiv preprint arXiv:2403.03964}, 2024.

\bibitem[()Kitaev]{kitaev2006anyons}
A.~Kitaev.
\newblock Anyons in an exactly solved model and beyond.
\newblock \emph{Annals of Physics}, 321\penalty0 (1):\penalty0 2--111, 2006.

\bibitem[Nayak et~al.(2008)Nayak, Simon, Stern, Freedman, and Das~Sarma]{nayak2008non}
C.~Nayak, S.~H. Simon, A.~Stern, M.~Freedman, and S.~Das~Sarma.
\newblock Non-abelian anyons and topological quantum computation.
\newblock \emph{Reviews of Modern Physics}, 80\penalty0 (3):\penalty0 1083--1159, 2008.

\bibitem[Hastings et~al.(2013)Hastings, Nayak, and Wang]{hastings2013metaplectic}
M.~B. Hastings, C.~Nayak, and Z.~Wang.
\newblock Metaplectic anyons, majorana zero modes, and their computational power.
\newblock \emph{Physical Review B—Condensed Matter and Materials Physics}, 87\penalty0 (16):\penalty0 165421, 2013.

\bibitem[Clarke et~al.(2013)Clarke, Alicea, and Shtengel]{clarke2013exotic}
D.~J. Clarke, J.~Alicea, and K.~Shtengel.
\newblock Exotic non-abelian anyons from conventional fractional quantum {Hall} states.
\newblock \emph{Nature communications}, 4\penalty0 (1):\penalty0 1348, 2013.

\bibitem[Lindner et~al.(2012)Lindner, Berg, Refael, and Stern]{lindner2012fractionalizing}
N.~H. Lindner, E.~Berg, G.~Refael, and A.~Stern.
\newblock Fractionalizing {M}ajorana fermions: Non-abelian statistics on the edges of abelian quantum {Hall} states.
\newblock \emph{Physical Review X}, 2\penalty0 (4):\penalty0 041002, 2012.

\bibitem[()Cheng]{cheng2012superconducting}
M.~Cheng.
\newblock Superconducting proximity effect on the edge of fractional topological insulators.
\newblock \emph{Physical Review B—Condensed Matter and Materials Physics}, 86\penalty0 (19):\penalty0 195126, 2012.

\bibitem[Barkeshli et~al.(2013)Barkeshli, Jian, and Qi]{barkeshli2013theory}
M.~Barkeshli, C.-M. Jian, and X.-L. Qi.
\newblock Theory of defects in abelian topological states.
\newblock \emph{Physical Review B—Condensed Matter and Materials Physics}, 88\penalty0 (23):\penalty0 235103, 2013.

\bibitem[({\natexlab{a}})Fendley]{fendley2012parafermionic}
P.~Fendley.
\newblock Parafermionic edge zero modes in {Zn}-invariant spin chains.
\newblock \emph{Journal of Statistical Mechanics: Theory and Experiment}, 2012\penalty0 (11):\penalty0 P11020, 2012{\natexlab{a}}.

\bibitem[({\natexlab{b}})Verlinde]{verlinde1988fusion}
E.~Verlinde.
\newblock Fusion rules and modular transformations in 2d conformal field theory.
\newblock \emph{Nuclear Physics B}, 300:\penalty0 360--376, 1988{\natexlab{b}}.

\bibitem[Fr{\"o}hlich et~al.(2004)Fr{\"o}hlich, Fuchs, Runkel, and Schweigert]{frohlich2004kramers}
J.~Fr{\"o}hlich, J.~Fuchs, I.~Runkel, and C.~Schweigert.
\newblock {Kramers-Wannier} duality from conformal defects.
\newblock \emph{Physical review letters}, 93\penalty0 (7):\penalty0 070601, 2004.

\bibitem[Fr{\"o}hlich et~al.(2007)Fr{\"o}hlich, Fuchs, Runkel, and Schweigert]{frohlich2007duality}
J.~Fr{\"o}hlich, J.~Fuchs, I.~Runkel, and C.~Schweigert.
\newblock Duality and defects in rational conformal field theory.
\newblock \emph{Nuclear Physics B}, 763\penalty0 (3):\penalty0 354--430, 2007.

\bibitem[Francesco et~al.(2012)Francesco, Mathieu, and S{\'e}n{\'e}chal]{francesco2012conformal}
P.~Francesco, P.~Mathieu, and D.~S{\'e}n{\'e}chal.
\newblock \emph{Conformal field theory}.
\newblock Springer Science \& Business Media, 2012.

\bibitem[Chang et~al.(2019)Chang, Lin, Shao, Wang, and Yin]{chang2019topological}
C.-M. Chang, Y.-H. Lin, S.-H. Shao, Y.~Wang, and X.~Yin.
\newblock Topological defect lines and renormalization group flows in two dimensions.
\newblock \emph{Journal of High Energy Physics}, 2019\penalty0 (1):\penalty0 1--85, 2019.

\bibitem[Huang et~al.(2024)Huang, Colmenarez, M{\"u}ller, and Diehl]{huang2024arxiv}
Z.-M. Huang, L.~Colmenarez, M.~M{\"u}ller, and S.~Diehl.
\newblock Coherent information as a mixed-state topological order parameter of fermions.
\newblock \emph{arXiv preprint arXiv:2412.12279}, 2024.

\bibitem[Fan et~al.(2024)Fan, Bao, Altman, and Vishwanath]{fan2024diagnostics}
R.~Fan, Y.~Bao, E.~Altman, and A.~Vishwanath.
\newblock Diagnostics of mixed-state topological order and breakdown of quantum memory.
\newblock \emph{PRX Quantum}, 5\penalty0 (2):\penalty0 020343, 2024.

\bibitem[Dennis et~al.(2002)Dennis, Kitaev, Landahl, and Preskill]{dennis2002topological}
E.~Dennis, A.~Kitaev, A.~Landahl, and J.~Preskill.
\newblock Topological quantum memory.
\newblock \emph{Journal of Mathematical Physics}, 43\penalty0 (9):\penalty0 4452--4505, 2002.

\bibitem[()Nishimori]{nishimori2001statistical}
H.~Nishimori.
\newblock Number.
\newblock 111. Clarendon Press, 2001.

\bibitem[Zaletel et~al.(2014)Zaletel, Mong, and Pollmann]{zaletel2014arxiv}
M.~P. Zaletel, R.~S. Mong, and F.~Pollmann.
\newblock Flux insertion, entanglement, and quantized responses.
\newblock \emph{Journal of Statistical Mechanics: Theory and Experiment}, 2014\penalty0 (10):\penalty0 P10007, 2014.

\bibitem[Cirac et~al.(2021)Cirac, Perez-Garcia, Schuch, and Verstraete]{cirac2021rmp}
J.~I. Cirac, D.~Perez-Garcia, N.~Schuch, and F.~Verstraete.
\newblock Matrix product states and projected entangled pair states: Concepts, symmetries, theorems.
\newblock \emph{Reviews of Modern Physics}, 93\penalty0 (4):\penalty0 045003, 2021.

\bibitem[Huang et~al.(2025)Huang, Diehl, and Sun]{huang2025topological}
Z.-M. Huang, S.~Diehl, and X.-Q. Sun.
\newblock Topological response in open quantum systems with weak symmetries.
\newblock \emph{arXiv preprint arXiv:2504.02941}, 2025.

\bibitem[Sun et~al.(2011)Sun, Gu, Katsura, and Das~Sarma]{sun2011nearly}
K.~Sun, Z.~Gu, H.~Katsura, and S.~Das~Sarma.
\newblock Nearly flatbands with nontrivial topology.
\newblock \emph{Physical review letters}, 106\penalty0 (23):\penalty0 236803, 2011.

\bibitem[Neupert et~al.(2011)Neupert, Santos, Chamon, and Mudry]{neupert2011fractional}
T.~Neupert, L.~Santos, C.~Chamon, and C.~Mudry.
\newblock {Fractional quantum Hall states at zero magnetic field}.
\newblock \emph{Physical review letters}, 106\penalty0 (23):\penalty0 236804, 2011.

\bibitem[Mai et~al.(2023{\natexlab{a}})Mai, Zhao, Feldman, and Phillips]{mai20231}
P.~Mai, J.~Zhao, B.~E. Feldman, and P.~W. Phillips.
\newblock 1/4 is the new 1/2 when topology is intertwined with {Mottness}.
\newblock \emph{Nature communications}, 14\penalty0 (1):\penalty0 5999, 2023{\natexlab{a}}.

\bibitem[Mai et~al.(2023{\natexlab{b}})Mai, Feldman, and Phillips]{mai2023topological}
P.~Mai, B.~E. Feldman, and P.~W. Phillips.
\newblock Topological {M}ott insulator at quarter filling in the interacting haldane model.
\newblock \emph{Physical Review Research}, 5\penalty0 (1):\penalty0 013162, 2023{\natexlab{b}}.

\bibitem[Mai et~al.(2024)Mai, Zhao, Maier, Bradlyn, and Phillips]{mai2024topological}
P.~Mai, J.~Zhao, T.~A. Maier, B.~Bradlyn, and P.~W. Phillips.
\newblock Topological phase transition without single particle gap closing in strongly correlated systems.
\newblock \emph{Physical Review B}, 110\penalty0 (7):\penalty0 075105, 2024.

\bibitem[({\natexlab{a}})Fu and Sachdev]{PhysRevB.94.035135}
W.~Fu and S.~Sachdev.
\newblock Numerical study of fermion and boson models with infinite-range random interactions.
\newblock \emph{Phys. Rev. B}, 94:\penalty0 035135, 2016{\natexlab{a}}.

\bibitem[({\natexlab{b}})Maldacena and Stanford]{PhysRevD.94.106002}
J.~Maldacena and D.~Stanford.
\newblock Remarks on the {Sachdev-Ye-Kitaev} model.
\newblock \emph{Phys. Rev. D}, 94:\penalty0 106002, 2016{\natexlab{b}}.

\bibitem[Izubuchi et~al.(2018)Izubuchi, Ji, Jin, Stewart, and Zhao]{Izubuchi:2018srq}
T.~Izubuchi, X.~Ji, L.~Jin, I.~W. Stewart, and Y.~Zhao.
\newblock Factorization theorem relating {Euclidean} and light-cone parton distributions.
\newblock \emph{Phys. Rev. D}, 98\penalty0 (5):\penalty0 056004, 2018.

\bibitem[Moch et~al.(2004)Moch, Vermaseren, and Vogt]{Moch:2004pa}
S.~Moch, J.~A.~M. Vermaseren, and A.~Vogt.
\newblock The three loop splitting functions in {QCD}: The nonsinglet case.
\newblock \emph{Nucl. Phys. B}, 688:\penalty0 101--134, 2004.

\bibitem[Su et~al.(2023)Su, Holligan, Ji, Yao, Zhang, and Zhang]{Su:2022fiu}
Y.~Su, J.~Holligan, X.~Ji, F.~Yao, J.-H. Zhang, and R.~Zhang.
\newblock Resumming quark's longitudinal momentum logarithms in {LaMET} expansion of lattice {PDFs}.
\newblock \emph{Nucl. Phys. B}, 991:\penalty0 116201, 2023.

\bibitem[Ji et~al.(2021)Ji, Liu, Liu, Zhang, and Zhao]{ji2021large}
X.~Ji, Y.~Liu, Y.-S. Liu, J.-H. Zhang, and Y.~Zhao.
\newblock Large-momentum effective theory.
\newblock \emph{Reviews of Modern Physics}, 93\penalty0 (3):\penalty0 035005, 2021.

\bibitem[({\natexlab{a}})Ji]{ji2013parton}
X.~Ji.
\newblock Parton physics on a euclidean lattice.
\newblock \emph{Physical Review Letters}, 110\penalty0 (26):\penalty0 262002, 2013{\natexlab{a}}.

\bibitem[({\natexlab{b}})Ji]{Ji:2014gla}
X.~Ji.
\newblock Parton physics from large-momentum effective field theory.
\newblock \emph{Sci. China Phys. Mech. Astron.}, 57:\penalty0 1407--1412, 2014{\natexlab{b}}.

\bibitem[Horodecki et~al.(2009)Horodecki, Horodecki, Horodecki, and Oppenheim]{horodecki2009general}
K.~Horodecki, M.~Horodecki, P.~Horodecki, and J.~Oppenheim.
\newblock General paradigm for distilling classical key from quantum states.
\newblock \emph{IEEE Transactions on Information Theory}, 55\penalty0 (4):\penalty0 1898--1929, 2009.

\bibitem[({\natexlab{a}})Smith and Wu]{smith2025additivity}
G.~Smith and P.~Wu.
\newblock Additivity of quantum capacities in simple non-degradable quantum channels.
\newblock \emph{IEEE Transactions on Information Theory}, 2025{\natexlab{a}}.

\bibitem[({\natexlab{b}})Lesniewski and Ruskai]{lesniewski1999monotone}
A.~Lesniewski and M.~B. Ruskai.
\newblock Monotone {R}iemannian metrics and relative entropy on noncommutative probability spaces.
\newblock \emph{Journal of Mathematical Physics}, 40\penalty0 (11):\penalty0 5702--5724, 1999{\natexlab{b}}.

\bibitem[({\natexlab{c}})Hiai and Ruskai]{hiai2016contraction}
F.~Hiai and M.~B. Ruskai.
\newblock Contraction coefficients for noisy quantum channels.
\newblock \emph{Journal of Mathematical Physics}, 57\penalty0 (1), 2016{\natexlab{c}}.

\bibitem[Hoang et~al.(2016)Hoang, Ma, Ahn, Bang, Robicheaux, Yin, and Li]{hoang2016torsional}
T.~M. Hoang, Y.~Ma, J.~Ahn, J.~Bang, F.~Robicheaux, Z.-Q. Yin, and T.~Li.
\newblock Torsional optomechanics of a levitated nonspherical nanoparticle.
\newblock \emph{Physical review letters}, 117\penalty0 (12):\penalty0 123604, 2016.

\bibitem[Zhang et~al.(2025)Zhang, Zhang, and Yin]{zhang2025scalable}
G.~Zhang, H.~Zhang, and Z.-q. Yin.
\newblock Scalable universal quantum gates between nitrogen-vacancy centers in levitated nanodiamonds arrays.
\newblock \emph{arXiv preprint arXiv:2504.08194}, 2025.

\bibitem[Rieser et~al.(2022)Rieser, Ciampini, Rudolph, Kiesel, Hornberger, Stickler, Aspelmeyer, and Deli{\'c}]{rieser2022tunable}
J.~Rieser, M.~A. Ciampini, H.~Rudolph, N.~Kiesel, K.~Hornberger, B.~A. Stickler, M.~Aspelmeyer, and U.~Deli{\'c}.
\newblock Tunable light-induced dipole-dipole interaction between optically levitated nanoparticles.
\newblock \emph{Science}, 377\penalty0 (6609):\penalty0 987--990, 2022.

\bibitem[Manceau et~al.(2017)Manceau, Khalili, and Chekhova]{manceau2017improving}
M.~Manceau, F.~Khalili, and M.~Chekhova.
\newblock Improving the phase super-sensitivity of squeezing-assisted interferometers by squeeze factor unbalancing.
\newblock \emph{New Journal of Physics}, 19\penalty0 (1):\penalty0 013014, 2017.

\bibitem[Nehra et~al.(2022)Nehra, Sekine, Ledezma, Guo, Gray, Roy, and Marandi]{nehra2022few}
R.~Nehra, R.~Sekine, L.~Ledezma, Q.~Guo, R.~M. Gray, A.~Roy, and A.~Marandi.
\newblock Few-cycle vacuum squeezing in nanophotonics.
\newblock \emph{Science}, 377\penalty0 (6612):\penalty0 1333--1337, 2022.

\bibitem[Murase et~al.(2018)Murase, Oikonomou, and Petropoulou]{Murase:2018iyl}
K.~Murase, F.~Oikonomou, and M.~Petropoulou.
\newblock Blazar flares as an origin of high-energy cosmic neutrinos?
\newblock \emph{Astrophys. J.}, 865\penalty0 (2):\penalty0 124, 2018.

\bibitem[Padovani et~al.(2019)Padovani, Oikonomou, Petropoulou, Giommi, and Resconi]{Padovani:2019xcv}
P.~Padovani, F.~Oikonomou, M.~Petropoulou, P.~Giommi, and E.~Resconi.
\newblock {TXS} 0506+056, the first cosmic neutrino source, is not a {BL Lac}.
\newblock \emph{Mon. Not. Roy. Astron. Soc.}, 484\penalty0 (1):\penalty0 L104--L108, 2019.

\bibitem[({\natexlab{a}})Boyer]{boyer1968quantum}
T.~H. Boyer.
\newblock Quantum electromagnetic zero-point energy of a conducting spherical shell and the casimir model for a charged particle.
\newblock \emph{Physical Review}, 174\penalty0 (5):\penalty0 1764, 1968{\natexlab{a}}.

\bibitem[({\natexlab{b}})Brown and Gabrielse]{brown1986geonium}
L.~S. Brown and G.~Gabrielse.
\newblock Geonium theory: Physics of a single electron or ion in a {Penning} trap.
\newblock \emph{Reviews of Modern Physics}, 58\penalty0 (1):\penalty0 233, 1986{\natexlab{b}}.

\bibitem[Brown et~al.(1986)Brown, Helmerson, and Tan]{brown1986cyclotron}
L.~S. Brown, K.~Helmerson, and J.~Tan.
\newblock Cyclotron motion in a spherical microwave cavity.
\newblock \emph{Physical Review A}, 34\penalty0 (4):\penalty0 2638, 1986.

\bibitem[()Barton and Fawcett]{barton1988quantum}
G.~Barton and N.~S. Fawcett.
\newblock {Quantum electromagnetics of an electron near mirrors}.
\newblock \emph{Physics Reports}, 170\penalty0 (1):\penalty0 1--95, 1988.

\bibitem[Fan et~al.(2023)Fan, Myers, Sukra, and Gabrielse]{fan2023measurement}
X.~Fan, T.~Myers, B.~Sukra, and G.~Gabrielse.
\newblock Measurement of the electron magnetic moment.
\newblock \emph{Physical review letters}, 130\penalty0 (7):\penalty0 071801, 2023.

\bibitem[()Kitagawa and Ueda]{Kitagawa1993squeezed}
M.~Kitagawa and M.~Ueda.
\newblock Squeezed spin states.
\newblock \emph{Physical Review A}, 47\penalty0 (6):\penalty0 5138, 1993.

\bibitem[Wineland et~al.(1994)Wineland, Bollinger, Itano, and Heinzen]{wineland1994squeezed}
D.~J. Wineland, J.~J. Bollinger, W.~M. Itano, and D.~J. Heinzen.
\newblock Squeezed atomic states and projection noise in spectroscopy.
\newblock \emph{Physical Review A}, 50\penalty0 (1):\penalty0 67, 1994.

\bibitem[Ma et~al.(2011)Ma, Wang, and Nori]{ma2011quantum}
J.~Ma, X.~Wang, and F.~Nori.
\newblock Quantum spin squeezing.
\newblock \emph{Physics Reports}, 509\penalty0 (2-3):\penalty0 89--165, 2011.

\bibitem[Chu et~al.(2021)Chu, He, Thompson, and Rey]{chu2021quantum}
A.~Chu, P.~He, J.~K. Thompson, and A.~M. Rey.
\newblock Quantum enhanced cavity {QED} interferometer with partially delocalized atoms in lattices.
\newblock \emph{Physical Review Letters}, 127\penalty0 (21):\penalty0 210401, 2021.

\bibitem[Barberena et~al.(2024)Barberena, Chu, Thompson, and Rey]{barberena2024trade}
D.~Barberena, A.~Chu, J.~K. Thompson, and A.~M. Rey.
\newblock Trade-offs between unitary and measurement induced spin squeezing in cavity {QED}.
\newblock \emph{Physical Review Research}, 6\penalty0 (3):\penalty0 L032037, 2024.

\bibitem[()Goluskin]{goluskin2016internally}
D.~Goluskin.
\newblock \emph{Internally heated convection and {Rayleigh-B{\'e}nard} convection}.
\newblock Springer, 2016.

\bibitem[Liu et~al.(2024)Liu, Sharma, Julien, and Knobloch]{liu2024fixed}
C.~Liu, M.~Sharma, K.~Julien, and E.~Knobloch.
\newblock Fixed-flux {Rayleigh--B{\'e}nard} convection in doubly periodic domains: generation of large-scale shear.
\newblock \emph{Journal of Fluid Mechanics}, 979:\penalty0 A19, 2024.

\bibitem[({\natexlab{a}})Chandrasekhar]{chandrasekhar2013hydrodynamic}
S.~Chandrasekhar.
\newblock \emph{Hydrodynamic and hydromagnetic stability}.
\newblock Courier Corporation, 2013{\natexlab{a}}.

\bibitem[({\natexlab{b}})Busse]{busse1967stability}
F.~H. Busse.
\newblock On the stability of two-dimensional convection in a layer heated from below.
\newblock \emph{Journal of Mathematics and Physics}, 46\penalty0 (1-4):\penalty0 140--150, 1967{\natexlab{b}}.

\bibitem[({\natexlab{c}})Manneville]{Manneville2006}
P.~Manneville.
\newblock {Rayleigh-B{\'e}nard} convection: Thirty years of experimental, theoretical, and modeling work.
\newblock pages, 41--65. Springer New York, New York, NY, 2006{\natexlab{c}}.

\bibitem[({\natexlab{d}})Liu and Knobloch]{liu2022single}
C.~Liu and E.~Knobloch.
\newblock Single-mode solutions for convection and double-diffusive convection in porous media.
\newblock \emph{Fluids}, 7\penalty0 (12):\penalty0 373, 2022{\natexlab{d}}.

\bibitem[({\natexlab{e}})Busse]{busse1978non}
F.~H. Busse.
\newblock Non-linear properties of thermal convection.
\newblock \emph{Reports on Progress in Physics}, 41\penalty0 (12):\penalty0 1929, 1978{\natexlab{e}}.

\bibitem[({\natexlab{f}})Nield and Bejan]{nield2006convection}
D.~A. Nield and A.~Bejan.
\newblock \emph{Convection in porous media}.
\newblock Springer, 2006{\natexlab{f}}.

\bibitem[({\natexlab{g}})Nield and Simmons]{nield2019brief}
D.~Nield and C.~T. Simmons.
\newblock A brief introduction to convection in porous media.
\newblock \emph{Transport in Porous Media}, 130\penalty0 (1):\penalty0 237--250, 2019{\natexlab{g}}.

\bibitem[({\natexlab{h}})Trevisan and Bejan]{trevisan1987mass}
O.~V. Trevisan and A.~Bejan.
\newblock Mass and heat transfer by high rayleigh number convection in a porous medium heated from below.
\newblock \emph{International Journal of Heat and Mass Transfer}, 30\penalty0 (11):\penalty0 2341--2356, 1987{\natexlab{h}}.

\bibitem[({\natexlab{i}})Hewitt]{hewitt2020vigorous}
D.~Hewitt.
\newblock Vigorous convection in porous media.
\newblock \emph{Proceedings of the Royal Society A}, 476\penalty0 (2239):\penalty0 20200111, 2020{\natexlab{i}}.

\bibitem[Beacom et~al.(2007)Beacom, Bell, and Mack]{Beacom:2006tt}
J.~F. Beacom, N.~F. Bell, and G.~D. Mack.
\newblock General upper bound on the dark matter total annihilation cross section.
\newblock \emph{Phys. Rev. Lett.}, 99:\penalty0 231301, 2007.

\bibitem[Alenezi et~al.(2025)Alenezi, Cesarotti, Gori, and Shelton]{Alenezi:2025kwl}
A.~Alenezi, C.~Cesarotti, S.~Gori, and J.~Shelton.
\newblock Discovery prospects for a minimal dark matter model at cosmic and intensity frontier experiments.
\newblock 2025.

\bibitem[Bartolotta et~al.(2022)Bartolotta, J{\"a}ger, Reilly, Norcia, Thompson, Smith, and Holland]{bartolotta2022entropy}
J.~P. Bartolotta, S.~B. J{\"a}ger, J.~T. Reilly, M.~A. Norcia, J.~K. Thompson, G.~Smith, and M.~J. Holland.
\newblock Entropy transfer from a quantum particle to a classical coherent light field.
\newblock \emph{Physical Review Research}, 4\penalty0 (1):\penalty0 013218, 2022.

\bibitem[({\natexlab{a}})Bellman and Lehman]{bellman_reciprocity_1961}
R.~Bellman and R.~S. Lehman.
\newblock The reciprocity formula for multidimensional theta functions.
\newblock \emph{Proceedings of the American Mathematical Society}, 12\penalty0 (6):\penalty0 954--961, 1961{\natexlab{a}}.

\bibitem[({\natexlab{b}})Jeffrey and Zwillinger]{jeffrey_table_2007}
A.~Jeffrey and D.~Zwillinger.
\newblock \emph{Table of integrals, series, and products}.
\newblock Academic Press, 2007{\natexlab{b}}.

\bibitem[Turner et~al.(2018)Turner, Michailidis, Abanin, Serbyn, and Papi{\'c}]{turner_quantum_2018}
C.~J. Turner, A.~A. Michailidis, D.~A. Abanin, M.~Serbyn, and Z.~Papi{\'c}.
\newblock Quantum scarred eigenstates in a {{Rydberg}} atom chain: Entanglement, breakdown of thermalization, and stability to perturbations.
\newblock \emph{Physical Review B}, 98\penalty0 (15), 2018.

\bibitem[Zhou et~al.(2023)Zhou, Guo, Xu, Chen, and Swingle]{zhou_hydrodynamic_2023}
T.~Zhou, A.~Y. Guo, S.~Xu, X.~Chen, and B.~Swingle.
\newblock Hydrodynamic theory of scrambling in chaotic long-range interacting systems.
\newblock \emph{Physical Review B}, 107\penalty0 (1):\penalty0 014201, 2023.

\bibitem[({\natexlab{a}})Hallatschek and Fisher]{hallatschek_acceleration_2014}
O.~Hallatschek and D.~S. Fisher.
\newblock Acceleration of evolutionary spread by long-range dispersal.
\newblock \emph{Proceedings of the National Academy of Sciences}, 111\penalty0 (46):\penalty0 E4911--E4919, 2014{\natexlab{a}}.

\bibitem[({\natexlab{b}})Zhou and Nahum]{zhou_emergent_2019}
T.~Zhou and A.~Nahum.
\newblock Emergent statistical mechanics of entanglement in random unitary circuits.
\newblock \emph{Physical Review B}, 99\penalty0 (17):\penalty0 174205, 2019{\natexlab{b}}.

\bibitem[({\natexlab{c}})Chen and Kotlarchyk]{chen2007interactions}
S.-H. Chen and M.~Kotlarchyk.
\newblock \emph{Interactions of photons and neutrons with matter}.
\newblock World Scientific, 2007{\natexlab{c}}.

\bibitem[Tsang et~al.(2016)Tsang, Nair, and Lu]{tsang2016quantum}
M.~Tsang, R.~Nair, and X.-M. Lu.
\newblock Quantum theory of superresolution for two incoherent optical point sources.
\newblock \emph{Physical Review X}, 6\penalty0 (3):\penalty0 031033, 2016.

\bibitem[({\natexlab{a}})Meyer and Wong]{meyer2015connectivity}
D.~A. Meyer and T.~G. Wong.
\newblock Connectivity is a poor indicator of fast quantum search.
\newblock \emph{Physical review letters}, 114\penalty0 (11):\penalty0 110503, 2015{\natexlab{a}}.

\bibitem[({\natexlab{b}})Collins and {\'S}niady]{collins2006integration}
B.~Collins and P.~{\'S}niady.
\newblock Integration with respect to the {Haar} measure on unitary, orthogonal and symplectic group.
\newblock \emph{Communications in Mathematical Physics}, 264\penalty0 (3):\penalty0 773--795, 2006{\natexlab{b}}.

\bibitem[Reilly et~al.(2024)Reilly, J{\"a}ger, Wilson, Cooper, Eggert, and Holland]{reilly2024speeding}
J.~T. Reilly, S.~B. J{\"a}ger, J.~D. Wilson, J.~Cooper, S.~Eggert, and M.~J. Holland.
\newblock Speeding up squeezing with a periodically driven dicke model.
\newblock \emph{Physical Review Research}, 6\penalty0 (3):\penalty0 033090, 2024.

\bibitem[Wilson et~al.(2024)Wilson, Reilly, Zhang, Luo, Chu, Thompson, Rey, and Holland]{wilson2024entangled}
J.~D. Wilson, J.~T. Reilly, H.~Zhang, C.~Luo, A.~Chu, J.~K. Thompson, A.~M. Rey, and M.~J. Holland.
\newblock Entangled matter waves for quantum enhanced sensing.
\newblock \emph{Physical Review A}, 110\penalty0 (4):\penalty0 L041301, 2024.

\bibitem[J{\"a}ger et~al.(2022)J{\"a}ger, Schmit, Morigi, Holland, and Betzholz]{jager2022lindblad}
S.~B. J{\"a}ger, T.~Schmit, G.~Morigi, M.~J. Holland, and R.~Betzholz.
\newblock Lindblad master equations for quantum systems coupled to dissipative bosonic modes.
\newblock \emph{Physical Review Letters}, 129\penalty0 (6):\penalty0 063601, 2022.

\bibitem[Zhang et~al.(2022)Zhang, Chen, and Chitambar]{zhang2022building}
Y.~Zhang, X.~Chen, and E.~Chitambar.
\newblock Building multiple access channels with a single particle.
\newblock \emph{Quantum}, 6:\penalty0 653, 2022.

\bibitem[()Horvat and Daki{\'c}]{horvat2021quantum}
S.~Horvat and B.~Daki{\'c}.
\newblock Quantum enhancement to information acquisition speed.
\newblock \emph{New Journal of Physics}, 23\penalty0 (3):\penalty0 033008, 2021.

\bibitem[Mu et~al.(2024)Mu, Sun, and Millis]{PhysRevB.109.115154}
A.~Mu, Z.~Sun, and A.~J. Millis.
\newblock Adequacy of the dynamical mean field theory for low density and dirac materials.
\newblock \emph{Phys. Rev. B}, 109:\penalty0 115154, 2024.

\bibitem[Mu et~al.(2022)Mu, Sun, and Millis]{PhysRevB.106.085142}
A.~Mu, Z.~Sun, and A.~J. Millis.
\newblock Optical conductivity of the two-dimensional hubbard model: Vertex corrections, emergent galilean invariance, and the accuracy of the single-site dynamical mean field approximation.
\newblock \emph{Phys. Rev. B}, 106:\penalty0 085142, 2022.

\bibitem[()Rosch and Howell]{PhysRevB.72.104510}
A.~Rosch and P.~C. Howell.
\newblock Zero-temperature optical conductivity of ultraclean fermi liquids and superconductors.
\newblock \emph{Phys. Rev. B}, 72:\penalty0 104510, 2005.

\bibitem[Greiner et~al.(2002)Greiner, Mandel, Esslinger, H{\"a}nsch, and Bloch]{greiner2002quantum}
M.~Greiner, O.~Mandel, T.~Esslinger, T.~W. H{\"a}nsch, and I.~Bloch.
\newblock Quantum phase transition from a superfluid to a {Mott} insulator in a gas of ultracold atoms.
\newblock \emph{nature}, 415\penalty0 (6867):\penalty0 39--44, 2002.

\bibitem[Bloch et~al.(2012)Bloch, Dalibard, and Nascimbene]{bloch2012quantum}
I.~Bloch, J.~Dalibard, and S.~Nascimbene.
\newblock Quantum simulations with ultracold quantum gases.
\newblock \emph{Nature Physics}, 8\penalty0 (4):\penalty0 267--276, 2012.

\bibitem[({\natexlab{a}})Gross and Bloch]{gross2017quantum}
C.~Gross and I.~Bloch.
\newblock Quantum simulations with ultracold atoms in optical lattices.
\newblock \emph{Science}, 357\penalty0 (6355):\penalty0 995--1001, 2017{\natexlab{a}}.

\bibitem[({\natexlab{b}})Gross and Bakr]{gross2021quantum}
C.~Gross and W.~S. Bakr.
\newblock Quantum gas microscopy for single atom and spin detection.
\newblock \emph{Nature Physics}, 17\penalty0 (12):\penalty0 1316--1323, 2021{\natexlab{b}}.

\bibitem[Young et~al.(2022)Young, Eckner, Schine, Childs, and Kaufman]{young2022tweezer}
A.~W. Young, W.~J. Eckner, N.~Schine, A.~M. Childs, and A.~M. Kaufman.
\newblock Tweezer-programmable {2D} quantum walks in a {Hubbard}-regime lattice.
\newblock \emph{Science}, 377\penalty0 (6608):\penalty0 885--889, 2022.

\bibitem[()Kim]{kim2017}
I.~H. Kim.
\newblock Holographic quantum simulation.
\newblock \emph{arXiv:1702.02093}, 2017.

\bibitem[Foss-Feig et~al.(2021)Foss-Feig, Hayes, Dreiling, Figgatt, Gaebler, Moses, Pino, and Potter]{fossfeig2020}
M.~Foss-Feig, D.~Hayes, J.~M. Dreiling, C.~Figgatt, J.~P. Gaebler, S.~A. Moses, J.~M. Pino, and A.~C. Potter.
\newblock {Holographic quantum algorithms for simulating correlated spin systems}.
\newblock \emph{Phys. Rev. Research}, 3:\penalty0 033002, 2021.

\bibitem[Barratt et~al.(2021)Barratt, Dborin, Bal, Stojevic, Pollmann, and Green]{Barratt2021}
F.~Barratt, J.~Dborin, M.~Bal, V.~Stojevic, F.~Pollmann, and A.~G. Green.
\newblock Parallel quantum simulation of large systems on small {NISQ} computers.
\newblock \emph{{npj Quantum Inf.}}, 7\penalty0 (1), 2021.

\bibitem[Chertkov et~al.(2022)Chertkov, Bohnet, Francois, Gaebler, Gresh, Hankin, Lee, Hayes, Neyenhuis, Stutz, Potter, and Foss-Feig]{Chertkov2022}
E.~Chertkov, J.~Bohnet, D.~Francois, J.~Gaebler, D.~Gresh, A.~Hankin, K.~Lee, D.~Hayes, B.~Neyenhuis, R.~Stutz, A.~C. Potter, and M.~Foss-Feig.
\newblock Holographic dynamics simulations with a trapped-ion quantum computer.
\newblock \emph{Nat. Phys.}, 18:\penalty0 1074, 2022.

\bibitem[Niu et~al.(2021)Niu, Haghshenas, Zhang, Foss-Feig, Chan, and Potter]{Niu2021}
D.~Niu, R.~Haghshenas, Y.~Zhang, M.~Foss-Feig, G.~K.-L. Chan, and A.~C. Potter.
\newblock Holographic simulation of correlated electrons on a trapped ion quantum processor.
\newblock \emph{arXiv:2112.10810}, 2021.

\bibitem[Zhang et~al.(2022)Zhang, Jahanbani, Niu, Haghshenas, and Potter]{Zhang2022}
Y.~Zhang, S.~Jahanbani, D.~Niu, R.~Haghshenas, and A.~C. Potter.
\newblock Qubit-efficient simulation of thermal states with quantum tensor networks.
\newblock \emph{arXiv:2205.06299}, 2022.

\bibitem[DeCross et~al.(2022)DeCross, Chertkov, Kohagen, and Foss-Feig]{DeCross2022}
M.~DeCross, E.~Chertkov, M.~Kohagen, and M.~Foss-Feig.
\newblock Qubit-reuse compilation with mid-circuit measurement and reset.
\newblock \emph{arXiv:2210.08039}, 2022.

\bibitem[()Chertkov and Clark]{PhysRevX.8.031029}
E.~Chertkov and B.~K. Clark.
\newblock Computational inverse method for constructing spaces of quantum models from wave functions.
\newblock \emph{Phys. Rev. X}, 8:\penalty0 031029, 2018.

\bibitem[Chertkov et~al.(2020)Chertkov, Villalonga, and Clark]{PhysRevResearch.2.023348}
E.~Chertkov, B.~Villalonga, and B.~K. Clark.
\newblock Engineering topological models with a general-purpose symmetry-to-{H}amiltonian approach.
\newblock \emph{Phys. Rev. Res.}, 2:\penalty0 023348, 2020.

\bibitem[({\natexlab{a}})Qi and Ranard]{Qi2019determininglocal}
X.-L. Qi and D.~Ranard.
\newblock Determining a local {H}amiltonian from a single eigenstate.
\newblock \emph{{Quantum}}, 3:\penalty0 159, 2019{\natexlab{a}}.
\newblock ISSN 2521-327X.

\bibitem[({\natexlab{b}})Powell]{Powell1956}
E.~O. Powell.
\newblock Growth rate and generation time of bacteria, with special reference to continuous culture.
\newblock \emph{Journal of General Microbiology}, 15\penalty0 (3):\penalty0 492--511, 1956{\natexlab{b}}.

\bibitem[Jafarpour et~al.(2018)Jafarpour, Wright, Gudjonson, Riebling, Dawson, Lo, Fiebig, Crosson, Dinner, and Iyer-Biswas]{Jafarpour2018}
F.~Jafarpour, C.~S. Wright, H.~Gudjonson, J.~Riebling, E.~Dawson, K.~Lo, A.~Fiebig, S.~Crosson, A.~R. Dinner, and S.~Iyer-Biswas.
\newblock Bridging the timescales of single-cell and population dynamics.
\newblock \emph{Physical Review X}, 8\penalty0 (2):\penalty0 021007, 2018.

\bibitem[Barber et~al.(2021)Barber, Min, Murray, and Amir]{Barber2021}
F.~Barber, J.~Min, A.~W. Murray, and A.~Amir.
\newblock Modeling the impact of single-cell stochasticity and size control on the population growth rate in asymmetrically dividing cells.
\newblock \emph{PLOS Computational Biology}, 17\penalty0 (6):\penalty0 e1009080, 2021.

\bibitem[()Hein and Jafarpour]{hein2024asymptotic}
Y.~Hein and F.~Jafarpour.
\newblock Asymptotic decoupling of population growth rate and cell size distribution.
\newblock \emph{Physical Review Research}, 6\penalty0 (4):\penalty0 043006, 2024.

\bibitem[Levien et~al.(2025)Levien, He{\"\i}n, and Jafarpour]{levien2025size}
E.~Levien, Y.~He{\"\i}n, and F.~Jafarpour.
\newblock Size-structured populations with growth fluctuations: {Feynman--Kac} formula and decoupling.
\newblock \emph{arXiv preprint arXiv:2508.14680}, 2025.

\bibitem[({\natexlab{a}})Hinshelwood]{hinshelwood1952136}
C.~N. Hinshelwood.
\newblock On the chemical kinetics of autosynthetic systems.
\newblock pages, 745--755, 1952{\natexlab{a}}.

\bibitem[({\natexlab{b}})Gray]{Gray2006CirculantReview}
R.~M. Gray.
\newblock Toeplitz and circulant matrices: A review.
\newblock \emph{Foundations and Trends in Communications and Information Theory}, 2\penalty0 (3):\penalty0 155--239, 2006{\natexlab{b}}.

\bibitem[({\natexlab{c}})Hein and Jafarpour]{hein2024competition}
Y.~Hein and F.~Jafarpour.
\newblock Competition between transient oscillations and early stochasticity in exponentially growing populations.
\newblock \emph{Physical Review Research}, 6\penalty0 (3):\penalty0 033320, 2024{\natexlab{c}}.

\bibitem[Dauxois et~al.(2009)Dauxois, Di~Patti, Fanelli, and McKane]{Dauxois2009AutocatRing}
T.~Dauxois, F.~Di~Patti, D.~Fanelli, and A.~J. McKane.
\newblock Enhanced stochastic oscillations in autocatalytic reactions.
\newblock \emph{Physical Review E}, 79:\penalty0 036112, 2009.

\bibitem[()Togashi and Kaneko]{TogashiKaneko2001}
Y.~Togashi and K.~Kaneko.
\newblock Transitions induced by the discreteness of molecules in a small autocatalytic system.
\newblock \emph{Physical Review Letters}, 86:\penalty0 2459--2462, 2001.

\bibitem[Zhao et~al.(2022)Zhao, Yeo, Huang, and Phillips]{zhao2022thermodynamics}
J.~Zhao, L.~Yeo, E.~W. Huang, and P.~W. Phillips.
\newblock Thermodynamics of an exactly solvable model for superconductivity in a doped mott insulator.
\newblock \emph{Physical Review B}, 105\penalty0 (18):\penalty0 184509, 2022.

\bibitem[Zhao et~al.(2023{\natexlab{a}})Zhao, Mai, Bradlyn, and Phillips]{zhao2023failure}
J.~Zhao, P.~Mai, B.~Bradlyn, and P.~Phillips.
\newblock Failure of topological invariants in strongly correlated matter.
\newblock \emph{Physical review letters}, 131\penalty0 (10):\penalty0 106601, 2023{\natexlab{a}}.

\bibitem[Zhao et~al.(2023{\natexlab{b}})Zhao, La~Nave, and Phillips]{zhao2023proof}
J.~Zhao, G.~La~Nave, and P.~W. Phillips.
\newblock Proof of a stable fixed point for strongly correlated electron matter.
\newblock \emph{Physical Review B}, 108\penalty0 (16):\penalty0 165135, 2023{\natexlab{b}}.

\bibitem[Mai et~al.(2024)Mai, Zhao, Tenkila, Hackner, Kush, Pan, and Phillips]{mai2024new}
P.~Mai, J.~Zhao, G.~Tenkila, N.~A. Hackner, D.~Kush, D.~Pan, and P.~W. Phillips.
\newblock New approach to strong correlation: Twisting {Hubbard} into the orbital {Hatsugai-Kohmoto} model.
\newblock \emph{arXiv preprint arXiv:2401.08746}, 2024.

\bibitem[Ma et~al.(2025)Ma, Zhao, Huang, Kush, Bradlyn, and Phillips]{ma2025charge}
Y.~Ma, J.~Zhao, E.~W. Huang, D.~Kush, B.~Bradlyn, and P.~W. Phillips.
\newblock Charge susceptibility and kubo response in {Hatsugai-Kohmoto-related} models.
\newblock \emph{Physical Review B}, 112\penalty0 (4):\penalty0 045109, 2025.

\bibitem[La~Nave et~al.(2025)La~Nave, Zhao, and Phillips]{la2025luttinger}
G.~La~Nave, J.~Zhao, and P.~W. Phillips.
\newblock The {Luttinger} count is the homotopy not the physical charge: Generalized anomalies characterize non-fermi liquids.
\newblock \emph{arXiv preprint arXiv:2506.04342}, 2025.

\bibitem[Jain et~al.(2023)Jain, Jensen, Liu, and Mefford]{jain2023dipole}
A.~Jain, K.~Jensen, R.~Liu, and E.~Mefford.
\newblock Dipole superfluid hydrodynamics.
\newblock \emph{Journal of High Energy Physics}, 2023\penalty0 (9):\penalty0 1--67, 2023.

\bibitem[Jain et~al.(2024)Jain, Jensen, Liu, and Mefford]{jain2024dipole}
A.~Jain, K.~Jensen, R.~Liu, and E.~Mefford.
\newblock Dipole superfluid hydrodynamics. part ii.
\newblock \emph{Journal of High Energy Physics}, 2024\penalty0 (7):\penalty0 1--50, 2024.

\bibitem[Stahl et~al.(2023)Stahl, Qi, Glorioso, Lucas, and Nandkishore]{stahl2023fracton}
C.~Stahl, M.~Qi, P.~Glorioso, A.~Lucas, and R.~Nandkishore.
\newblock Fracton superfluid hydrodynamics.
\newblock \emph{Physical Review B}, 108\penalty0 (14):\penalty0 144509, 2023.

\bibitem[Glorioso et~al.(2023)Glorioso, Huang, Guo, Rodriguez-Nieva, and Lucas]{glorioso2023goldstone}
P.~Glorioso, X.~Huang, J.~Guo, J.~F. Rodriguez-Nieva, and A.~Lucas.
\newblock Goldstone bosons and fluctuating hydrodynamics with dipole and momentum conservation.
\newblock \emph{Journal of High Energy Physics}, 2023\penalty0 (5):\penalty0 1--43, 2023.

\bibitem[Yang et~al.(2020{\natexlab{a}})Yang, Gorelov, Pierleoni, Ceperley, and Holzmann]{yang2020electronic}
Y.~Yang, V.~Gorelov, C.~Pierleoni, D.~M. Ceperley, and M.~Holzmann.
\newblock Electronic band gaps from quantum {Monte Carlo} methods.
\newblock \emph{Physical Review B}, 101\penalty0 (8):\penalty0 085115, 2020{\natexlab{a}}.

\bibitem[Yang et~al.(2020{\natexlab{b}})Yang, Hiraoka, Matsuda, Holzmann, and Ceperley]{yang2020quantum}
Y.~Yang, N.~Hiraoka, K.~Matsuda, M.~Holzmann, and D.~M. Ceperley.
\newblock Quantum {Monte Carlo} compton profiles of solid and liquid lithium.
\newblock \emph{Physical Review B}, 101\penalty0 (16):\penalty0 165125, 2020{\natexlab{b}}.

\bibitem[Hiraoka et~al.(2020)Hiraoka, Yang, Hagiya, Niozu, Matsuda, Huotari, Holzmann, and Ceperley]{hiraoka2020direct}
N.~Hiraoka, Y.~Yang, T.~Hagiya, A.~Niozu, K.~Matsuda, S.~Huotari, M.~Holzmann, and D.~Ceperley.
\newblock Direct observation of the momentum distribution and renormalization factor in lithium.
\newblock \emph{Physical Review B}, 101\penalty0 (16):\penalty0 165124, 2020.

\bibitem[Holzmann et~al.(2016)Holzmann, Clay~III, Morales, Tubman, Ceperley, and Pierleoni]{holzmann2016theory}
M.~Holzmann, R.~C. Clay~III, M.~A. Morales, N.~M. Tubman, D.~M. Ceperley, and C.~Pierleoni.
\newblock Theory of finite size effects for electronic quantum {Monte Carlo} calculations of liquids and solids.
\newblock \emph{Physical Review B}, 94\penalty0 (3):\penalty0 035126, 2016.

\bibitem[Drummond et~al.(2008)Drummond, Needs, Sorouri, and Foulkes]{drummond2008finite}
N.~Drummond, R.~Needs, A.~Sorouri, and W.~Foulkes.
\newblock Finite-size errors in continuum quantum {Monte Carlo} calculations.
\newblock \emph{Physical Review B—Condensed Matter and Materials Physics}, 78\penalty0 (12):\penalty0 125106, 2008.

\bibitem[Chiesa et~al.(2006)Chiesa, Ceperley, Martin, and Holzmann]{chiesa2006finite}
S.~Chiesa, D.~M. Ceperley, R.~M. Martin, and M.~Holzmann.
\newblock Finite-size error in many-body simulations with long-range interactions.
\newblock \emph{Physical review letters}, 97\penalty0 (7):\penalty0 076404, 2006.

\bibitem[({\natexlab{a}})G{\'e}rard and Lenzmann]{gerard2018lax}
P.~G{\'e}rard and E.~Lenzmann.
\newblock A lax pair structure for the half-wave maps equation.
\newblock \emph{Letters in Mathematical Physics}, 108\penalty0 (7):\penalty0 1635--1648, 2018{\natexlab{a}}.

\bibitem[({\natexlab{b}})Lenzmann and Sok]{lenzmann2020derivation}
E.~Lenzmann and J.~Sok.
\newblock Derivation of the half-wave maps equation from {Calogero--Moser} spin systems.
\newblock \emph{arXiv preprint arXiv:2007.15323}, 2020{\natexlab{b}}.

\bibitem[({\natexlab{c}})Zhou and Stone]{zhou2015solitons}
T.~Zhou and M.~Stone.
\newblock Solitons in a continuous classical {Haldane--Shastry} spin chain.
\newblock \emph{Physics Letters A}, 379\penalty0 (43-44):\penalty0 2817--2825, 2015{\natexlab{c}}.

\bibitem[({\natexlab{d}})Lenzmann]{lenzmann2018short}
E.~Lenzmann.
\newblock A short primer on the half-wave maps equation.
\newblock pages, 1--12, 2018{\natexlab{d}}.

\bibitem[({\natexlab{e}})Stanley]{stanley1986enumerative}
R.~P. Stanley.
\newblock What is enumerative combinatorics?
\newblock pages, 1--63. Springer, 1986{\natexlab{e}}.

\bibitem[({\natexlab{f}})Simion]{simion2000noncrossing}
R.~Simion.
\newblock Noncrossing partitions.
\newblock \emph{Discrete Mathematics}, 217\penalty0 (1-3):\penalty0 367--409, 2000{\natexlab{f}}.

\bibitem[({\natexlab{g}})Zhou and Nahum]{zhou2019emergent}
T.~Zhou and A.~Nahum.
\newblock Emergent statistical mechanics of entanglement in random unitary circuits.
\newblock \emph{Physical Review B}, 99\penalty0 (17):\penalty0 174205, 2019{\natexlab{g}}.

\bibitem[({\natexlab{h}})Hsu and Fradkin]{hsu2013dynamical}
B.~Hsu and E.~Fradkin.
\newblock Dynamical stability of the quantum {Lifshitz} theory in 2+ 1 dimensions.
\newblock \emph{Physical Review B—Condensed Matter and Materials Physics}, 87\penalty0 (8):\penalty0 085102, 2013{\natexlab{h}}.

\bibitem[({\natexlab{i}})Fradkin]{fradkin2009scaling}
E.~Fradkin.
\newblock Scaling of entanglement entropy at 2d quantum {Lifshitz} fixed points and topological fluids.
\newblock \emph{Journal of Physics A: Mathematical and Theoretical}, 42\penalty0 (50):\penalty0 504011, 2009{\natexlab{i}}.

\bibitem[Parker et~al.(2017)Parker, Vasseur, and Moore]{parker2017entanglement}
D.~E. Parker, R.~Vasseur, and J.~E. Moore.
\newblock Entanglement entropy in excited states of the quantum {Lifshitz} model.
\newblock \emph{Journal of Physics A: Mathematical and Theoretical}, 50\penalty0 (25):\penalty0 254003, 2017.

\bibitem[({\natexlab{a}})Brunet]{brunet2016some}
{\'E}.~Brunet.
\newblock \emph{Some aspects of the {Fisher-KPP} equation and the branching {Brownian} motion}.
\newblock PhD thesis, UPMC, 2016{\natexlab{a}}.

\bibitem[({\natexlab{b}})Chatterjee and S.~Dey]{chatterjee2016multiple}
S.~Chatterjee and P.~S.~Dey.
\newblock Multiple phase transitions in long-range first-passage percolation on square lattices.
\newblock \emph{Communications on Pure and Applied Mathematics}, 69\penalty0 (2):\penalty0 203--256, 2016{\natexlab{b}}.

\bibitem[Turner et~al.(2018)Turner, Michailidis, Abanin, Serbyn, and Papi{\'c}]{turner2018weak}
C.~J. Turner, A.~A. Michailidis, D.~A. Abanin, M.~Serbyn, and Z.~Papi{\'c}.
\newblock Weak ergodicity breaking from quantum many-body scars.
\newblock \emph{Nature Physics}, 14\penalty0 (7):\penalty0 745--749, 2018.

\bibitem[Nahum et~al.(2017)Nahum, Ruhman, Vijay, and Haah]{nahum2017quantum}
A.~Nahum, J.~Ruhman, S.~Vijay, and J.~Haah.
\newblock Quantum entanglement growth under random unitary dynamics.
\newblock \emph{Physical Review X}, 7\penalty0 (3):\penalty0 031016, 2017.

\bibitem[({\natexlab{a}})Zhou and Luitz]{zhou2017operator}
T.~Zhou and D.~J. Luitz.
\newblock Operator entanglement entropy of the time evolution operator in chaotic systems.
\newblock \emph{Physical Review B}, 95\penalty0 (9):\penalty0 094206, 2017{\natexlab{a}}.

\bibitem[({\natexlab{b}})NOYES]{noyes1970strong}
H.~P. NOYES.
\newblock page, 1. North-Holland Publishing Company, 1970{\natexlab{b}}.

\bibitem[({\natexlab{c}})Efimov]{PhysRevC.47.1876}
V.~Efimov.
\newblock Effective interaction of three resonantly interacting particles and the force range.
\newblock \emph{Phys. Rev. C}, 47:\penalty0 1876--1884, 1993{\natexlab{c}}.

\bibitem[({\natexlab{d}})Castin and Werner]{PhysRevA.83.063614}
Y.~Castin and F.~Werner.
\newblock Single-particle momentum distribution of an {Efimov} trimer.
\newblock \emph{Phys. Rev. A}, 83:\penalty0 063614, 2011{\natexlab{d}}.

\bibitem[Colussi et~al.(2018)Colussi, Corson, and D'Incao]{PhysRevLett.120.100401}
V.~E. Colussi, J.~P. Corson, and J.~P. D'Incao.
\newblock Dynamics of three-body correlations in quenched unitary bose gases.
\newblock \emph{Phys. Rev. Lett.}, 120:\penalty0 100401, 2018.

\bibitem[Colussi et~al.(2019)Colussi, van Zwol, D'Incao, and Kokkelmans]{PhysRevA.99.043604}
V.~E. Colussi, B.~E. van Zwol, J.~P. D'Incao, and S.~J. J. M.~F. Kokkelmans.
\newblock Bunching, clustering, and the buildup of few-body correlations in a quenched unitary bose gas.
\newblock \emph{Phys. Rev. A}, 99:\penalty0 043604, 2019.

\bibitem[({\natexlab{a}})Fefferman and Graham]{fefferman2012ambient}
C.~Fefferman and C.~R. Graham.
\newblock \emph{The ambient metric ({AM-178})}.
\newblock Princeton University Press, 2012{\natexlab{a}}.

\bibitem[({\natexlab{b}})Graham]{graham2009extended}
C.~R. Graham.
\newblock Extended obstruction tensors and renormalized volume coefficients.
\newblock \emph{Advances in Mathematics}, 220\penalty0 (6):\penalty0 1956--1985, 2009{\natexlab{b}}.

\bibitem[({\natexlab{c}})Jia and Karydas]{jia2021obstruction}
W.~Jia and M.~Karydas.
\newblock Obstruction tensors in {Weyl} geometry and holographic {Weyl} anomaly.
\newblock \emph{Physical Review D}, 104\penalty0 (12):\penalty0 126031, 2021{\natexlab{c}}.

\bibitem[Jia et~al.(2023)Jia, Karydas, and Leigh]{jia2023weyl}
W.~Jia, M.~Karydas, and R.~G. Leigh.
\newblock Weyl-ambient geometries.
\newblock \emph{Nuclear Physics B}, 991:\penalty0 116224, 2023.

\bibitem[({\natexlab{a}})Henningson and Skenderis]{henningson1998holographic}
M.~Henningson and K.~Skenderis.
\newblock The holographic {Weyl} anomaly.
\newblock \emph{Journal of High Energy Physics}, 1998\penalty0 (07):\penalty0 023, 1998{\natexlab{a}}.

\bibitem[({\natexlab{b}})Ciambelli and Leigh]{ciambelli2020weyl}
L.~Ciambelli and R.~G. Leigh.
\newblock Weyl connections and their role in holography.
\newblock \emph{Physical Review D}, 101\penalty0 (8):\penalty0 086020, 2020{\natexlab{b}}.

\bibitem[Safronova et~al.(2012)Safronova, Porsev, and Clark]{safronova2012ytterbium}
M.~Safronova, S.~Porsev, and C.~W. Clark.
\newblock Ytterbium in quantum gases and atomic clocks: van der {W}aals interactions and blackbody shifts.
\newblock \emph{Physical review letters}, 109\penalty0 (23):\penalty0 230802, 2012.

\bibitem[Mitroy et~al.(2010)Mitroy, Safronova, and Clark]{mitroy2010theory}
J.~Mitroy, M.~S. Safronova, and C.~W. Clark.
\newblock Theory and applications of atomic and ionic polarizabilities.
\newblock \emph{Journal of Physics B: Atomic, Molecular and Optical Physics}, 43\penalty0 (20):\penalty0 202001, 2010.

\bibitem[Le~Kien et~al.(2013)Le~Kien, Schneeweiss, and Rauschenbeutel]{le2013dynamical}
F.~Le~Kien, P.~Schneeweiss, and A.~Rauschenbeutel.
\newblock Dynamical polarizability of atoms in arbitrary light fields: general theory and application to cesium.
\newblock \emph{The European Physical Journal D}, 67\penalty0 (5):\penalty0 92, 2013.

\bibitem[Tang et~al.(2018)Tang, Yu, Jiang, and Dong]{Tang_2018}
Z.-M. Tang, Y.-M. Yu, J.~Jiang, and C.-Z. Dong.
\newblock Magic wavelengths for the $6{s}^{2}{}^{1}{S}_{0}\mbox{--}6s6p{}^{3}{P}_{1}^{o}$ transition in ytterbium atom.
\newblock \emph{Journal of Physics B: Atomic, Molecular and Optical Physics}, 51\penalty0 (12):\penalty0 125002, 2018.

\bibitem[Fujita et~al.(2011)Fujita, Takayanagi, and Tonni]{fujita2011aspects}
M.~Fujita, T.~Takayanagi, and E.~Tonni.
\newblock Aspects of {AdS/BCFT}.
\newblock \emph{Journal of High Energy Physics}, 2011\penalty0 (11):\penalty0 1--40, 2011.

\bibitem[({\natexlab{a}})Grinberg and Maldacena]{grinberg2021proper}
M.~Grinberg and J.~Maldacena.
\newblock Proper time to the black hole singularity from thermal one-point functions.
\newblock \emph{Journal of High Energy Physics}, 2021\penalty0 (3):\penalty0 1--31, 2021{\natexlab{a}}.

\bibitem[({\natexlab{b}})Geng and Jiang]{geng2025microscopic}
H.~Geng and Y.~Jiang.
\newblock Microscopic origin of the entropy of single-sided black holes.
\newblock \emph{Journal of High Energy Physics}, 2025\penalty0 (4):\penalty0 1--29, 2025{\natexlab{b}}.

\bibitem[({\natexlab{c}})Chua and Jiang]{chua2024hartle}
W.~Z. Chua and Y.~Jiang.
\newblock {Hartle-Hawking} state and its factorization in 3d gravity.
\newblock \emph{Journal of High Energy Physics}, 2024\penalty0 (3):\penalty0 1--81, 2024{\natexlab{c}}.

\bibitem[({\natexlab{d}})Horvat and Daki{\'c}]{horvat2021interference}
S.~Horvat and B.~Daki{\'c}.
\newblock Interference as an information-theoretic game.
\newblock \emph{Quantum}, 5:\penalty0 404, 2021{\natexlab{d}}.

\bibitem[Chen et~al.(2024)Chen, Zhang, Winter, Lorenz, and Chitambar]{chen2024information}
X.~Chen, Y.~Zhang, A.~Winter, V.~O. Lorenz, and E.~Chitambar.
\newblock Information carried by a single particle in quantum multiple-access channels.
\newblock \emph{Physical Review A}, 109\penalty0 (6):\penalty0 062420, 2024.

\bibitem[Maisriml et~al.(2025)Maisriml, Horvat, and Daki{\'c}]{maisriml2025acquisition}
J.~Maisriml, S.~Horvat, and B.~Daki{\'c}.
\newblock Acquisition of delocalized information via classical and quantum carriers.
\newblock \emph{arXiv preprint arXiv:2506.11254}, 2025.

\bibitem[({\natexlab{a}})Kochen and Specker]{kochen2011problem}
S.~Kochen and E.~P. Specker.
\newblock The problem of hidden variables in quantum mechanics.
\newblock pages, 235--263. Springer, 2011{\natexlab{a}}.

\bibitem[({\natexlab{b}})Spekkens]{spekkens2005contextuality}
R.~W. Spekkens.
\newblock Contextuality for preparations, transformations, and unsharp measurements.
\newblock \emph{Physical Review A—Atomic, Molecular, and Optical Physics}, 71\penalty0 (5):\penalty0 052108, 2005{\natexlab{b}}.

\bibitem[Klyachko et~al.(2008)Klyachko, Can, Binicio{\u{g}}lu, and Shumovsky]{klyachko2008simple}
A.~A. Klyachko, M.~A. Can, S.~Binicio{\u{g}}lu, and A.~S. Shumovsky.
\newblock Simple test for hidden variables in spin-1 systems.
\newblock \emph{Physical review letters}, 101\penalty0 (2):\penalty0 020403, 2008.

\bibitem[()Kunjwal and Spekkens]{kunjwal2015kochen}
R.~Kunjwal and R.~W. Spekkens.
\newblock From the {Kochen-Specker} theorem to noncontextuality inequalities without assuming determinism.
\newblock \emph{Physical review letters}, 115\penalty0 (11):\penalty0 110403, 2015.

\bibitem[Zhang et~al.(2025)Zhang, Y{\=\i}ng, and Schmid]{zhang2025quantifiers}
Y.~Zhang, Y.~Y{\=\i}ng, and D.~Schmid.
\newblock Quantifiers and witnesses for the nonclassicality of measurements and of states.
\newblock \emph{arXiv preprint arXiv:2504.02944}, 2025.

\bibitem[Zhang et~al.()Zhang, Schmid, Y{\i}ng, and Spekkens]{zhang2503reassessing}
Y.~Zhang, D.~Schmid, Y.~Y{\i}ng, and R.~Spekkens.
\newblock Reassessing the boundary between classical and nonclassical for individual quantum processes, arxiv (2025).
\newblock \emph{arXiv preprint arXiv:2503.05884}.

\bibitem[()Dotsenko and Fateev]{dotsenko1984conformal}
V.~S. Dotsenko and V.~A. Fateev.
\newblock Conformal algebra and multipoint correlation functions in 2d statistical models.
\newblock \emph{Nuclear Physics B}, 240\penalty0 (3):\penalty0 312--348, 1984.

\bibitem[Dijkgraaf et~al.(1988)Dijkgraaf, Verlinde, and Verlinde]{dijkgraaf1988c}
R.~Dijkgraaf, E.~Verlinde, and H.~Verlinde.
\newblock C= 1 conformal field theories on {Riemann} surfaces.
\newblock \emph{Communications in Mathematical Physics}, 115\penalty0 (4):\penalty0 649--690, 1988.

\bibitem[Aharony et~al.(2004)Aharony, Marsano, Minwalla, Papadodimas, and RAAMSDONK]{aharony2004hagedorn}
O.~Aharony, J.~Marsano, S.~Minwalla, K.~Papadodimas, and M.~V. RAAMSDONK.
\newblock The {Hagedorn}/deconfinement phase transition in weakly coupled large n gauge theories.
\newblock pages, 161--203. World Scientific, 2004.

\bibitem[Kinney et~al.(2007)Kinney, Maldacena, Minwalla, and Raju]{kinney2007index}
J.~Kinney, J.~Maldacena, S.~Minwalla, and S.~Raju.
\newblock An index for 4 dimensional super conformal theories.
\newblock \emph{Communications in mathematical physics}, 275\penalty0 (1):\penalty0 209--254, 2007.

\bibitem[Yu et~al.(2025{\natexlab{a}})Yu, Herzog-Arbeitman, and Bernevig]{yu2025universal}
J.~Yu, J.~Herzog-Arbeitman, and B.~A. Bernevig.
\newblock Universal {Wilson} loop bound of quantum geometry.
\newblock \emph{Physical Review Letters}, 135\penalty0 (8):\penalty0 086401, 2025{\natexlab{a}}.

\bibitem[Yu et~al.(2025{\natexlab{b}})Yu, Lian, and Ryu]{yu2025wilson}
J.~Yu, B.~Lian, and S.~Ryu.
\newblock Wilson-loop-ideal bands and general idealization.
\newblock \emph{arXiv preprint arXiv:2509.05410}, 2025{\natexlab{b}}.

\bibitem[Yu et~al.(2024)Yu, Bernevig, Queiroz, Rossi, T{\"o}rm{\"a}, and Yang]{yu2024quantum}
J.~Yu, B.~A. Bernevig, R.~Queiroz, E.~Rossi, P.~T{\"o}rm{\"a}, and B.-J. Yang.
\newblock Quantum geometry in quantum materials.
\newblock \emph{arXiv preprint arXiv:2501.00098}, 2024.

\bibitem[()Roy]{roy2014band}
R.~Roy.
\newblock Band geometry of fractional topological insulators.
\newblock \emph{Physical Review B}, 90\penalty0 (16):\penalty0 165139, 2014.

\bibitem[Yang et~al.(2025)Yang, Liu, Schindler, and Liu]{yang2025engineering}
K.~Yang, Y.~Liu, F.~Schindler, and C.-X. Liu.
\newblock Engineering miniband topology via band folding in moir{\'e} superlattice materials.
\newblock \emph{Physical Review B}, 111\penalty0 (24):\penalty0 L241104, 2025.

\bibitem[Hern{\'a}ndez-Garc{\'\i}a et~al.(2013)Hern{\'a}ndez-Garc{\'\i}a, Pic{\'o}n, San~Rom{\'a}n, and Plaja]{hernandez2013attosecond}
C.~Hern{\'a}ndez-Garc{\'\i}a, A.~Pic{\'o}n, J.~San~Rom{\'a}n, and L.~Plaja.
\newblock Attosecond extreme ultraviolet vortices from high-order harmonic generation.
\newblock \emph{Physical review letters}, 111\penalty0 (8):\penalty0 083602, 2013.

\bibitem[Dorney et~al.(2019)Dorney, Rego, Brooks, San~Rom{\'a}n, Liao, Ellis, Zusin, Gentry, Nguyen, Shaw, et~al.]{dorney2019controlling}
K.~M. Dorney, L.~Rego, N.~J. Brooks, J.~San~Rom{\'a}n, C.-T. Liao, J.~L. Ellis, D.~Zusin, C.~Gentry, Q.~L. Nguyen, J.~M. Shaw, et~al.
\newblock Controlling the polarization and vortex charge of attosecond high-harmonic beams via simultaneous spin--orbit momentum conservation.
\newblock \emph{Nature photonics}, 13\penalty0 (2):\penalty0 123--130, 2019.

\bibitem[Rego et~al.(2019)Rego, Dorney, Brooks, Nguyen, Liao, San~Rom{\'a}n, Couch, Liu, Pisanty, Lewenstein, et~al.]{rego2019generation}
L.~Rego, K.~M. Dorney, N.~J. Brooks, Q.~L. Nguyen, C.-T. Liao, J.~San~Rom{\'a}n, D.~E. Couch, A.~Liu, E.~Pisanty, M.~Lewenstein, et~al.
\newblock Generation of extreme-ultraviolet beams with time-varying orbital angular momentum.
\newblock \emph{Science}, 364\penalty0 (6447):\penalty0 eaaw9486, 2019.

\bibitem[Brooks et~al.(2024)Brooks, de~las Heras, Wang, Binnie, Serrano, San~Rom{\'a}n, Plaja, Kapteyn, Hernandez-Garcia, and Murnane]{brooks2024circularly}
N.~J. Brooks, A.~de~las Heras, B.~Wang, I.~Binnie, J.~Serrano, J.~San~Rom{\'a}n, L.~Plaja, H.~C. Kapteyn, C.~Hernandez-Garcia, and M.~M. Murnane.
\newblock Circularly polarized attosecond pulses enabled by an azimuthal phase and polarization grating.
\newblock \emph{ACS Photonics}, 12\penalty0 (1):\penalty0 495--504, 2024.

\bibitem[Hook et~al.(2018)Hook, Kahn, Safdi, and Sun]{PhysRevLett.121.241102}
A.~Hook, Y.~Kahn, B.~R. Safdi, and Z.~Sun.
\newblock Radio signals from axion dark matter conversion in neutron star magnetospheres.
\newblock \emph{Phys. Rev. Lett.}, 121:\penalty0 241102, 2018.

\bibitem[Leroy et~al.(2020)Leroy, Chianese, Edwards, and Weniger]{PhysRevD.101.123003}
M.~Leroy, M.~Chianese, T.~D.~P. Edwards, and C.~Weniger.
\newblock Radio signal of axion-photon conversion in neutron stars: A ray tracing analysis.
\newblock \emph{Phys. Rev. D}, 101:\penalty0 123003, 2020.

\bibitem[Witte et~al.(2021)Witte, Noordhuis, Edwards, and Weniger]{PhysRevD.104.103030}
S.~J. Witte, D.~Noordhuis, T.~D.~P. Edwards, and C.~Weniger.
\newblock Axion-photon conversion in neutron star magnetospheres: The role of the plasma in the {Goldreich-Julian} model.
\newblock \emph{Phys. Rev. D}, 104:\penalty0 103030, 2021.

\bibitem[()Grote and Stadnik]{PhysRevResearch.1.033187}
H.~Grote and Y.~V. Stadnik.
\newblock Novel signatures of dark matter in laser-interferometric gravitational-wave detectors.
\newblock \emph{Phys. Rev. Res.}, 1:\penalty0 033187, 2019.

\bibitem[Aiello et~al.(2022)Aiello, Richardson, Vermeulen, Grote, Hogan, Kwon, and Stoughton]{PhysRevLett.128.121101}
L.~Aiello, J.~W. Richardson, S.~M. Vermeulen, H.~Grote, C.~Hogan, O.~Kwon, and C.~Stoughton.
\newblock Constraints on scalar field dark matter from colocated {Michelson} interferometers.
\newblock \emph{Phys. Rev. Lett.}, 128:\penalty0 121101, 2022.

\bibitem[Vermeulen et~al.(2021)Vermeulen, Relton, Grote, Raymond, Affeldt, Bergamin, Bisht, Brinkmann, Danzmann, Doravari, Kringel, Lough, L{\"u}ck, Mehmet, Mukund, Nadji, Schreiber, Sorazu, Strain, Vahlbruch, Weinert, Willke, and Wittel]{Vermeulen:2021uf}
S.~M. Vermeulen, P.~Relton, H.~Grote, V.~Raymond, C.~Affeldt, F.~Bergamin, A.~Bisht, M.~Brinkmann, K.~Danzmann, S.~Doravari, V.~Kringel, J.~Lough, H.~L{\"u}ck, M.~Mehmet, N.~Mukund, S.~Nadji, E.~Schreiber, B.~Sorazu, K.~A. Strain, H.~Vahlbruch, M.~Weinert, B.~Willke, and H.~Wittel.
\newblock Direct limits for scalar field dark matter from a gravitational-wave detector.
\newblock \emph{Nature}, 600\penalty0 (7889):\penalty0 424--428, 2021.

\bibitem[()Hall and Aggarwal]{hall2022advancedligolisacosmic}
E.~Hall and N.~Aggarwal.
\newblock Advanced {LIGO, LISA, and Cosmic Explorer} as dark matter transducers, 2022.

\bibitem[Morisaki et~al.(2021)Morisaki, Fujita, Michimura, Nakatsuka, and Obata]{PhysRevD.103.L051702}
S.~Morisaki, T.~Fujita, Y.~Michimura, H.~Nakatsuka, and I.~Obata.
\newblock Improved sensitivity of interferometric gravitational-wave detectors to ultralight vector dark matter from the finite light-traveling time.
\newblock \emph{Phys. Rev. D}, 103:\penalty0 L051702, 2021.

\bibitem[Pierce et~al.(2018)Pierce, Riles, and Zhao]{PhysRevLett.121.061102}
A.~Pierce, K.~Riles, and Y.~Zhao.
\newblock Searching for dark photon dark matter with gravitational-wave detectors.
\newblock \emph{Phys. Rev. Lett.}, 121:\penalty0 061102, 2018.

\bibitem[Abbott et~al.(2022)Abbott, Abbott, Acernese, Ackley, Adams, Adhikari, Adhikari, Adya, Affeldt, Agarwal, et~al.]{PhysRevD.105.063030}
R.~Abbott, T.~D. Abbott, F.~Acernese, K.~Ackley, C.~Adams, N.~Adhikari, R.~X. Adhikari, V.~B. Adya, C.~Affeldt, D.~Agarwal, et~al.
\newblock Constraints on dark photon dark matter using data from {LIGO's} and {Virgo's} third observing run.
\newblock \emph{Phys. Rev. D}, 105:\penalty0 063030, 2022.

\bibitem[Miller et~al.(2021)Miller, Astone, Bruno, Clesse, D'Antonio, Depasse, De~Lillo, Frasca, La~Rosa, Leaci, Palomba, Piccinni, Pierini, Rei, and Tanasijczuk]{PhysRevD.103.103002}
A.~L. Miller, P.~Astone, G.~Bruno, S.~Clesse, S.~D'Antonio, A.~Depasse, F.~De~Lillo, S.~Frasca, I.~La~Rosa, P.~Leaci, C.~Palomba, O.~J. Piccinni, L.~Pierini, L.~Rei, and A.~Tanasijczuk.
\newblock Probing new light gauge bosons with gravitational-wave interferometers using an adapted semicoherent method.
\newblock \emph{Phys. Rev. D}, 103:\penalty0 103002, 2021.

\bibitem[Sunko et~al.(2023)Sunko, Sun, Vranas, Homes, Lee, Donoway, Wang, Balguri, Mahendru, Ruiz, et~al.]{sunko2023spin}
V.~Sunko, Y.~Sun, M.~Vranas, C.~C. Homes, C.~Lee, E.~Donoway, Z.-C. Wang, S.~Balguri, M.~B. Mahendru, A.~Ruiz, et~al.
\newblock Spin-carrier coupling induced ferromagnetism and giant resistivity peak in {EuCd}$_2${P}$_2$.
\newblock \emph{Physical Review B}, 107\penalty0 (14):\penalty0 144404, 2023.

\bibitem[Donoway et~al.(2024)Donoway, Trevisan, Liebman-Pel{\'a}ez, Day, Yamakawa, Sun, Soh, Prabhakaran, Boothroyd, Fernandes, et~al.]{donoway2024multimodal}
E.~Donoway, T.~Trevisan, A.~Liebman-Pel{\'a}ez, R.~Day, K.~Yamakawa, Y.~Sun, J.~Soh, D.~Prabhakaran, A.~Boothroyd, R.~Fernandes, et~al.
\newblock Multimodal approach reveals the symmetry-breaking pathway to the broken helix in {EuIn} $_2${As}$_2$.
\newblock \emph{Physical Review X}, 14\penalty0 (3):\penalty0 031013, 2024.

\bibitem[Sie et~al.(2017)Sie, Lui, Lee, Fu, Kong, and Gedik]{sie2017large}
E.~J. Sie, C.~H. Lui, Y.-H. Lee, L.~Fu, J.~Kong, and N.~Gedik.
\newblock Large, valley-exclusive {Bloch-Siegert} shift in monolayer {WS}$_2$.
\newblock \emph{Science}, 355\penalty0 (6329):\penalty0 1066--1069, 2017.

\bibitem[Sie et~al.(2015)Sie, McIver, Lee, Fu, Kong, and Gedik]{sie2015Valley}
E.~J. Sie, J.~W. McIver, Y.-H. Lee, L.~Fu, J.~Kong, and N.~Gedik.
\newblock Valley-selective optical {S}tark effect in monolayer {WS$_2$}.
\newblock \emph{Nature Materials}, 14\penalty0 (3):\penalty0 290--294, 2015.

\bibitem[()Overhauser]{overhauser1971observability}
A.~Overhauser.
\newblock Observability of charge-density waves by neutron diffraction.
\newblock \emph{Physical Review B}, 3\penalty0 (10):\penalty0 3173, 1971.

\bibitem[Pospelov et~al.(2008)Pospelov, Ritz, and Voloshin]{Pospelov:2007mp}
M.~Pospelov, A.~Ritz, and M.~B. Voloshin.
\newblock Secluded {WIMP} dark matter.
\newblock \emph{Phys. Lett. B}, 662:\penalty0 53--61, 2008.

\bibitem[()Shklovskii]{PhysRevB.76.233411}
B.~I. Shklovskii.
\newblock Simple model of {Coulomb} disorder and screening in graphene.
\newblock \emph{Phys. Rev. B}, 76:\penalty0 233411, 2007.

\bibitem[Adam et~al.(2007)Adam, Hwang, Galitski, and Das~Sarma]{adam2007self}
S.~Adam, E.~Hwang, V.~Galitski, and S.~Das~Sarma.
\newblock A self-consistent theory for graphene transport.
\newblock \emph{Proceedings of the National Academy of Sciences}, 104\penalty0 (47):\penalty0 18392--18397, 2007.

\bibitem[Li et~al.(2020)Li, Zhu, Benalcazar, and Hughes]{PhysRevB.101.115115}
T.~Li, P.~Zhu, W.~A. Benalcazar, and T.~L. Hughes.
\newblock Fractional disclination charge in two-dimensional ${C}_{n}$-symmetric topological crystalline insulators.
\newblock \emph{Phys. Rev. B}, 101:\penalty0 115115, 2020.

\bibitem[Zhu et~al.(2020)Zhu, Loehr, and Hughes]{PhysRevB.101.115140}
P.~Zhu, K.~Loehr, and T.~L. Hughes.
\newblock Identifying ${C}_{n}$-symmetric higher-order topology and fractional corner charge using entanglement spectra.
\newblock \emph{Phys. Rev. B}, 101:\penalty0 115140, 2020.

\bibitem[()Zschocke and Vojta]{PhysRevB.92.014403}
F.~Zschocke and M.~Vojta.
\newblock Physical states and finite-size effects in {Kitaev's} honeycomb model: Bond disorder, spin excitations, and nmr line shape.
\newblock \emph{Phys. Rev. B}, 92:\penalty0 014403, 2015.

\bibitem[Zhu et~al.(2025)Zhu, Feng, Wang, Xiang, and Trivedi]{zhu2025emergent}
P.~Zhu, S.~Feng, K.~Wang, T.~Xiang, and N.~Trivedi.
\newblock Emergent quantum {Majorana} metal from a chiral spin liquid.
\newblock \emph{Nature Communications}, 16\penalty0 (1):\penalty0 2420, 2025.

\bibitem[Wang et~al.(2025)Wang, Feng, Zhu, Chi, Liao, Trivedi, and Xiang]{PhysRevB.111.L100402}
K.~Wang, S.~Feng, P.~Zhu, R.~Chi, H.-J. Liao, N.~Trivedi, and T.~Xiang.
\newblock Fractionalization signatures in the dynamics of quantum spin liquids.
\newblock \emph{Phys. Rev. B}, 111:\penalty0 L100402, 2025.

\bibitem[Feng et~al.(2025)Feng, Zhu, Knolle, and Knap]{feng2025transient}
S.~Feng, P.~Zhu, J.~Knolle, and M.~Knap.
\newblock Transient localization from fractionalization: vanishingly small heat conductivity in gapless quantum magnets.
\newblock \emph{arXiv preprint arXiv:2509.07062}, 2025.

\bibitem[({\natexlab{a}})Su and MacDonald]{su2017spatially}
J.-J. Su and A.~H. MacDonald.
\newblock Spatially indirect exciton condensate phases in double bilayer graphene.
\newblock \emph{Physical Review B}, 95\penalty0 (4):\penalty0 045416, 2017{\natexlab{a}}.

\bibitem[({\natexlab{b}})Abbamonte and Fink]{abbamonte2025collective}
P.~Abbamonte and J.~Fink.
\newblock Collective charge excitations studied by electron energy-loss spectroscopy.
\newblock \emph{Annual Review of Condensed Matter Physics}, 16\penalty0 (1):\penalty0 465--480, 2025{\natexlab{b}}.

\bibitem[Mitrano et~al.(2024)Mitrano, Johnston, Kim, and Dean]{mitrano2024exploring}
M.~Mitrano, S.~Johnston, Y.-J. Kim, and M.~Dean.
\newblock Exploring quantum materials with resonant inelastic x-ray scattering.
\newblock \emph{Physical Review X}, 14\penalty0 (4):\penalty0 040501, 2024.

\bibitem[Zhou et~al.(2025)Zhou, Chessa, Chitambar, and Leditzky]{zhou2025distinguishability}
J.~Zhou, S.~Chessa, E.~Chitambar, and F.~Leditzky.
\newblock On the distinguishability of geometrically uniform quantum states, 2025.

\bibitem[Hirche et~al.(2022)Hirche, Rouz{\'e}, and Fran{\c{c}}a]{hirche2022contraction}
C.~Hirche, C.~Rouz{\'e}, and D.~S. Fran{\c{c}}a.
\newblock On contraction coefficients, partial orders and approximation of capacities for quantum channels.
\newblock \emph{Quantum}, 6:\penalty0 862, 2022.

\bibitem[Wu et~al.(2019)Wu, Lovorn, Tutuc, Martin, and MacDonald]{PhysRevLett.122.086402}
F.~Wu, T.~Lovorn, E.~Tutuc, I.~Martin, and A.~H. MacDonald.
\newblock Topological insulators in twisted transition metal dichalcogenide homobilayers.
\newblock \emph{Phys. Rev. Lett.}, 122:\penalty0 086402, 2019.

\bibitem[Reddy et~al.(2023)Reddy, Alsallom, Zhang, Devakul, and Fu]{PhysRevB.108.085117}
A.~P. Reddy, F.~Alsallom, Y.~Zhang, T.~Devakul, and L.~Fu.
\newblock Fractional quantum anomalous hall states in twisted bilayer {MoTe}$_{2}$ and {WSe}$_{2}$.
\newblock \emph{Phys. Rev. B}, 108:\penalty0 085117, 2023.

\bibitem[Wang et~al.(2024)Wang, Zhang, Liu, He, Xu, Ran, Cao, and Xiao]{PhysRevLett.132.036501}
C.~Wang, X.-W. Zhang, X.~Liu, Y.~He, X.~Xu, Y.~Ran, T.~Cao, and D.~Xiao.
\newblock Fractional {C}hern insulator in twisted bilayer {MoTe}$_{2}$.
\newblock \emph{Phys. Rev. Lett.}, 132:\penalty0 036501, 2024.

\bibitem[Yu et~al.(2024)Yu, Herzog-Arbeitman, Wang, Vafek, Bernevig, and Regnault]{PhysRevB.109.045147}
J.~Yu, J.~Herzog-Arbeitman, M.~Wang, O.~Vafek, B.~A. Bernevig, and N.~Regnault.
\newblock Fractional {C}hern insulators versus nonmagnetic states in twisted bilayer {MoTe}$_{2}$.
\newblock \emph{Phys. Rev. B}, 109:\penalty0 045147, 2024.

\bibitem[Jia et~al.(2024)Jia, Yu, Liu, Herzog-Arbeitman, Qi, Pi, Regnault, Weng, Bernevig, and Wu]{PhysRevB.109.205121}
Y.~Jia, J.~Yu, J.~Liu, J.~Herzog-Arbeitman, Z.~Qi, H.~Pi, N.~Regnault, H.~Weng, B.~A. Bernevig, and Q.~Wu.
\newblock Moir\'e fractional {C}hern insulators. i. first-principles calculations and continuum models of twisted bilayer {MoTe}$_{2}$.
\newblock \emph{Phys. Rev. B}, 109:\penalty0 205121, 2024.

\bibitem[({\natexlab{a}})AI~Security~Institute]{UK_AI_Security_Institute_Inspect_AI_Framework_2024}
U.~AI~Security~Institute.
\newblock Inspect {AI:} framework for large language model evaluations, {\natexlab{a}}.
\newblock URL \url{https://github.com/UKGovernmentBEIS/inspect_ai}.

\bibitem[({\natexlab{b}}){Together AI}]{togetherai}
{Together AI}.
\newblock {Together AI Platform}.
\newblock \url{https://www.together.ai/}, 2025{\natexlab{b}}.
\newblock Accessed: 2025-08-24.

\end{thebibliography}
